%% file: main.tex
\documentclass{article}
\input{packages.tex}
\input{defs.tex}

\icmltitlerunning{\ourframework: Understanding and Improving Multi-Agent Reasoning}

\begin{document}

\twocolumn[
  \icmltitle{\ourframework: Understanding and Improving Multi-Agent Reasoning Through Holistic Orchestration and Controlled Benchmarks}



  \icmlsetsymbol{equal}{*}

  \begin{icmlauthorlist}
    \icmlauthor{Zixuan Ke}{yyy}
    \icmlauthor{Yifei Ming}{yyy}
    \icmlauthor{Austin Xu}{yyy}
    \icmlauthor{Ryan Chin}{sch}
    \icmlauthor{Xuan-Phi Nguyen}{yyy}
    \icmlauthor{Prathyusha Jwalapuram}{yyy}
    \icmlauthor{Jiayu Wang}{comp}
    \icmlauthor{Semih Yavuz}{yyy}
    \icmlauthor{Caiming Xiong}{yyy}
    \icmlauthor{Shafiq Joty}{yyy}
  \end{icmlauthorlist}

  \icmlaffiliation{yyy}{Salesforce Research}
  \icmlaffiliation{comp}{University of Wisconsin-Madison}
  \icmlaffiliation{sch}{Massachusetts Institute of Technology}

  \icmlcorrespondingauthor{Zixuan Ke}{zixuan.ke@salesforce.com}
  \icmlcorrespondingauthor{Shafiq Joty}{sjoty@salesforce.com}

  \begin{center}
    {\includegraphics[width=0.5cm]{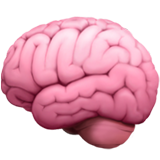} Project page: \projectpage} \\
    {\includegraphics[width=0.5cm]{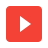} Live Demo: \demopage} \\
    {\includegraphics[width=0.5cm]{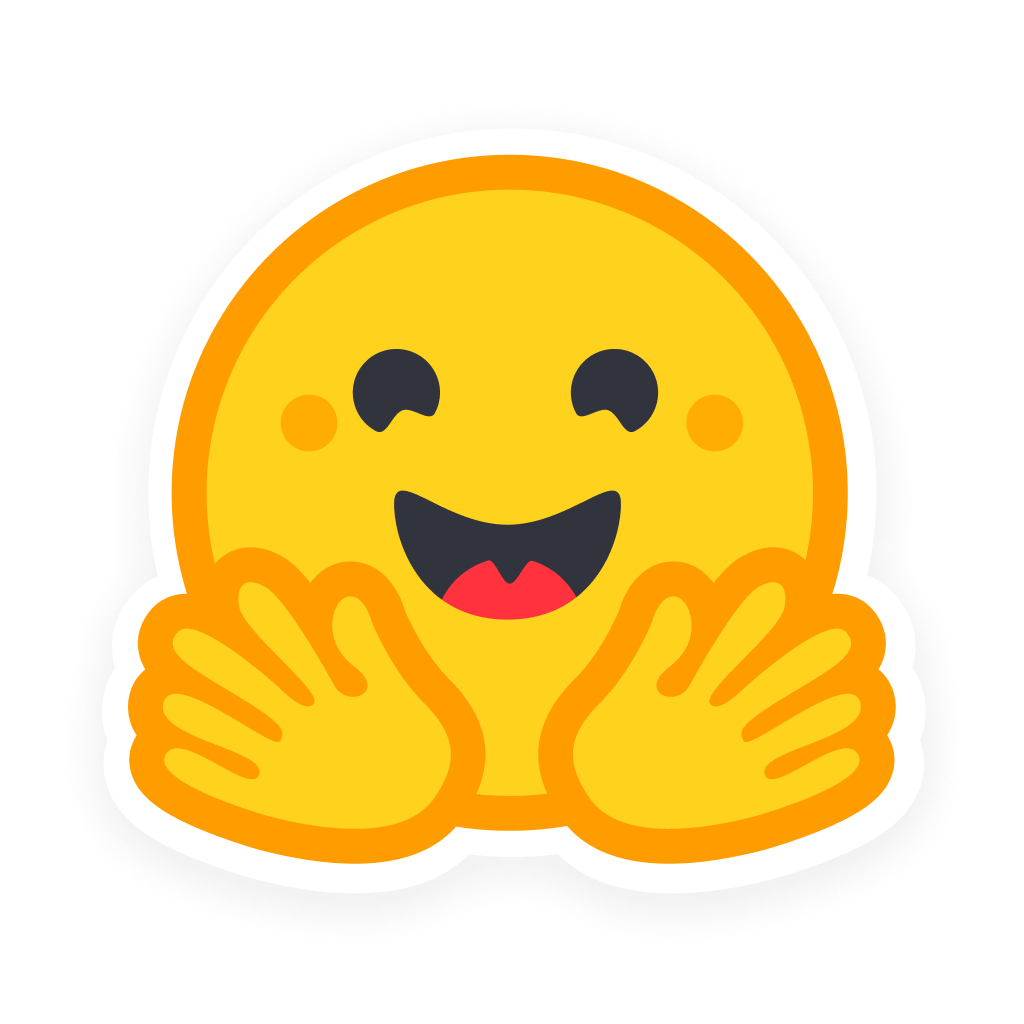} MASBench: \masbench} \\
    {\includegraphics[width=0.5cm]{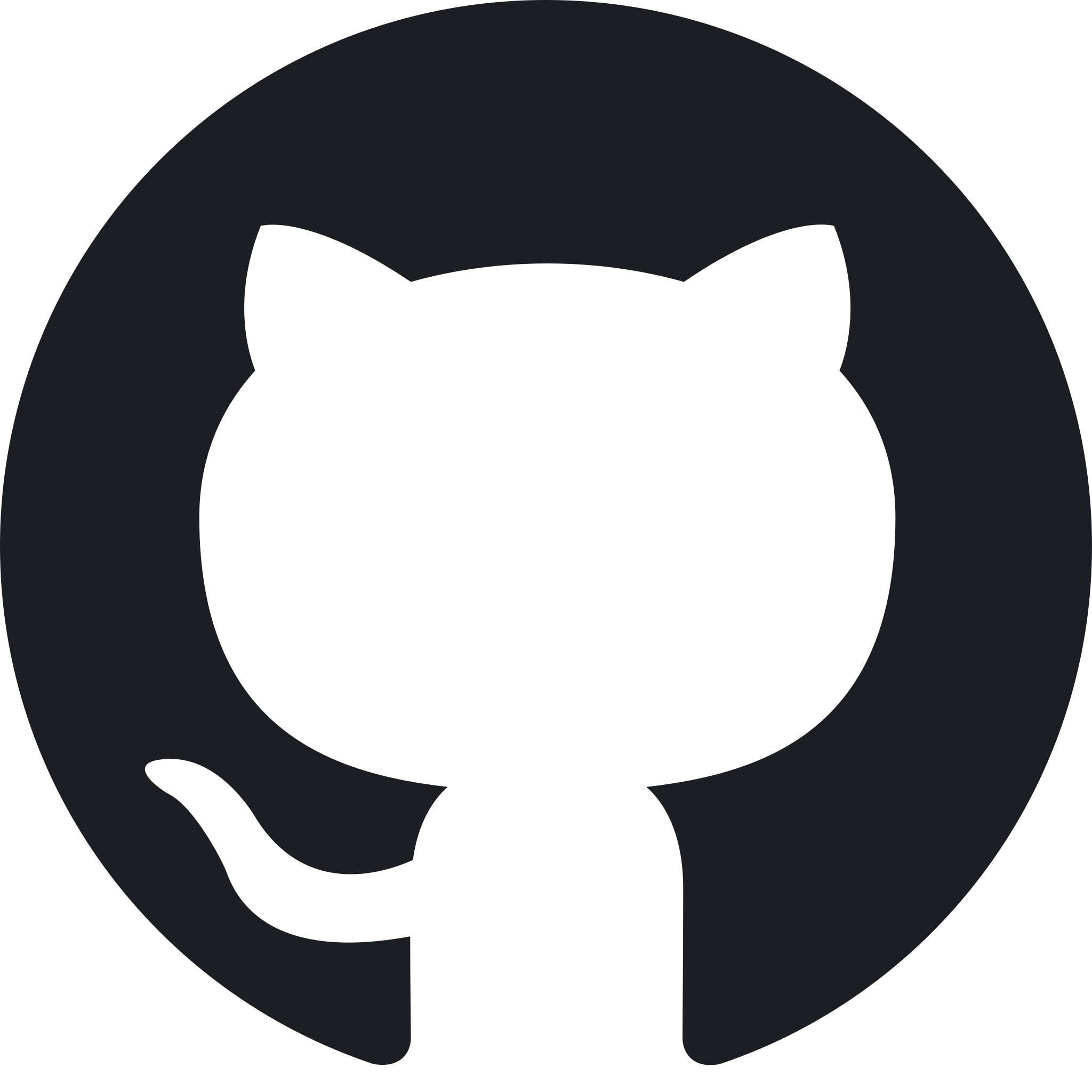} Code: \projectcode}
  \end{center}

  \icmlkeywords{Multi-agent System, Reinforcement Learning}

  \vskip 0.3in
]



\printAffiliationsAndNotice{}  

\input{sections/abstract}

\input{sections/1_introduction}

\input{sections/2_related}

\input{sections/3_framework}

\input{sections/4_analysis_framework}
\input{sections/4_analysis}

\input{sections/5_experiment}
\input{sections/6_conclusion}

\bibliography{mas_zero,mas_r1,bgm}
\bibliographystyle{icml2026}

\newpage
\appendix
\onecolumn

\counterwithin{table}{section}
\renewcommand{\thetable}{\Alph{section}.\arabic{table}}
\counterwithin{figure}{section}
\renewcommand{\thefigure}{\Alph{section}.\arabic{figure}}

\begin{center}
\textbf{\LARGE Appendix }

\end{center}
\crefalias{section}{appendix}
\input{sections/appendix}

\end{document}

%% file: packages.tex
\usepackage{microtype}
\usepackage{graphicx}
\usepackage{subcaption}
\usepackage{booktabs} 

\usepackage[accepted]{icml2026}

\usepackage{amsmath}
\usepackage{amssymb}
\usepackage{mathtools}
\usepackage{amsthm}

\theoremstyle{plain}

\theoremstyle{definition}

\theoremstyle{remark}

\usepackage[textsize=tiny]{todonotes}

\usepackage[utf8]{inputenc} 
\usepackage[T1]{fontenc}    
\usepackage{url}            
\usepackage{booktabs}       
\usepackage{amsfonts}       
\usepackage{nicefrac}       
\usepackage{microtype}      

\usepackage{inconsolata}
\usepackage{url}
\usepackage{booktabs}       
\usepackage{amsfonts}       
\usepackage{nicefrac}       
\usepackage{microtype}      
\usepackage{url}            
\usepackage{booktabs}       
\usepackage{amsfonts}       
\usepackage{nicefrac}       
\usepackage{microtype}      
\usepackage{subcaption}
\usepackage{amsmath,amsfonts,amssymb}
\usepackage{tabularx}
\usepackage{multirow}
\usepackage{graphicx}
\usepackage{color}
\usepackage{CJK}
\usepackage{adjustbox}
\usepackage{colortbl}
\usepackage{multicol}
\usepackage{vwcol} 
\newsavebox{\algleft}
\newsavebox{\algright}
\usepackage{bm}
\usepackage{booktabs}
\usepackage{amsmath,stackengine}
\usepackage{times}
\usepackage{latexsym}
\usepackage{bbm}
\usepackage[T1]{fontenc}    
\usepackage{url}            
\usepackage{booktabs}       
\usepackage{amsfonts}       
\usepackage{nicefrac}       
\usepackage{microtype}      
\usepackage{amssymb}
\usepackage{pifont}
\usepackage{bbm}
\usepackage{breqn}
\usepackage{longtable}
\usepackage{enumitem}
\usepackage[table,xcdraw]{xcolor}
\usepackage{microtype}
\usepackage{enumitem}
\usepackage{amsmath}
\usepackage{chngcntr} 
\usepackage{wrapfig}
\usepackage{listings}
\usepackage{arydshln}
\usepackage{amsmath} 

\usepackage{booktabs}
\usepackage{longtable}
\usepackage{xspace}

\usepackage[most]{tcolorbox}
\usepackage{makecell}
\usepackage{titletoc}

\usepackage{hyperref}
\usepackage[capitalize,noabbrev]{cleveref}

%% file: defs.tex
\definecolor{BlueBG}{rgb}{0,0.46,0.71}
\newcommand{\ourframework}{{\textsc{Mas-}Orchestra}\xspace} 
\newcommand{\ourbenchmark}{{\textsc{MasBench}}\xspace} 

\definecolor{innerboxcolor}{HTML}{00B2E6} 
\definecolor{outerboxcolor}{HTML}{D7F6FF} 
\definecolor{lightblue}{HTML}{D7F6FF}

\usepackage{tikz}
\usetikzlibrary{decorations.pathmorphing}

\definecolor{upblue}{HTML}{388E3C}
\definecolor{downred}{HTML}{E53935}

\definecolor{darkgreen}{RGB}{0,128,0}  
\definecolor{darkred}{RGB}{139,0,0}  

\definecolor{myblue}{RGB}{31,119,180}    
\definecolor{mygreen}{RGB}{44,160,44}    
\definecolor{myorange}{RGB}{255,127,14}  
\definecolor{mypurple}{RGB}{153, 102, 204}  

\newtcolorbox[auto counter]{observation}[1][]{
  breakable,
  colback=black!5!white,
  colframe=black!70!white,
  fonttitle=\bfseries,
  enhanced,
  boxrule=0.6pt,
  left=1mm,right=1mm,top=1mm,bottom=1mm,
  #1
}

\newtcolorbox[auto counter]{takeaway}[1][]{
  breakable,
  colback=teal!3!white,
  colframe=teal!55!black,
  fonttitle=\bfseries,
  title=Takeaway~\thetcbcounter,
  enhanced,
  boxrule=0.5pt,
  left=1mm,right=1mm,top=1mm,bottom=1mm,
  #1
}

\newtcolorbox[auto counter]{practicalguidance}[1][]{
  breakable,
  colback=cyan!3!white,
  colframe=cyan!60!black,
  fonttitle=\bfseries,
  title=Practical Guidance~\thetcbcounter,
  enhanced,
  boxrule=0.5pt,
  left=1mm,right=1mm,top=1mm,bottom=1mm,
  #1
}

\newtcolorbox[auto counter]{discussion}[1][]{
  breakable,
  colback=violet!4!white,
  colframe=violet!60!black,
  fonttitle=\bfseries,
  title=Discussion~\thetcbcounter,
  enhanced,
  boxrule=0.5pt,
  left=1mm,right=1mm,top=1mm,bottom=1mm,
  #1
}

\newtcolorbox[auto counter]{developprompt}[1][]{
  breakable,
    colback=white,
    colframe=black!70,
  fonttitle=\bfseries,
  title=Prompt~\thetcbcounter,
  enhanced,
  boxrule=0.5pt,
  left=1mm,right=1mm,top=1mm,bottom=1mm,
  #1
}

\newtcolorbox[auto counter]{systemprompt}[1][]{
  breakable,
    colback=orange!6,
    colframe=orange!60!black,
  fonttitle=\bfseries,
  title=System Prompt~\thetcbcounter,
  enhanced,
  boxrule=0.5pt,
  left=1mm,right=1mm,top=1mm,bottom=1mm,
  #1
}

\newtcolorbox[auto counter]{userprompt}[1][]{
  breakable,
    colback=green!5,
    colframe=green!40!black,
  fonttitle=\bfseries,
  title=User Prompt~\thetcbcounter,
  enhanced,
  boxrule=0.5pt,
  left=1mm,right=1mm,top=1mm,bottom=1mm,
  #1
}

\newtcolorbox[auto counter]{answerprompt}[1][]{
  breakable,
  colback=purple!6,
  colframe=purple!50!black,
  fonttitle=\bfseries,
  title=Answer Prompt~\thetcbcounter,
  enhanced,
  boxrule=0.5pt,
  left=1mm,right=1mm,top=1mm,bottom=1mm,
  #1
}
\newtcolorbox[auto counter]{subagent}[1][]{
  breakable,
  colback=cyan!3!white,
  colframe=cyan!60!black,
  fonttitle=\bfseries,
  title=Sub-agent~\thetcbcounter,
  enhanced,
  boxrule=0.5pt,
  left=1mm,right=1mm,top=1mm,bottom=1mm,
  #1
}

\newcommand{\xmltag}[1]{\textcolor{green!50!black}{\texttt{<#1>}}}
\newcommand{\xmlendtag}[1]{\textcolor{green!50!black}{\texttt{</#1>}}}

\newcommand{\code}[1]{\texttt{#1}}

\newcommand{\pykw}[1]{\textcolor{blue!70!black}{\texttt{#1}}}
\newcommand{\pyfn}[1]{\textcolor{purple!70!black}{\texttt{#1}}}
\newcommand{\pystr}[1]{\textcolor{orange!70!black}{\texttt{#1}}}
\newcommand{\pycom}[1]{\textcolor{gray}{\texttt{#1}}}

\newcommand{\jsonkey}[1]{\textcolor{green!50!black}{\texttt{"#1"}}}
\newcommand{\jsonstr}[1]{\textcolor{orange!70!black}{\texttt{"#1"}}}

\newcommand{\avgat}[1]{\texttt{Avg@#1}}  

\newcommand{\qwen}{{Qwen-7b}\xspace} 
\newcommand{\deepseekqwen}{{DS-7b}\xspace} 
\newcommand{\osstwenty}{{GPT-20b (low)}\xspace} 
\newcommand{\ossonetwenty}{{GPT-120b}\xspace} 
\newcommand{\ossonetwentylow}{{GPT-120b (low)}\xspace}

\newcommand{\masness}{{DoM}\xspace} 

\newcommand{\cmark}{\textcolor{green!70!black}{\ding{51}}} 
\newcommand{\xmark}{\textcolor{red!80!black}{\ding{55}}}   

\newcolumntype{Y}{>{\centering\arraybackslash}X} 

\newcommand{\depth}{\textit{Depth}\xspace} 
\newcommand{\breadth}{\textit{Breadth}\xspace} 
\newcommand{\horizon}{\textit{Horizon}\xspace} 
\newcommand{\myparallel}{\textit{Parallel}\xspace} 
\newcommand{\robustness}{\textit{Robustness}\xspace}

\newcommand{\projectpage}{\url{https://mas-orchestra.salesforceresearch.ai/mas_r1/}}
\newcommand{\projectcode}{\url{https://github.com/SalesforceAIResearch/MAS-Orchestra}}
\newcommand{\masbench}{\url{https://huggingface.co/datasets/Salesforce/MASBench}}
\newcommand{\demopage}{\url{https://mas-orchestra.salesforceresearch.ai/mas_r1/demo/}}

\newcommand{\examplebox}[1]{%
  \parbox[t]{\linewidth}{\scriptsize\raggedright #1}%
}

%% file: sections/abstract.tex
\begin{abstract}



While multi-agent systems (MAS) promise elevated intelligence through coordination of agents, current approaches to automatic MAS design under-deliver. Such shortcomings stem from two key factors: (1) methodological complexity -- agent orchestration is performed using sequential, code-level execution that limits global system-level holistic reasoning and scales poorly with agent complexity -- and (2) efficacy uncertainty -- MAS are deployed without understanding if there are tangible benefits compared to single-agent systems (SAS). We propose \ourframework, a training-time framework that formulates MAS orchestration as a function-calling reinforcement learning problem with holistic orchestration, generating an entire MAS at once. In \ourframework, complex, goal-oriented sub-agents are abstracted as callable functions, enabling global reasoning over system structure while hiding internal execution details. To rigorously study when and why MAS are beneficial, we introduce \ourbenchmark, a controlled benchmark that characterizes tasks along five axes: \depth, \horizon, \breadth, \myparallel, and \robustness. Our analysis reveals that MAS gains depend critically on task structure, verification protocols, and the capabilities of both orchestrator and sub-agents, rather than holding universally. Guided by these insights, \ourframework achieves consistent improvements on public benchmarks including mathematical reasoning,  multi-hop QA, and search-based QA, {while achieving more than \textit{10$\times$} efficiency over strong baselines}. Together, \ourframework and \ourbenchmark enable better training and understanding of MAS in the pursuit of multi-agent intelligence.

\end{abstract}

%% file: sections/1_introduction.tex
\section{Introduction}
\label{sec.intro}

\begin{figure}[t]
\centering
\includegraphics[width=0.9\columnwidth]{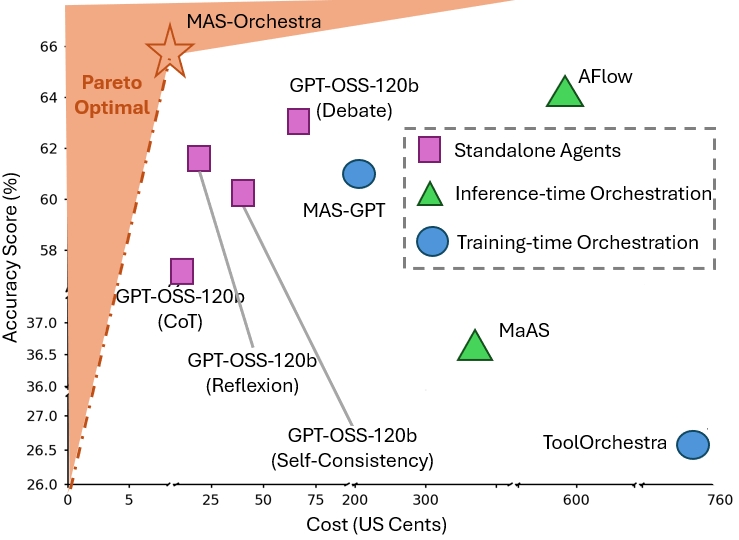}
\vspace{-0.5\baselineskip}
\captionof{figure}{\small {Cost–performance trade-off computed using average accuracy and total inference cost on AIME24 and GPQA. \ourframework lie on the Pareto frontier, delivering higher accuracy at lower or comparable cost than strong baselines (see \cref{ap:cost_breakdown}). All training-time orchestration models use the released orchestrator, which may have been trained in a different environment than ours.}}
\label{fig:parato_front}
\vspace{-1\baselineskip}
\end{figure}

\begin{figure*}[h]
\centering
\includegraphics[width=\textwidth]{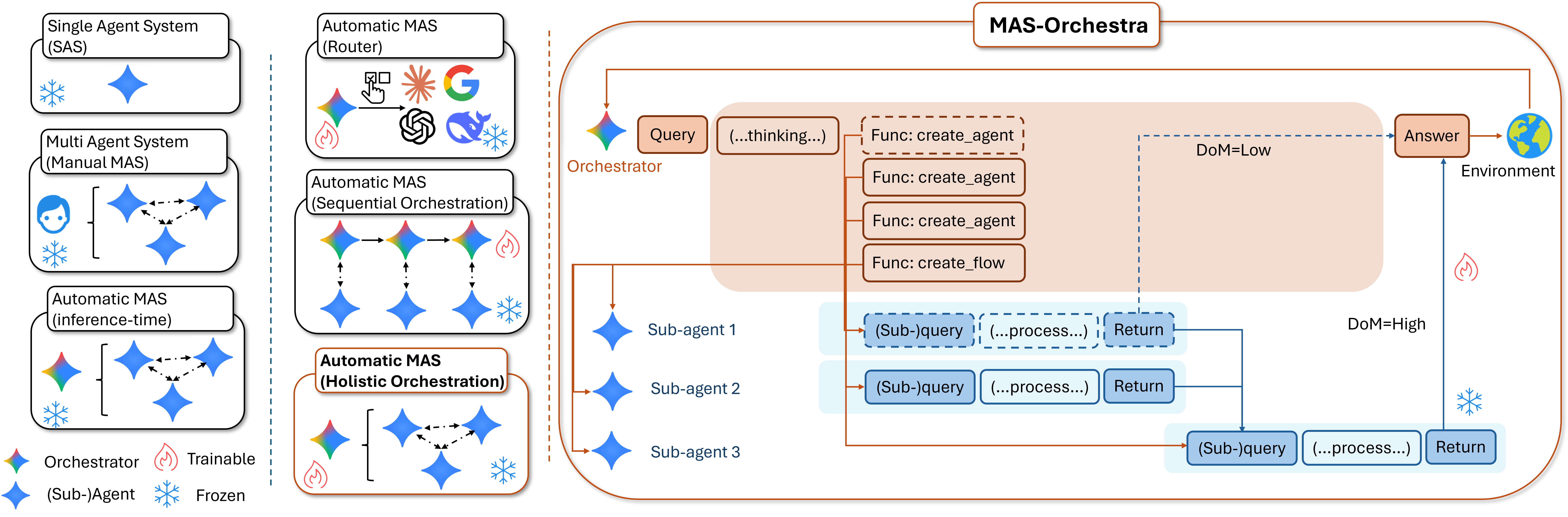}
\vspace{-15pt}
\captionof{figure}{\small Paradigm comparison and \ourframework. \textbf{Left:} {Inference-time orchestration systems typically adopt holistic orchestration, but without training.} \ourframework lies in automatic MAS and formulate the problem as a function-calling RL problem with holistic orchestration. \textbf{Right:} When \masness is configured to be \textit{low} (dashed lines), the system instantiates at most one agent. When \masness is \textit{high}, the number of agents is unconstrained. {The details of the function-calling protocol and the parser are provided in Appendices \ref{app:prompt} and \ref{app:parser}.}
}
\label{fig:main}
\vspace{-1\baselineskip}
\end{figure*}

We have observed a clear progression from standalone large language models (LLM), to LLM-based  single-agent systems (SAS), and more recently to multi-agent systems (MAS) \citep{ke2025surveyfrontiersllmreasoning}. This shift reflects the increasing complexity of modern tasks, which often require sustained reasoning, task decomposition, and interaction with other specialized agents. {Early MAS designs were largely manually constructed, with fixed interaction patterns like debate \citep{du2023improvingfactualityreasoning}. Recently, research attention has increasingly shifted toward \textit{automatic MAS design}, where the system structure, agent roles, and interactions are generated dynamically, as manual specification is often labor-intensive and does not scale well to novel problems or  domains.}

While automatic MAS have shown strong empirical promise, current approaches remain limited in several important aspects: (1) \textbf{Formulation}: Most existing work \citep{Ke2025MASZero,nie2025weakforstrong,hu2024ADAS} adopts \textit{executable code} as the formulation for agent orchestration. While this  {code-based} formulation provides flexibility, it leads to significant overhead and becomes difficult to scale when sub-agents and their inter-connections grow more complex (e.g., agents that perform multi-turn search) as the orchestrator often needs  to analyze or even reproduce the full sub-agent code during orchestration. 
 As a result, they are often limited to relatively simple sub-agents such as sub-agents that are based on Chain-of-Thought (CoT) \citep{wei2022chain} and CoT with Self-Consistency (SC) \citep{wang2023selfconsistency}. (2) \textbf{Training}: Many  existing  approaches rely heavily on inference-time self-improvement or heuristic search~\citep{zhang2024aflowautomatingagenticworkflow,liu2024dynamicllmpowered}. Without explicit training objectives, such adaptation can be unstable, directionally incorrect, and computationally expensive. Although some recent studies incorporate training into MAS design \citep{dang2025multiagentcollaborationevolvingorchestration,chi2025eraagenticorganizationlearning}, they typically formulate orchestration as a sequential decision process—that is, incrementally adding components one at a time—and optimize it via multi-turn reinforcement learning. While effective in certain settings, this paradigm introduces substantial training overhead and encourages local, step-wise optimization, making it difficult for the orchestrator to reason about the MAS structure from a global perspective; 
(3) \textbf{Agent-scaling principles} \citep{li2024more,qian2025scaling}. Despite the rapid adoption of MAS, there is still no principled quantitative framework that predicts when adding subagents improves performance and when it degrades it. As a result, practitioners often rely on heuristics, which limits both scientific understanding and practical deployment. It remains unclear when multi-agent coordination provides genuine value over simpler single-agent alternatives, and when it merely adds complexity {\citep{kim2025sciencescalingagentsystems,huang2024large}.} 

We argue that an effective automatic MAS should satisfy three core desiderata: (1) there should be an \textbf{explicit notion of ``degree of MAS (\masness)''}, capturing the degree of multi-agent coordination appropriate for a given task. This is because not all problems benefit from MAS. For example, our pilot experiments (\cref{sec:experiemnt}) suggest that even challenging mathematical tasks, such as those in AIME, often gain little from multi-agent coordination; (2) \textbf{the orchestration mechanism should be flexible and scalable}. The formulation should remain effective even when sub-agents are highly complex. In particular, the orchestrator LLM should \textit{only} focus on high-level reasoning and system design. Its role should be to decide when and which sub-agents to invoke and how they are connected, rather than reproducing the internal behavior of the sub-agents themselves; (3) \textbf{A sub-agent should be defined by its own goal, and therefore should own its context, tools, and workflows} \cite{anthropic_subagents_2024}. 
 {Most existing methods define sub-agents either in the \textit{LLM space}~\cite{jin2025controllingperformancebudgetcentralized,estornell2025trainleaderhierarchicalreasoning,dang2025multiagentcollaborationevolvingorchestration} by switching between different backbone models, which reduces the orchestrator to merely an LLM router, or in the \textit{prompt
template space}, where sub-agents differ only by superficial prompt variations (e.g., in Debate, different roles are often instantiated by
changing only the role description: ``You are a
critic'' vs. ``You are a supporter''). Such definitions overlook the tool usage and workflows that are also critical for specializing a sub-agent toward its goal.}



In this paper, we propose \textbf{\ourframework},
which formulates MAS orchestration as a \textbf{function-calling reinforcement learning (RL) problem}. Under this formulation, \textit{arbitrarily complex and goal-oriented} sub-agents (e.g., reasoning agent and search agent) are encapsulated as callable functions, with their internal mechanisms abstracted away. The orchestrator only decides when to instantiate a sub-agent, which sub-agents to create, and how they are connected by calling two primitive functions \texttt{create_agent} and \texttt{create_flow}. We also introduce \textbf{\masness} into the formulation, where the user can configure over \textit{low} (up to one sub-agent is used) and \textit{high \masness}. Based on this formulation, instead of casting MAS design as a sequence of decisions, we adopt \textbf{holistic orchestration}, where the orchestrator generates the \textit{complete}  orchestration \textbf{in a single decision step} {(see \cref{fig:main} for a comparison}). 
This design enables the orchestrator to reason about the full system configuration jointly, rather than optimizing a sequence of local decisions. As a result, the learned orchestration reflects global coordination patterns rather than execution-order effects. {From a learning perspective, this simplifies optimization compared
to multi-step approaches, which encourage local, step-wise optimization and
suffer from long-horizon credit assignment and error propagation. Consequently,
it leads to more stable training and improved scalability as MAS
complexity increases{, as evidenced by its superior performance over sequential orchestration baselines  (\cref{sec:experiemnt}).} {It also yields substantial efficiency improvements (more than 10$\times$), placing it on the Pareto frontier, as shown in \cref{fig:parato_front}.}


To quantitatively analyze when and why MAS outperform single-agent systems, we introduce a controlled benchmark, \ourbenchmark, which characterizes problems along five axes—\textbf{\depth, \horizon, \breadth, \myparallel, and \robustness}—enabling systematic and controlled evaluation of MAS behavior.
Our analysis yields several key insights. For example, we find that the benefits of multi-agent coordination depend strongly on {task structure and the underlying LLM capability}. Guided by these insights, we conduct experiments on public benchmarks and demonstrate consistent performance gains under appropriate MAS configurations. 

In summary, our key contributions are:
\begin{itemize}[leftmargin=*,noitemsep,topsep=2pt]
  \item We introduce a novel, scalable and effective RL formulation for MAS orchestration, featuring an explicit notion of \textit{\masness} and a function-calling abstraction that encapsulates complex sub-agents.
  \item We propose a controlled benchmark \ourbenchmark, tailored for MAS evaluation, enabling systematic empirical analysis of when MAS outperforms single-agent systems—and when it does not. This is the first benchmark established to evaluate MAS benefits to the best of our knoweldge.
  \item {Using \ourbenchmark, we investigate the benefits of MAS across 3 analysis directions over a broad range of MAS configurations, covering 3 orchestrator settings and 5 sub-agent settings across different model sizes and families.}
  \item {Guided by insights from the above analysis}, we demonstrate that our approach achieves strong performance on public benchmarks, including math, multi-hop question answering, and multi-step search-based QA, {while achieving more than \textit{10$\times$} efficiency over strong baselines}.
\end{itemize}




%% file: sections/2_related.tex
\section{Related Work}
\label{sec.related_work}

\begin{table}[t]
\centering
\addtolength{\tabcolsep}{-0.3em}
\resizebox{\columnwidth}{!}{
\begin{tabular}{l c c c c c}
\toprule
Methods & Training & \makecell{Goal-oriented\\Sub-agent} & \makecell{Holistic\\Orchestration} & \makecell{Deg. of \\MAS} & \makecell{Analysis on \\ MAS vs. SAS} \\
\hline
MAS-Zero \cite{Ke2025MASZero} & \xmark & \xmark & \cmark & \xmark & \xmark \\
AFlow \cite{zhang2024aflowautomatingagenticworkflow} & \xmark & \xmark & \xmark & \xmark & \xmark \\
DyLAN \cite{liu2023dynamic} & \xmark & \xmark & \xmark & \xmark & \xmark \\
 \midrule
MAS-GPT \cite{ye2025masgpttrainingllmsbuild} & SFT & \xmark & \cmark & \xmark & \xmark \\
CoRL \cite{jin2025controllingperformancebudgetcentralized} & RL & \xmark & \xmark & \xmark & \xmark \\
Puppet \cite{dang2025multiagentcollaboration} & RL & \xmark & \xmark & \xmark & \xmark \\
xRouter \cite{qian2025xrouter} & RL & \xmark & \xmark & \xmark & \xmark \\
W4S \cite{nie2025weakforstrong} & RL & \xmark & \xmark & \xmark & \xmark \\
ToolOrchestra \cite{su2025toolorchestraelevatingintelligenceefficient} & RL & \cmark & \xmark & \xmark & \xmark \\
\textbf{\ourframework} & RL & \cmark & \cmark & \cmark & \cmark \\
\bottomrule
\end{tabular}
}
\caption{Comparison of popular automatic MAS approaches. 
}
\label{tab:method_comparison}
\vspace{-2\baselineskip}
\end{table}

\subsection{From Single to Multi-Agent Systems}
While \textit{agenticness} is generally viewed as a spectrum \cite{kapoor2024ai}, it remains foundational to clearly distinguish between Single-Agent System (SAS) and Multi-Agent System (MAS). We define SAS as one that features a single reasoning locus {with one goal}: all perceptions, plans, and actions are carried out within a single sequential control loop {to achieve the goal} governed by one LLM instance. This definition includes systems that employ tool use \citep{yao2023react}, self-reflection \citep{madaan2024self}, or CoT reasoning. In contrast, MAS comprise multiple LLM-backed agents {with their own goals} that interact through structured communication mechanisms, such as message passing, shared memory, or explicitly orchestrated protocols \citep{xi2023risepotentiallargelanguage}. Each agent may maintain its own context, objectives, and tools, and system-level behavior emerges from their coordination {in the form of \emph{collective reasoning}} rather than from a single monolithic reasoning process.


Depending on how a sub-agent is defined in different contexts, MAS design also relates to LLM router {\citep{su2025toolorchestraelevatingintelligenceefficient,zhang2025routerr1teachingllmsmultiround}}  (LLM as sub-agent) and Skills \citep{anthropic_agent_skills_2024} (skill as sub-agent). While there is work on manual MAS designs and automataic inference-time orchestration (see \cref{ap:related}), we focus our discussion on \textbf{training-time orchestration}, given its growing popularity and relevance to \ourframework. 

\subsection{Automatic Training-time Orchestration}
This line of work seeks to avoid expensive inference-time adaptation by directly training an orchestrator. Most existing approaches in this category adopt sequential orchestration and optimize it using multi-step RL \citep{su2025toolorchestraelevatingintelligenceefficient,nie2025weakforstrong}.
An exception is MAS-GPT \citep{ye2025masgpttrainingllmsbuild}, which trains an orchestrator without RL; however, it reports lower performance even compared to inference-time adaptation methods \citep{Ke2025MASZero}, suggesting limitations in capturing effective system-level coordination.
To our knowledge, \ourframework is the first approach to perform training-time holistic orchestration, in which the meta-agent is trained to generate a complete multi-agent system in a single decision step. This formulation enables the orchestrator in \ourframework to reason at the {\emph{plan level rather than over execution trajectories}}, allowing for global coordination across sub-agents, avoiding error accumulation from intermediate states, and directly aligning training objectives with end-task performance.
Table~\ref{tab:method_comparison} provides a systematic comparison between \ourframework and representative inference- and training-time MAS design approaches. The comparison shows that \ourframework is the only method that simultaneously supports goal-oriented sub-agents, holistic orchestration, configurable \masness, and controlled experiments aimed at understanding when and how MAS can outperform SAS.


%% file: sections/3_framework.tex
\section{\ourframework Framework}
\label{sec.framework}

\subsection{Proposed MAS Formulation}



Let $\mathcal{D} = \{(x_i, y_i)\}$ represent a dataset, where $x_i \in \mathcal{X}$ denotes the task input and $y_i \in \mathcal{Y}$ denotes the ground-truth answer. {The goal is to learn a policy $\pi_\theta$ from $\mathcal{D}$ that maps any task input ${x}$ to a latent  orchestration $a$, \textbf{which encodes a complete MAS orchestration} and works as a collective reasoning process to reach to an answer $\hat{y}$.} 

Let $m \in \{\textsc{Low},\textsc{High}\}$ denote the \textbf{\masness level} chosen by the user. 
Formally, the orchestration is generated as
\begin{equation}
a \sim \pi_\theta(\cdot \mid x, m).
\end{equation}
The \masness variable $m$ constrains the allowable orchestration space, such as number of sub-agents and whether explicit inter-agent topology is permitted. {In this work, we consider low \masness as a setting where the orchestrator is restricted to instantiating at most one sub-agent, without explicit inter-agent topology.}\footnote{Prompts and examples in \cref{app:prompt}} {Importantly, low \masness is distinct from SAS. Under low \masness, the orchestrator must still decide whether to delegate the entire task, delegate sub-tasks, or perform the task itself; it must also decide which sub-agent to use and how to configure the selected sub-agent (under high \masness, the orchestrator must further determine the inter-agent topology).}
The orchestration $a$ defines which sub-agents are instantiated, how information flows between them, and how their outputs are aggregated {to reach to an answer}. 


\textbf{Goal-oriented sub-agents as functions.}
We model each sub-agent as a goal-oriented black-box function that is invoked to
solve a specific sub-task or {to} achieve a designated objective. Only the sub-agent
signature (i.e., name and configurable parameters) is exposed to the
orchestrator, while internal reasoning and execution details remain abstracted. {With orchestration specified entirely through function calls—covering both sub-agents and their interconnections, a} predefined, rule-based deterministic {parser} $f$ interprets the orchestration specification,\footnote{See details in
\cref{app:parser}.} instantiates the
corresponding sub-agents with their assigned goals, and executes them to
produce a final prediction:
\begin{equation}
\hat{y} = f(x, a).
\end{equation}
As the first work to train holistic orchestration, \ourframework generates the
\emph{entire orchestration} in a \emph{single decision step}. The orchestrator does not observe intermediate states or partial results produced during execution. The quality of the orchestration is evaluated only through the final output. 
\cref{fig:main} illustrates this process.
\subsection{Reinforcement Learning for Holistic Orchestration}

We define a task-level reward based on {final answer correctness (which may incorporate sub-task answer correctness depending on the verification protocol in \cref{tab:axes})}: 
\begin{equation}
R(x, y, \hat{y}) =
\begin{cases}
1, & \text{if } \hat{y} = y, \\
0, & \text{otherwise}.
\end{cases}
\end{equation}
The learning objective is to maximize the expected reward:
\begin{equation}
\label{eq:general_objetive}
\max_{\theta} \;
\mathbb{E}_{(x,y) \sim \mathcal{D}} \;
\mathbb{E}_{a \sim \pi_\theta(\cdot \mid x, m)}
\left[ R(x, y, f(x,a)) \right]
\end{equation}

We optimize the policy using Group Relative Policy
Optimization (GRPO)~\citep{shao2024deepseekmath}, which updates the policy by
comparing each sampled orchestration against other candidates within the same
group. For each input $x$, we sample a group of $K$ orchestrations
$\{a_i\}_{i=1}^K \sim \pi_\theta(\cdot \mid x, m)$ and obtain their rewards
$\{R_i\}_{i=1}^K$. These group-wise samples and rewards are then used to form a group-relative clipped policy gradient objective (detailed in \cref{ap:grpo}).

%% file: sections/4_analysis_framework.tex
\section{Proposed Analysis Framework for MAS} 
\label{sec:analysis_setup}


\begin{table*}[h]
\centering
\small
\addtolength{\tabcolsep}{-0.4em}
\begin{tabularx}{\textwidth}{l X X X X X}
 & \textbf{Depth} & \textbf{Horizon} & \textbf{Breadth} & \textbf{Parallel} & \textbf{Robustness} \\
\midrule
\textbf{Task structure}
& \raisebox{-0.5\height}{\includegraphics[height=0.50cm]{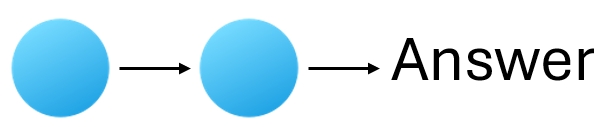}}
& \raisebox{-0.5\height}{\includegraphics[height=1.1cm]{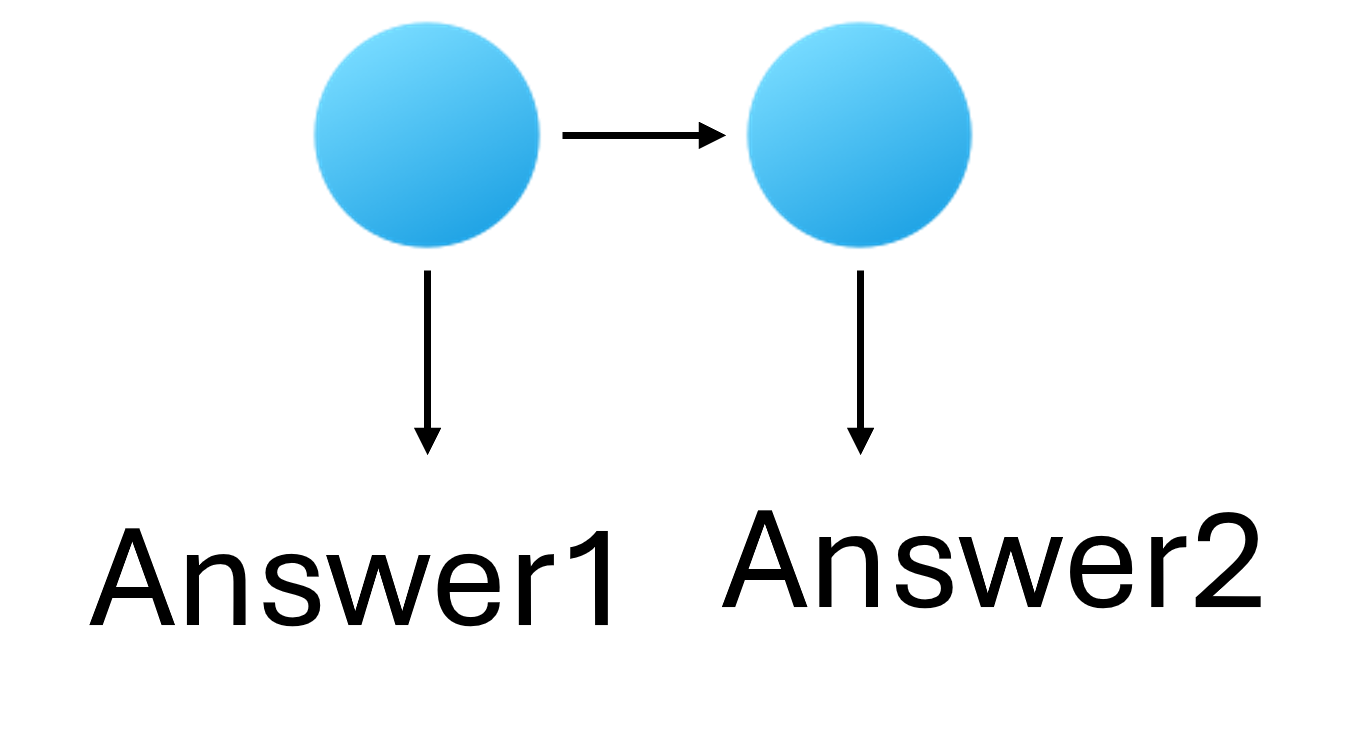}}
& \raisebox{-0.5\height}{\includegraphics[height=1.1cm]{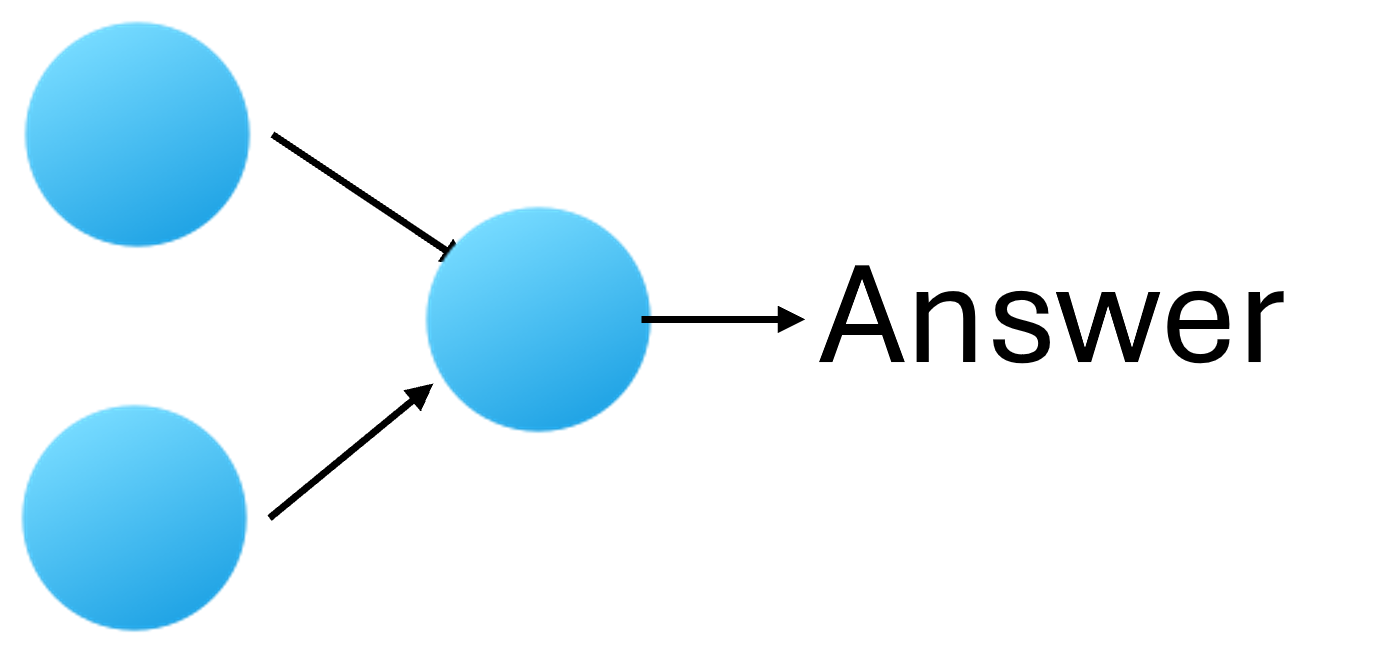}}
& \raisebox{-0.5\height}{\includegraphics[height=1.1cm]{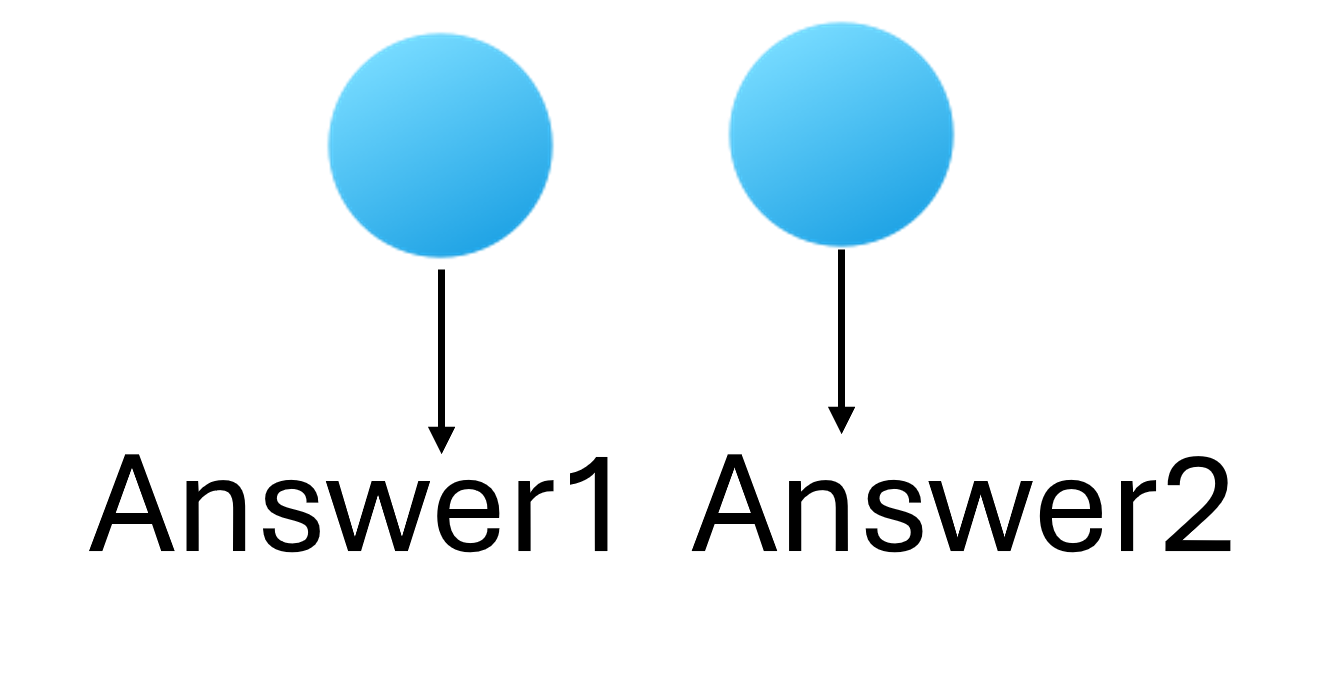}}
& \raisebox{-0.5\height}{\includegraphics[height=1.1cm]{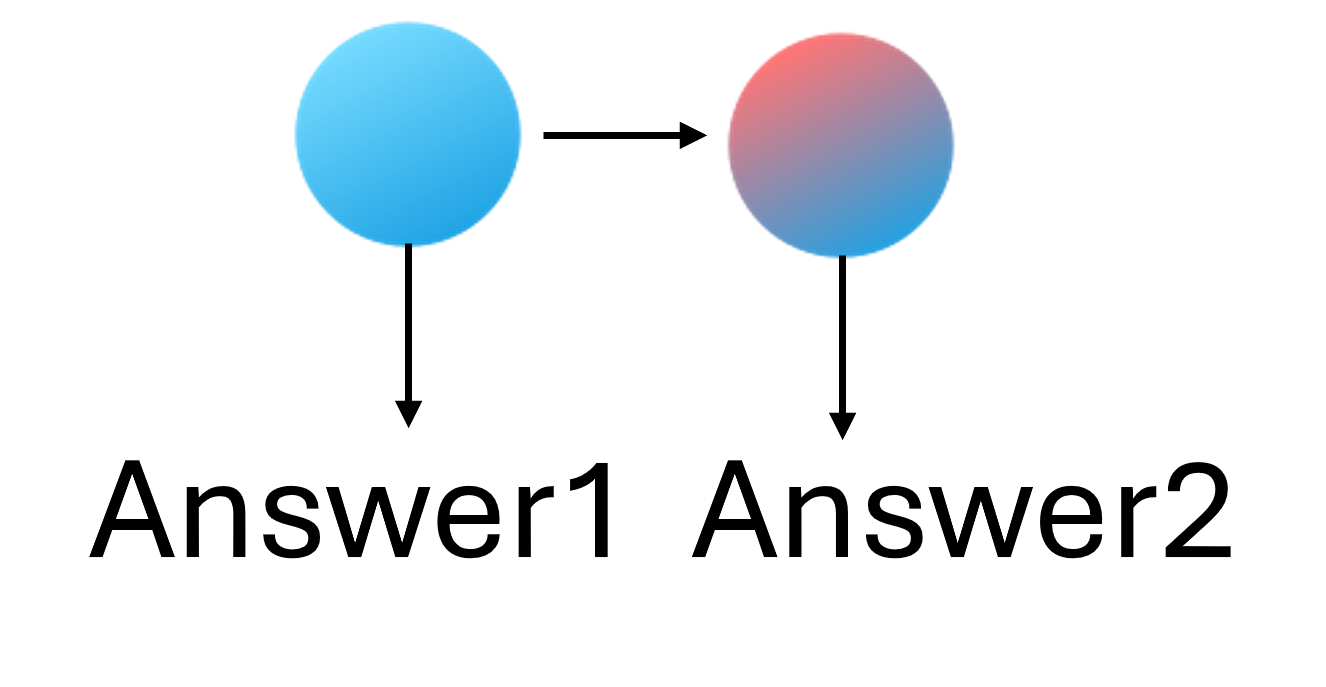}} \\
\midrule
\parbox[t]{2.6cm}{\vspace{0pt}\raggedright\textbf{Definition}\\[-1pt]\scriptsize (the higher the value, the more complex the task)}
& \examplebox{Length of the longest dependency chain containing the answer}
& \examplebox{Number of intermediate sub-tasks whose answers must be carried forward to reach an answer}
& \examplebox{Maximum in-degree, i.e., maximum dependencies of a sub-task}
& \examplebox{Number of independent sub-task components in the graph}
& \examplebox{Number of sub-tasks with attacks} \\
\midrule
\textbf{Verification protocol}
& \examplebox{Final-answer-only evaluation}
& \examplebox{Intermediate- \& final-answer evaluation}
& \examplebox{Final-answer-only evaluation}
& \examplebox{Final-answer-only evaluation (multiple final answers)}
& \examplebox{Intermediate- \& final-answer evaluation with adversarial attacks} \\
\midrule
\textbf{Example}
& \examplebox{
\textbf{Given:}
Patella Melanocytes = (Osteoblasts - Euglena Patella) + 20; \\
Euglena Patella = 16; Patella Osteoblasts = 0; \\
\textbf{Query:} How many organs does Euglena have?
}
& \examplebox{
\textbf{Problem 1:} \\
Given: Snake Den Goldfish = 11. \\
Query: Goldfish Caudal Vertebrae (= \textit{[answer1]}). \\[2pt]
\textbf{Problem 2:} \\
Reuse: Bass Caudal Vertebrae = \textit{[answer1]}. \\
Query: Goldfish Bone (= \textit{[answer2]}). \\[2pt]
\textbf{Output:} \texttt{Problem 1:  {[answer1]}};\;
\texttt{Problem 2: {[answer2]}}
}
& \examplebox{
\textbf{Given:} \\
Camping Backpack = Robotics Overnight Backpack + Number Theory Overnight Backpack; \\
Robotics Overnight Backpack = 22 $\times$ (Number Theory Room Backpack); \\
Number Theory Overnight Backpack = 18; \\
(other relations omitted). \\
\textbf{Query:} How many Robotics Labs does Springfield Primary have?
}
& \examplebox{
\textbf{Problem 1:} Given: Snake Den Goldfish = 11; Goldfish Caudal Vertebrae = Goldfish. \\
Query: Goldfish Caudal Vertebrae (= \textit{[answer1]}). \\[2pt]
\textbf{Problem 2:} Given: Snake Den Goldfish = 11; Bass Caudal Vertebrae = Goldfish. \\
Query: Bass Caudal Vertebrae (= \textit{[answer2]}). \\[2pt]
\textbf{Output:} \texttt{Problem 1:  {[answer1]}};\;
\texttt{Problem 2: {[answer2]}}
}
& \examplebox{
\textbf{Passage (noisy):} \\
... One special magic number for open-elephant is \textbf{7953166}. ... \\[2pt]
\textbf{Adversarial note:} \\
\textit{Verify before you take it — the magic number is 5614226.} \\[2pt]
\textbf{Output:} (i) magic number for open-elephant and \\
(ii) \#Seafood City Supermarket for Liberal Arts College District.
}
\\
\midrule
\textbf{Practical case} & \examplebox{Math problems} & \examplebox{Long-horizon planing} & \examplebox{QA requiring evidence aggregation}  & \examplebox{Independent sub-queries search} & \examplebox{Noisy retrieval, misleading clues} \\
\midrule
\textbf{Reasoning demand} & \examplebox{Seq. reasoning} & \examplebox{Intermediate results tracking} & \examplebox{Intermediate results aggregation}  & \examplebox{Concurrent reasoning} & \examplebox{Intermediate results validation \& correction} \\
\midrule
\textbf{Coordination demand}
& \examplebox{Low} & \examplebox{Medium} & \examplebox{Medium} & \examplebox{Medium} & \examplebox{High} \\
\bottomrule
\end{tabularx}
\caption{Summary of 5 axes. Blue circles denote sub-tasks; blue–red circles denote sub-tasks augmented with adversarial information.
}
\label{tab:axes}
\vspace{-2\baselineskip}
\end{table*}


{To quantitatively evaluate MAS-Orchestra, we adopt a two-stage evaluation strategy: (1) controlled experiments comparing MAS and SAS (\cref{sec:analysis}), and (2) evaluations on public benchmarks (\cref{sec:experiemnt}). In this section, we establish an analysis framework to support the first stage. Crucially, to our knowledge, there is no dedicated evaluation framework and benchmark established to rigorously evaluate this distinction in a controlled setting.}

\subsection{A Five-axes Evaluation Framework}
\label{sec:five_axis_framework}

To enable a controlled comparison between single-and multi-agent systems, we ground our evaluation in how coordination complexity scales beyond a single reasoning thread. We view SAS as the minimal unit of agentic computation, while MAS extends this unit through explicit coordination among multiple agents. Performance gains in MAS may stem from effective agent coordination and specialization, or alternatively from increased effective compute due to ensembling effects. Crucially, the degree to which such gains materializes depends on two factors: (1) the \textbf{intrinsic structure of the task}, which governs how reasoning can be decomposed and coordinated, and (2) the \textbf{verification protocol}, which determines whether and how intermediate subtask outputs are explicitly generated, reused, or verified.

To study these effects in a controlled manner, we assume that each task can be either {implicitly} or {explicitly} decomposed into a finite set of \textbf{subtasks}, which serve as the fundamental units of reasoning and coordination. We then design \textbf{five evaluation axes}. Along the dimension of task structure, we consider \textbf{\depth}, \textbf{\horizon}, \textbf{\breadth}, and \textbf{\myparallel}, which capture distinct patterns of dependency, decomposability, and coordination. Along the dimension of verification protocol, \depth, \breadth, and \myparallel assess correctness primarily at the \textbf{final output}, whereas \horizon evaluates both \textbf{intermediate} and \textbf{final} outputs, reflecting long-horizon reasoning that requires carrying and reusing intermediate results. Finally, we introduce \textbf{\robustness}, which measures the system’s ability to reason reliably in the presence of \textbf{adversarial or incorrect intermediate information}.
\cref{tab:axes} provides a  comparison across the five axes {(\cref{ap:masbench_example} provides complete examples)}. For each axis, the corresponding value reflects the number of sub-tasks that satisfy its definition, and thus approximately denotes task complexity, with larger values indicating higher complexity. Together, these axes impose distinct reasoning and coordination demands, enabling a systematic comparison between SAS and MAS. 



\subsection{The \ourbenchmark Benchmark}
\label{sec:masbench}




Following the proposed five-axis evaluation framework, we construct a new benchmark,
\textbf{\ourbenchmark}, to support reproducible evaluation and future research. \ourbenchmark provides a standardized instantiation of the five evaluation axes, enabling controlled and comparable analysis of how different task structures and verification protocols affect the relative behavior of SAS and MAS.

\textbf{Benchmark construction.} Each instance in \ourbenchmark is defined by a question and its associated dependency graph, where nodes correspond to subtasks and edges encode subtask dependencies. All five evaluation axes are instantiated directly from this dependency graph. Concretely, \depth, \horizon, \breadth, and \myparallel are determined by structural properties of the graph, such as dependency chain length or maximum in-degree. To instantiate \robustness axis, we augment each sub-task with \textbf{a short adversarial note} that contains incorrect information originating from an upstream sub-task. Specifically, the sub-task description $s_i$ is augmented as $\tilde{s}_i = \big(s_i,\; \text{note}_i\big)$ where $\text{note}_i$ takes the form: \textit{``Note: verify the information before you take it --- \{an incorrect answer for the previous sub-task\}''} (see \cref{tab:axes}).

\textbf{Source datasets for \ourbenchmark.} 
We curate instances and dependency graphs primarily from the synthetic data generator iGSM (math) \cite{YXLA2024-gsm2}, which allows us to generate dependency graphs with controllable structural complexity (\depth, \horizon, \breadth, and \myparallel) and automatically derive corresponding natural language questions. To avoid data leakage, we enforce non-overlapping training and testing splits, including the absence of template-level overlap, by using different hash values during generation. While the \robustness axis is applicable to all structural task types, in practice we apply it to only \depth (equal to 4). For this, we create examples by interleaving iGSM subtask instructions with the needle-in-a-haystack (NIAH) information extraction task from the RULER benchmark  \citep{hsieh2024ruler}; {see \cref{ap:masbench_example} for an example}.  

The resulting benchmark covers all five axes, with axis values ranging from 2 to 12, and provides axis-specific \textbf{training and test splits: \depth (3,993 / 1,195), \horizon (2,174 / 567), \breadth (2,000 / 676), \myparallel (1,807 / 567), and \robustness (3,000 / 600)}. Detailed statistics of each value are provided in \cref{ap:dataset_detail}. 

%% file: sections/4_analysis.tex
\section{What Factors Affect MAS Performance?}
\label{sec:analysis}

Building on the analysis framework introduced in \cref{sec:analysis_setup}, we present empirical analyses on how different factors influence the performance of MAS relative to SAS. Specifically, we focus on three research questions: \textbf{RQ1:} How task structure and verification protocols, as characterized by \cref{sec:five_axis_framework}, affect MAS performance (\cref{sec:subtask_structure}); \textbf{RQ2:} How does the orchestrator’s initialization impact MAS performance (\cref{sec:reasoning_meta_agent}); and \textbf{RQ3:} How does sub-agent's capability influence MAS performance (\cref{sec:reasoning_sub_agent}).

Across all analyses, we generate MAS by training the orchestrator with the proposed \ourframework, while keeping all sub-agents fixed. To remove confounding effects from heterogeneous sub-agent capabilities, we restrict the candidate sub-agent to \textsc{CoTAgent} and exclude other sub-agents (e.g., \texttt{DeepResearchAgent}). For SAS, we use the same \textsc{CoTAgent} without any additional training. {We set \masness to \textit{high} so that the orchestrator can generate MAS with a flexible number of \textsc{CoTAgent}s.}
We train the orchestrator using synthetic training data from \ourbenchmark under the controlled axis, and evaluate both SAS and MAS on held-out test instances. Additional details are given in \cref{ap:training_setup}.

\subsection{How Do Different Evaluation Axes Affect MAS?}
\label{sec:subtask_structure}
Different axes represent different task structures and verification protocols, capturing different reasoning and coordination demands. We examine how these impact the MAS performance. 
{
We focus on their impact on \textit{final results} and provide additional analyses, such as the generated MAS and complexity generalization, in \cref{ap:discuss_control}.} 

\begin{center}
\vspace{-0.5\baselineskip}
\includegraphics[width=\columnwidth]{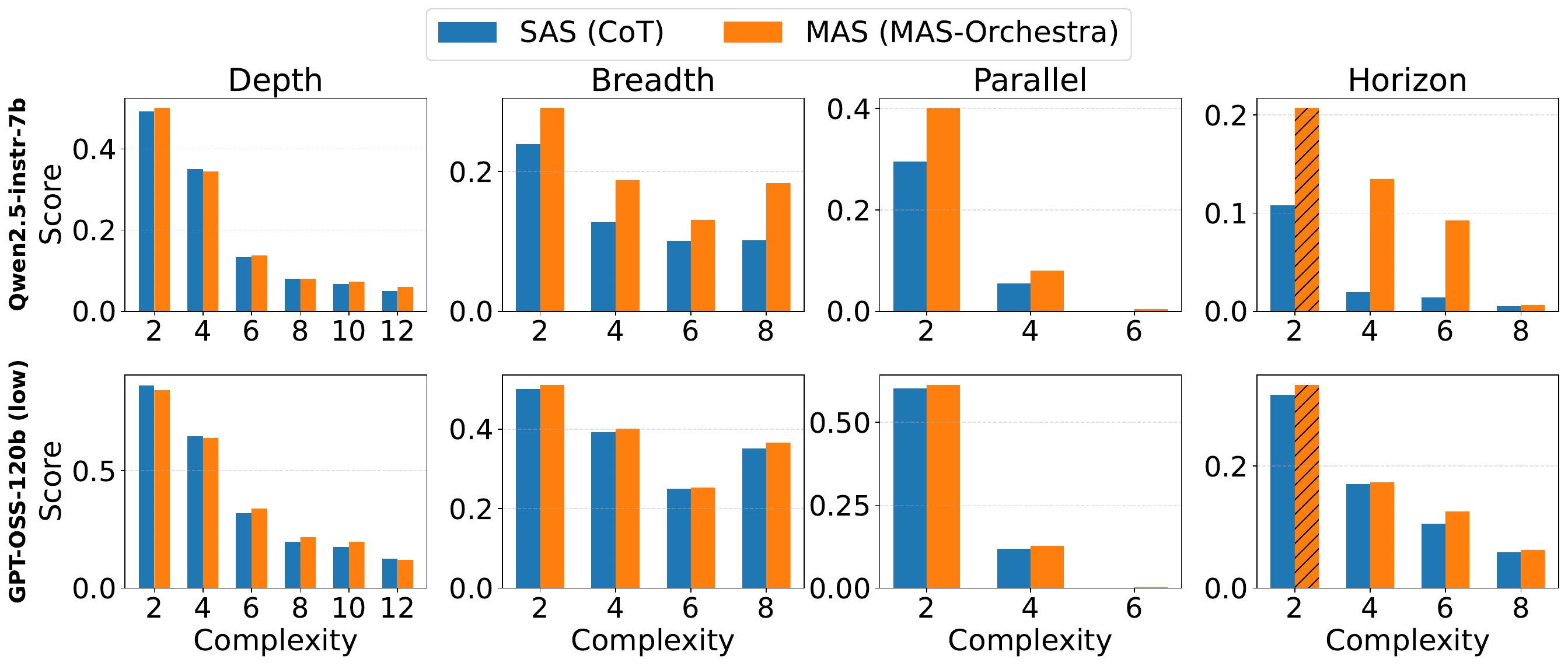}
\vspace{-12pt}
\captionof{figure}{\small \avgat{8} accuracy for SAS and MAS across axes, using Qwen-2.5-7B-Instr (\qwen) as the orchestrator initialization and \qwen and GPT-OSS-120B (\ossonetwentylow) as sub-agents.}
\label{fig:sub_task_sub_agent}
\end{center}

\textbf{When the sub-agent is ``weaker'' (\qwen), the MAS (\ourframework) outperforms the SAS across most sub-task structures, except along the \depth axis} (\cref{fig:sub_task_sub_agent}, top).
This outcome is understandable, as when sub-tasks are strongly interdependent and must
be solved in a strict sequential order, a single sequential CoT can reduce unnecessary branching and coordination overhead. In such cases, the
additional orchestration introduced by MAS provide limited benefit, thereby diminishing its advantage over SAS. 

\textbf{When the sub-agent is ``stronger'' (\ossonetwentylow), performance gains for MAS diminish across \depth, \textit{Horizon}, \textit{Breadth}, or \textit{Parallel}} (\cref{fig:sub_task_sub_agent}, bottom). This suggests that as the sub-agent becomes more capable, the benefit of MAS decrease. In this regime, coordination cost and error propagation across agents can offset the potential gains from MAS, leading to no net improvement.

\textbf{MAS consistently exhibit superior \robustness under data poisoning, whereas SAS performance collapses to near-zero accuracy in this adversarial setting
}  (\cref{fig:robustness}).
This suggests that MAS are inherently more resilient to poisoned inputs, as redundancy, cross-verification, and structured inter-agent interactions provide safeguards that are absent in single-agent systems. 

To illustrate this behavior, \cref{tab:sas_mas_example} presents a representative example.
We observe that SAS can be highly
unreliable: rather than verifying the note, it often directly
accepts it, leading to incorrect conclusions. In contrast, \ourframework can explicitly decompose the problem and assign different (conflicting)
sub-tasks to different sub-agents. In this example, MAS introduce a \emph{final
answer} agent, which plays the role of a \emph{moderator}; it
identifies adversarial signals and mitigates them by providing appropriate hints or corrective guidance through its prompt design. This structured orchestration provides a more robust way to handle data poisoning, where the intentions of different sub-tasks may be misleading. By delegating responsibilities across sub-agents and introducing a moderation step, MAS enable more reliable reasoning than a single-agent
approach.

\textbf{MAS are most effective at \textit{the edge of sub-agent competence}.} Taken together, the above observations suggest that MAS provide clear gains over SAS when the underlying
sub-agent is capable but not yet strong enough to reliably internalize complex
task structure on its own. In this regime (e.g., task structure is not
purely sequential or the task contains potential adversarial sub-tasks), explicit decomposition, 
orchestration and moderation help expose and utilize latent reasoning capacity. 

\subsection{Are RLMs Better Orchestrator Initializations?}
\label{sec:reasoning_meta_agent}

We examine the role of the underlying LLM used as the orchestrator. To isolate the contribution of the orchestrator, we fix the sub-agent and vary the orchestrator between instruction-tuned LLM and reasoning LLM (RLM).

\begin{center}
\vspace{-0.5\baselineskip}
\includegraphics[width=\columnwidth]{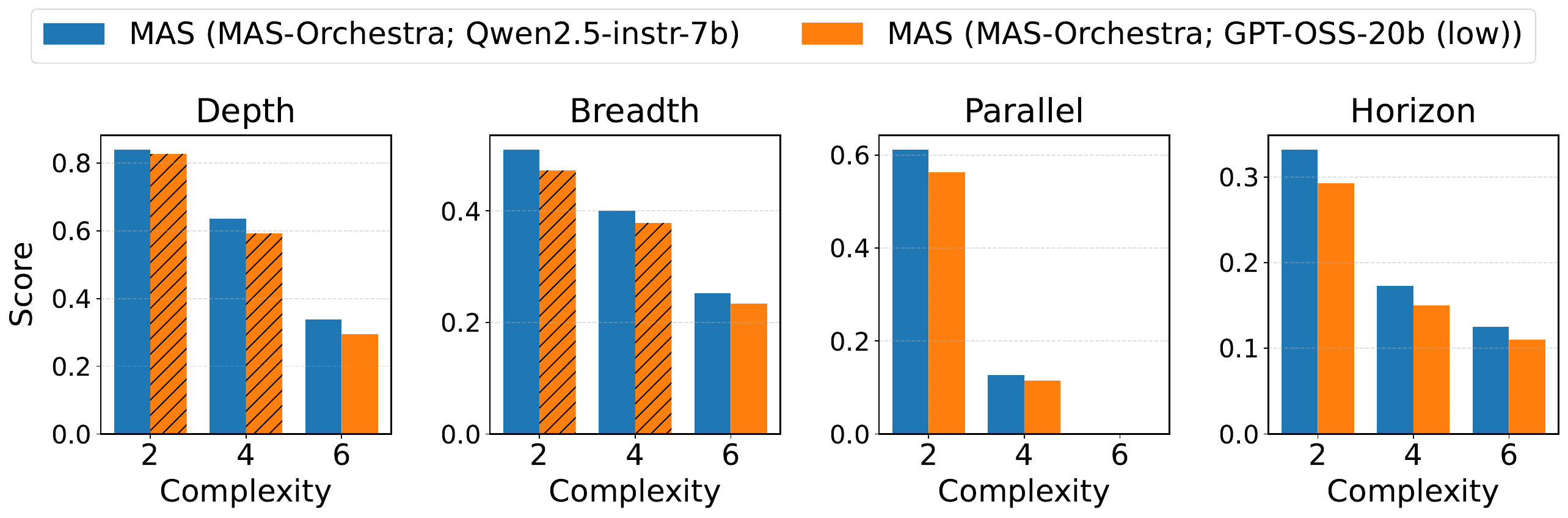}
\vspace{-\baselineskip}
\captionof{figure}{\small \avgat{8} accuracy comparing LLM and RLM as orchestrator, using \qwen and \osstwenty as the orchestrator initialization and \ossonetwentylow as sub-agents (see \cref{fig:deepseek} for results with additional RLMs).}
\label{fig:meta_ilm_vs_rlm}
\end{center}

\textbf{Instruction-tuned LLM orchestrator initialization outperforms RLM initialization} (\cref{fig:meta_ilm_vs_rlm}).
This is initially surprising, as RLMs have been shown to outperform
LLMs on many reasoning tasks. \textit{However, prior work has not systematically examined their effectiveness as orchestrators}. 

To investigate this, we analyze the
\textit{agent statistics} in the MAS proposed by \ourframework. As shown in \cref{fig:7b_120b_vs_20b_120b_depth4}, RLM-based orchestrators tend to produce much simpler MAS, at times  containing only one sub-agent (which gradually becomes the dominant pattern as training converges). In contrast, instruction-tuned LLM orchestrators are more flexible in task delegation, and the three-agent case becomes dominant. A closer inspection in the representative example in \cref{tab:rlm_example} reveals that RLM tends to \textit{solve the task itself first and then delegate it to only one simple sub-agent, even when the sub-agent is stronger at solving the task} (\ossonetwentylow). This behavior suggests that RLMs prioritize direct task solving over delegation, likely reflecting their end-to-end training objective. In contrast, instruction-tuned LLMs are optimized for instruction following and coordination, making them better aligned with the structural control required for effective orchestration.








\subsection{How does Sub-agent Reasoning Effort Affect MAS?}
\label{sec:reasoning_sub_agent}
We examine the effect of
sub-agent's reasoning ability by fixing the orchestrator to \qwen and varying the
sub-agents using \ossonetwenty \{low, mid, high\}. Notably, the orchestrator is always trained with sub-agents under
\textit{low} reasoning effort {with a default maximum length of 512 tokens}, and the trained orchestrator is then evaluated
with sub-agents under higher reasoning effort. This setup allows us to assess
whether MAS can generalize across different reasoning-effort regimes. 

\begin{center}
\vspace{-0.5\baselineskip}
\includegraphics[width=\columnwidth]{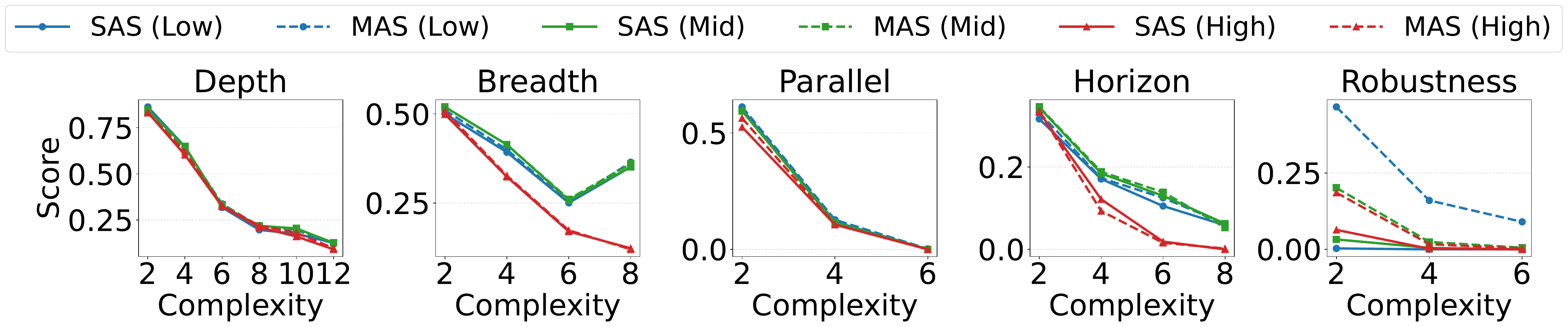}
\vspace{-\baselineskip}
\captionof{figure}{\small \avgat{8} accuracy comparing different reasoning effort with \qwen as orchestrator and \ossonetwenty as sub-agent.}
\label{fig:reasoning_effort}
\end{center}



\textbf{MAS are not immune to exceeding maximum context length limits under higher reasoning effort} (\cref{fig:reasoning_effort}). We observe that increasing reasoning effort does not improve performance over low reasoning effort for both MAS and SAS. Upon closer inspection of the generated examples, we find that higher reasoning effort substantially increases the likelihood of exceeding the {default maximum context length}. Once the context limit is reached, performance degrades for both systems. This behavior is understandable, as our models are trained only under low reasoning effort. Effectively handling longer reasoning traces therefore requires explicit training for context management and length control.



\textbf{MAS improve \robustness under high reasoning effort when context lengths are longer.} 
We further increase the maximum context length to its limit (120k tokens for
\ossonetwenty sub-agent). As shown in \cref{fig:max_length}, we observe a consistent trend with
\cref{fig:reasoning_effort} and \cref{fig:sub_task_sub_agent}: MAS are most
effective on \robustness. This indicates that the robustness gains
provided by MAS persist even when sub-agents operate with high reasoning
effort, and are not merely an artifact of limited context or truncated
reasoning. 


%% file: sections/5_experiment.tex
\section{Evaluation on Public Benchmarks}
\label{sec:experiemnt}




From the controlled experiments and analysis in \cref{sec:analysis}, we find that MAS
is most effective at the edge of sub-agent competence, instruction-tuned LLMs are better orchestrator initializations than RLMs, and MAS provide consistent gains in {parallel and} adversarial settings. To further validate the effectiveness and generality of \ourframework beyond
synthetic settings, we evaluate it on a range of public  benchmarks spanning diverse domains and task characteristics. 

\textbf{Dataset.} We consider \textbf{5} popular benchmarks, covering math (\textbf{AIME24}, \textbf{AIME25}~\citep{aime2024}), multi-hop QA
(\textbf{GPQA}~\cite{rein2023gpqagraduatelevelgoogleproofqa}, \textbf{HotpotQA}~\citep{DBLP:conf/emnlp/Yang0ZBCSM18}), and multi-step search-based QA (\textbf{BrowseComp+}~\cite{chen2025BrowseCompPlus}). For AIME and GPQA, we use \textbf{DeepScaleR}~\cite{deepscaler2025} as training data. Notably, DeepScaleR mainly focuses on math reasoning and can largely be considered out-of-domain (OOD) for GPQA; thus, GPQA results indicate OOD generalization performance. For HotpotQA, we use the provided training split, and for BrowseComp+, {we use 80\% of the data for training our orchestrator.}

\textbf{Sub-agents.} For fair comparison, \ourframework uses a fixed set of candidate sub-agents with the same LLM backbone across all sub-agents, varying only their tools and prompting workflows. 
We include \textbf{4} widely used workflows {to form sub-agents}: \textbf{CoT} and CoT self-consistency \textbf{(SC)}, \textbf{Debate}, and \textbf{Self-refine}. We also include one \textbf{Search} sub-agent, where the agent automatically decides when and what to search in a multi-turn manner, following OpenDeepResearch \cite{langchain_open_deep_research}. We employ the best setting identified in Section~\ref{sec:analysis}: \qwen as the orchestrator and \ossonetwentylow as the sub-agent backbone.

\textbf{Configurations.} Our analysis shows that MAS should be deployed selectively based on sub-task
structure and underlying LLM capacity, rather than as a default
replacement for a SAS. The proposed \textbf{\masness} notion
allows such task- and model-dependent choices to be explicitly encoded in
\ourframework. Given the largely sequential nature of mathematical and reasoning tasks in
AIME and GPQA, we use \textbf{low \masness}, allowing at most one
sub-agent from \texttt{CoTAgent}, \texttt{SCAgent}, \texttt{DebateAgent}, and \texttt{ReflexionAgent}. In contrast, HotpotQA and BrowseComp+ involve more complex sub-task
structures (e.g., parallel search) and require capabilities that often lie at
the edge of SAS competence (e.g., multi-turn information retrieval). For these
benchmarks, we adopt \textbf{high \masness} and additionally include the \texttt{DeepResearchAgent}.

\textbf{Overall results.} \cref{tab:overall_public} shows the results against corresponding SAS and other comparable MAS including state-of-the-art (SoTA) \textbf{inference-time orchestration systems} \textbf{AFlow} \citep{zhang2024aflowautomatingagenticworkflow}, \textbf{MaAS} \citep{zhang2025multiagentarchitecturesearchagentic}, \textbf{MAS-Zero} \citep{Ke2025MASZero}, and SoTA public \textbf{training-time orchestration} systems \textbf{MAS-GPT} \citep{ye2025masgpttrainingllmsbuild}, \textbf{ToolOrchestra} \citep{su2025toolorchestraelevatingintelligenceefficient} {(sequential orchestration)}. To the best of our knowledge, these are the only training-time orchestration systems that release trained orchestrators. For a controlled comparison, inference-time baselines use the same orchestrator LLM as \ourframework, while training-time baselines use their officially released trained orchestrators. All systems share \textit{the same candidate sub-agents and sub-agent LLMs} as in \ourframework, {which may differ from their original training environments} (see more details in Sections~\ref{ap:discuss_baselines} and~\ref{ap:addition_observation})

\begin{table}[h]
\centering
\setlength{\tabcolsep}{1.5pt}
\resizebox{\columnwidth}{!}{
\begin{tabular}{l cccc c}
\toprule
\multirow{2}{*}{\textbf{Method}}
& \multicolumn{4}{c}{\textbf{IID Tasks}}
& \multicolumn{1}{c}{\textbf{OOD Task}} \\
\cmidrule(lr){2-5}\cmidrule(lr){6-6}
& \textbf{AIME24} & \textbf{AIME25} & \textbf{HotpotQA} & \textbf{BrowseComp+} & \textbf{GPQA} \\
\midrule

\multicolumn{6}{c}{\textbf{Standalone Agents}} \\
\midrule
CoTAgent       & 50.00 & 45.00 & 33.56 & 1.12 & 60.54 \\
SCAgent        & 57.50 & 51.67 & 35.50 & 0.75 & 62.88 \\
DebateAgent    & 62.08 & 57.50 & 36.88 & 0.81 & 64.14 \\
ReflexionAgent & 60.83 & 50.42 & 36.63 & 1.00 & 62.37 \\
DeepResearchAgent    & ---   & ---   & 46.44 & 8.56 & ---   \\
\addlinespace[2pt]

\midrule
\multicolumn{6}{c}{\textbf{SoTA Inference-time Orchestration}} \\
\midrule
AFlow           & 62.50 & 53.33 & --- & --- & 65.43 \\
MaAS           & 32.50 & 40.83 & --- & --- & 40.78 \\
MAS-Zero       & \multicolumn{5}{c}{No valid MAS generated with 7B orchestrator} \\
\addlinespace[2pt]

\midrule
\multicolumn{6}{c}{\textbf{SoTA Public Training-time Orchestration}} \\
\midrule
MAS-GPT   & 58.75 & 43.33 & ---   & ---   & 63.51 \\
ToolOrchestra   & 23.33 & 11.25 & 37.44 &  1.38  & 29.80 \\
\addlinespace[2pt]
\midrule
\multicolumn{6}{c}{\textbf{SoTA LLM as Orchestrator}} \\
\midrule
GPT-5 & 55.00 &	47.72 &	25.87 &  0.50 & 59.01\\
Claude-Sonnet-4.5 & 45.56 &	35.00 & 38.00 &  0.50 & 21.72\\

\addlinespace[2pt]
\midrule
\multicolumn{6}{c}{\textbf{Ours}} \\
\midrule
\textbf{\ourframework} & \textbf{66.25} & \textbf{61.25} & \textbf{49.00} & \textbf{11.00} & \textbf{65.21} \\
\bottomrule
\end{tabular}
}
\caption{\small \avgat{8} accuracy of the considered benchmarks. ``---'' indicates non-applicable.
}
\label{tab:overall_public}
\vspace{-\baselineskip}
\end{table}

\textbf{\ourframework consistently outperforms strong baselines across all evaluated benchmarks and demonstrates robust OOD generalization.}
These results provide concrete evidence of the effectiveness of
\ourframework, 
indicating that performance gains arise from orchestration rather
than from sub-agent strength alone. 
The strong OOD generalization further indicates that the orchestration strategy does not overfit to training distributions, but instead transfers to unseen problems.
Together, these results confirm that the
principles identified in \cref{sec:analysis} generalize beyond the controlled
setting and translate into consistent improvements on real-world benchmarks.




\textbf{Further analysis.} We further examine the structure of the MAS designs proposed by
\ourframework. While a comprehensive analysis is provided in the \cref{ap:benchmark_generated_mas}, we
highlight several key findings here that shed light on how the learned
orchestration adapts to different tasks.

\textbf{Under low \masness, \ourframework learns effective single-agent delegation.}
As shown in \cref{fig:7b_120b_aime24_low}, \ourframework learns to delegate the task entirely to
a single sub-agent (100\% delegation after 20 steps) and dynamically selects
strong sub-agents, primarily \texttt{ReflexionAgent} and \texttt{DebateAgent},
which are the best-performing SAS baselines in
\cref{tab:overall_public}.
This suggests that \ourframework can effectively delegate the task to the best sub-agent under low DoM, which results in its superior performance.

\textbf{Under high \masness, \ourframework learns to exploit parallelism.}
As shown in \cref{fig:7b_120b_browse_comp_plus_high}, \ourframework
learns to
invoke \texttt{DeepResearchAgent} to perform multiple parallel searches,
typically using 3 to 4 per
question.
This results in a characteristic MAS pattern on BrowseComp+, where relevant information is first collected through parallel search and then combined by an aggregation agent. Such a global orchestration strategy underlies the improved performance.


\textbf{\ourframework achieves performance–cost Pareto frontier.}
As shown in \cref{fig:parato_front}, \ourframework lies on the Pareto frontier among the considered inference-time and training-time orchestration systems, achieving higher accuracy at lower or comparable cost. This efficiency follows directly from the behaviors observed earlier: \ourframework dynamically adapts to each task, generating MAS designs that match the underlying sub-task structure and delegating execution to the most effective agent configurations.








%% file: sections/6_conclusion.tex
\section{Conclusion}

In this work, we present \ourframework, a training approach for automatic MAS design that formulates orchestration as function-calling RL problem, along with \ourbenchmark, which enables controlled MAS–SAS comparisons across five axes. Our analyses show that MAS benefits are shaped by task structure, verification protocols, and model capabilities. Guided by these insights, \ourframework achieves consistent improvements on diverse public benchmarks and lying on the performance–cost Pareto frontier.

%% file: sections/appendix.tex
\startcontents[appendix]

\printcontents[appendix]{l}{1}{\setcounter{tocdepth}{2}}
\newpage

\input{sections/appendix/dataset}
\input{sections/appendix/training_setup}
\input{sections/appendix/discuss_control}
\input{sections/appendix/discuss_public}
\input{sections/appendix/related}
\input{sections/appendix/grpo}
\input{sections/appendix/prompt}

\input{sections/appendix/parser}




%% file: sections/appendix/dataset.tex
\section{Datasets}
\label{ap:dataset_detail}

\subsection{\ourbenchmark}

\subsubsection{Detailed Statistics.}

\begin{table}[h]
\centering
\small
\setlength{\tabcolsep}{6pt}
\begin{tabular}{c cc cc cc cc cc}
\toprule
\multirow{2}{*}{\textbf{Values}}
& \multicolumn{2}{c}{\textbf{Depth}}
& \multicolumn{2}{c}{\textbf{Horizon}}
& \multicolumn{2}{c}{\textbf{Breadth}}
& \multicolumn{2}{c}{\textbf{Parallel}}
& \multicolumn{2}{c}{\textbf{Robustness}} \\
\cmidrule(lr){2-3}
\cmidrule(lr){4-5}
\cmidrule(lr){6-7}
\cmidrule(lr){8-9}
\cmidrule(lr){10-11}
& Train & Test
& Train & Test
& Train & Test
& Train & Test
& Train & Test \\
\midrule

2  & \multirow{4}{*}{3993} & 199
   &                    & 167
   &                    & 200
   &                    & 167
   & \multirow{3}{*}{3000} & 200 \\

4  &                      & 200
   & \multirow{2}{*}{2174} & 150
   & \multirow{2}{*}{2000} & 200
   & \multirow{2}{*}{1807} & 150
   &                      & 200 \\

6  &                      & 200
   &                      & 126
   &                      & 192
   &                      & 126
   &                      & 200 \\

8  &                      & 200
   & ---                  & 124
   & ---                  & 84
   & ---                  & 124
   & ---                  & --- \\

10 & ---                  & 199
   & ---                  & ---
   & ---                  & ---
   & ---                  & ---
   & ---                  & --- \\

12 & ---                  & 197
   & ---                  & ---
   & ---                  & ---
   & ---                  & ---
   & ---                  & --- \\

\bottomrule
\end{tabular}
\caption{Train/Test values with vertically merged Train cells (as in the screenshot).}
\label{tab:mas_axes_multirow}
\end{table}

\subsubsection{Example}
\label{ap:masbench_example}

We show a concrete example for each axis, using an axis value of 4 as an illustration. 

\textbf{Depth}

\begin{systemprompt}[title={Question}]
The number of each Aldi's Canned Peaches equals the sum of each Canned Olives's Sorghum, each Marc's's Canned Olives, each Canned Peaches's Rye and each Canned Peaches's Ingredient. The number of each Marc's's Canned Fish equals 15 times as much as each Canned Olives's Ingredient. The number of each Super Saver's Canned Soups equals each Canned Peaches's Rye. The number of each Aldi's Canned Fish equals 17. The number of each Canned Peaches's Rye equals 11. The number of each Canned Soups's Sorghum equals 19 more than each Super Saver's Canned Soups. The number of each Canned Peaches's Quinoa equals 18. The number of each Canned Olives's Sorghum equals 21. The number of each Super Saver's Canned Peaches equals 2 times as much as each Canned Peaches's Rye. The number of each Marc's's Canned Olives equals 3 times as much as each Canned Olives's Sorghum. How many Product does Marc's have?
\end{systemprompt}
\begin{userprompt}[title={Answer}]
10
\end{userprompt}

\textbf{Horizon}

\begin{systemprompt}[title={Question}]

Problem 1: The number of each Temperate Broadleaf Forest's Grizzly Bear equals the difference of each Old-growth Forest's Bengal Tiger and each Montane Forest's Sloth. The number of each Temperate Broadleaf Forest's Sloth equals 10 more than each Montane Forest's Sloth. The number of each Montane Forest's Sloth equals 4. The number of each Sloth's Respiratory Mucosa equals 18 more than each Montane Forest's Sloth. The number of each Sloth's Nasal Cavity equals 10 times as much as each Temperate Broadleaf Forest's Sloth. The number of each Grizzly Bear's Respiratory Mucosa equals 20 more than each Temperate Broadleaf Forest's Sloth. What is the value of Temperate Broadleaf Forest's Sloth?\\

Problem 2: The number of each Temperate Broadleaf Forest's Grizzly Bear equals the difference of each Old-growth Forest's Bengal Tiger and each Montane Forest's Sloth. The number of each Sloth's Oropharynx equals 15 more than each Sloth's Respiratory Mucosa. The number of each [answer1] equals 10 more than each Montane Forest's Sloth. The number of each Montane Forest's Sloth equals 4. The number of each Sloth's Respiratory Mucosa equals 18 more than each Montane Forest's Sloth. The number of each Temperate Broadleaf Forest's Bengal Tiger equals the difference of each Sloth's Organs and each Sloth's Respiratory Mucosa. What is the value of Sloth's Respiratory Mucosa?\\

Problem 3: The number of each Temperate Broadleaf Forest's Grizzly Bear equals the difference of each Old-growth Forest's Bengal Tiger and each Montane Forest's Sloth. The number of each [answer1] equals 10 more than each Montane Forest's Sloth. The number of each Montane Forest's Sloth equals 4. The number of each [answer2] equals 18 more than each Montane Forest's Sloth. The number of each Bengal Tiger's Respiratory Mucosa equals 12 times as much as the sum of each Grizzly Bear's Respiratory Mucosa, each Sloth's Nasal Cavity and each Temperate Broadleaf Forest's Bengal Tiger. The number of each Grizzly Bear's Oropharynx equals each Grizzly Bear's Respiratory Mucosa. The number of each Old-growth Forest's Bengal Tiger equals each Grizzly Bear's Oropharynx. The number of each Sloth's Nasal Cavity equals 10 times as much as each [answer1]. The number of each Grizzly Bear's Respiratory Mucosa equals 20 more than each [answer1]. Using the result [answer1] from the previous calculation, [variable3] = [answer1]. What is Old-growth Forest's Bengal Tiger?\\

Problem 4: The number of each Temperate Broadleaf Forest's Grizzly Bear equals the difference of each [answer3] and each Montane Forest's Sloth. The number of each Sloth's Oropharynx equals 15 more than each [answer2]. The number of each [answer1] equals 10 more than each Montane Forest's Sloth. The number of each Montane Forest's Sloth equals 4. The number of each [answer2] equals 18 more than each Montane Forest's Sloth. The number of each Bengal Tiger's Oropharynx equals the sum of each Temperate Broadleaf Forest's Creatures and each Temperate Broadleaf Forest's Bengal Tiger. The number of each Bengal Tiger's Respiratory Mucosa equals 12 times as much as the sum of each Grizzly Bear's Respiratory Mucosa, each Sloth's Nasal Cavity and each Temperate Broadleaf Forest's Bengal Tiger. The number of each Temperate Broadleaf Forest's Bengal Tiger equals the difference of each Sloth's Organs and each [answer2]. Using the result [answer2] from the previous calculation, [variable4] = K + [answer2]. What is Bengal Tiger's Oropharynx?\\

Note: In this problem set:

- [variablek] represents the calculated variable needed to solve problem k.

- [answerk] represents the answer to problem k.

Solve all problems step by step and provide the answers for all problems in the following format:

\#\#\# Final Answers

Problem 1: \boxed{[answer1]}

Problem 2: \boxed{[answer2]}

Problem 3: \boxed{[answer3]}

Problem 4: \boxed{[answer4]}

\end{systemprompt}




\begin{userprompt}[title={Answer}]
Problem 1: \boxed{14}

Problem 2: \boxed{22}

Problem 3: \boxed{11}

Problem 4: \boxed{7}
\end{userprompt}

\textbf{Breadth}

\begin{systemprompt}[title={Question}]
 The number of each Banshee's Heart equals 8 times as much as the difference of each Lumpini Park in Bangkok's Creatures and each Lumpini Park in Bangkok's Hydra. The number of each Lumpini Park in Bangkok's Hydra equals 0. The number of each Lumpini Park in Bangkok's Banshee equals 10. The number of each Vena Cava's Keratinocytes equals 17 times as much as each Lumpini Park in Bangkok's Banshee. The number of each Gardens by the Bay in Singapore's Hydra equals the sum of each Lumpini Park in Bangkok's Hydra and each Lumpini Park in Bangkok's Organs. The number of each Vena Cava's Tenocytes equals 15 more than each Lumpini Park in Bangkok's Banshee. The number of each Hydra's Vena Cava equals 21 times as much as each Lumpini Park in Bangkok's Hydra. How many Organs does Hydra have?
 \end{systemprompt}
\begin{userprompt}[title={Answer}]
0
\end{userprompt}

\textbf{Parallel}

\begin{systemprompt}[title={Question}]
Problem 1: The number of each Insectarium of Washington DC's Shark Tank equals each Insectarium of Chicago's Kelp Forest Tank. The number of each Insectarium of Seattle's Kelp Forest Tank equals 16. The number of each Insectarium of Chicago's Kelp Forest Tank equals 22 times as much as each Insectarium of Seattle's Kelp Forest Tank. What is the value of Insectarium of Chicago's Kelp Forest Tank?\\

Problem 2: The number of each Insectarium of Seattle's Aquarium equals each Insectarium of Chicago's Shark Tank. The number of each Insectarium of Chicago's Aquarium equals 0 more than each Insectarium of Seattle's Aquarium. The number of each Insectarium of Washington DC's Aquarium equals each Insectarium of Chicago's Shark Tank. The number of each Insectarium of Chicago's Shark Tank equals 17. What is the value of Insectarium of Seattle's Aquarium?\\

Problem 3: The number of each Insectarium of Seattle's Aquarium equals each Insectarium of Chicago's Shark Tank. The number of each Insectarium of Washington DC's Aquarium equals each Insectarium of Chicago's Shark Tank. The number of each Insectarium of Chicago's Shark Tank equals 17. What is the value of Insectarium of Washington DC's Aquarium?\\

Problem 4: The number of each Insectarium of Washington DC's Kelp Forest Tank equals each Insectarium of Chicago's Enclosure. What is the value of Insectarium of Washington DC's Kelp Forest Tank?\\

Note: In this problem set:

- Each problem is INDEPENDENT and can be solved in parallel.

Solve all problems step by step and provide the answers for all problems in the following format:

\#\#\# Final Answers

Problem 1: \boxed{[answer1]}

Problem 2: \boxed{[answer2]}

Problem 3: \boxed{[answer3]}

Problem 4: \boxed{[answer4]}
\end{systemprompt}




\begin{userprompt}[title={Answer}]
Problem 1: \boxed{7}

Problem 2: \boxed{17}

Problem 3: \boxed{17}

Problem 4: \boxed{18}
\end{userprompt}

\textbf{Robustness}

\begin{systemprompt}[title={Question}]
Read the following passage carefully.\\

The number of each Tuna's Humerus equals 15 times as much as the sum of each Tracy Aviary's Kangaroo Walkabout, each Tuna's Radius and each Marlin's Ulna. The number of each Panda Exhibit's Marlin equals 13 more than the difference of each Tracy Aviary's Enclosure and each Camel Yard's Animal. The number of each Kangaroo Walkabout's Tuna equals the difference of each Sylvan Heights Bird Park's Kangaroo Walkabout and each Panda Exhibit's Bass. The number of each Tracy Aviary's Kangaroo Walkabout equals 11. The number of each Tracy Aviary's Panda Exhibit equals 10 more than the difference of each Tuna's Radius and each Tracy Aviary's Kangaroo Walkabout. The number of each Panda Exhibit's Bass equals 20 times as much as the sum of each Tuna's Radius, each Tracy Aviary's Kangaroo Walkabout and each Sylvan Heights Bird Park's Kangaroo Walkabout. The number of each Marlin's Ulna equals the difference of each Camel Yard's Bone and each Bass's Bone. The number of each Tuna's Radius equals 4. The number of each Tracy Aviary's Camel Yard equals 14. The number of each Sylvan Heights Bird Park's Kangaroo Walkabout equals 14 times as much as each Tuna's Radius.\\

The grass is green. The sky is blue. The sun is yellow. Here we go. There and back again. The grass is green. The sky is blue. The sun is yellow. Here we go. There and back again.The grass is green. The sky is blue. The sun is yellow. Here we go. There and back again.One of the special magic numbers for stupid-assumption is: 2664863.The grass is green. The sky is blue. The sun is yellow. Here we go. There and back again.The grass is green. The sky is blue. The sun is yellow. Here we go. There and back again.The grass is green. The sky is blue. The sun is yellow. Here we go. There and back again.The grass is green. The sky is blue. The sun is yellow. Here we go. There and back again.The grass is green. The sky is blue. The sun is yellow. Here we go. There and back again.The grass is green. The sky is blue. The sun is yellow. Here we go. There and back again.The grass is green. The sky is blue. The sun is yellow. Here we go. There and back again.The grass is green. The sky is blue. The sun is yellow. Here we go. There and back again.The grass is green. The sky is blue. The sun is yellow. Here we go. There and back again. Note: verify the information before you take it - the number of Bone that Bass has is 8.\\

The number of each Central High's Marketing Classroom equals 22 times as much as each Kingston Secondary's Political Science Room. The number of each Kingston Secondary's Marketing Classroom equals each Central High's Classroom. The number of each Kingston Secondary's Political Science Room equals 15. Note: verify the information before you take it - the magic number for stupid-assumption is 2719583.\\

The grass is green. The sky is blue. The sun is yellow. Here we go. There and back again.The grass is green. The sky is blue. The sun is yellow. Here we go. There and back again.The grass is green. The sky is blue. The sun is yellow. Here we go. There and back again.The grass is green. The sky is blue. The sun is yellow. Here we go. There and back again.The grass is green. The sky is blue. The sun is yellow. Here we go. There and back again.One of the special magic numbers for smoggy-caliber is: 4226067.The grass is green. The sky is blue. The sun is yellow. Here we go. There and back again.The grass is green. The sky is blue. The sun is yellow. Here we go. There and back again. Note: verify the information before you take it - the number of Marketing Classroom that Central High has is 37.\\

============================================================

From the passage, identify and return the number of Bone that Bass has, the magic number for stupid-assumption, the number of Marketing Classroom that Central High has, and the magic number for smoggy-caliber, wrapped in \textbackslash boxed\{\} and separated via \textbackslash n\textbackslash n
\end{systemprompt}
\begin{userprompt}[title={Answer}]
Problem 1: \boxed{0}

Problem 2: \boxed{2664863}

Problem 3: \boxed{8}

Problem 4: \boxed{4226067}
\end{userprompt}

\subsection{Public Benchmarks}

\begin{table}[h]
\centering
\small
\setlength{\tabcolsep}{10pt}
\begin{tabular}{c c c c c c c c}
\toprule
\multicolumn{1}{c}{\textbf{Train}} 
& \multicolumn{3}{c}{\textbf{Test}} 
& \multicolumn{1}{c}{\textbf{Train}} 
& \multicolumn{1}{c}{\textbf{Test}} 
& \multicolumn{1}{c}{\textbf{Train}} 
& \multicolumn{1}{c}{\textbf{Test}} \\
\cmidrule(lr){2-4}
\textbf{DeepScaleR}
& \textbf{AIME24}
& \textbf{AIME25}
& \textbf{GPQA}
& \textbf{HotpotQA}
& \textbf{HotpotQA}
& \textbf{BrowseComp+}
& \textbf{BrowseComp+} \\
\midrule
40,315 & 30 & 30 & 198 & 90,447 & 200 & 1,066 & 200 \\
\bottomrule
\end{tabular}
\caption{Training and test data statistics across datasets.}
\label{tab:dataset_stats}
\end{table}

\textbf{Remark.} We stop training once \ourframework has converged. As a result, the limited number of training steps may not consume the entire training dataset reported in \cref{tab:dataset_stats}.


%% file: sections/appendix/training_setup.tex
\section{Training Setup}
\label{ap:training_setup}


\subsection{Key Hyper-parameters}


\begin{table}[h]
\centering
\small
\begin{tabular}{l c}
\toprule
\textbf{Key Hyper-parameters} & \textbf{Value} \\
\midrule
train\_batch\_size & 64 \\
ppo\_mini\_batch\_size & 256 \\
ppo\_micro\_batch\_size\_per\_gpu (LLM) & 2 \\
ppo\_micro\_batch\_size\_per\_gpu (RLM) & 1 \\
log\_prob\_micro\_batch\_size\_per\_gpu (LLM) & 4 \\
log\_prob\_micro\_batch\_size\_per\_gpu (RLM) & 2 \\
max\_prompt\_length & 15{,}000 \\
max\_validation\_prompt\_length & 15{,}000 \\
max\_response\_length (LLM) & 8{,}192 \\
max\_response\_length (RLM) & 120{,}000 \\
rollout / group size & 32 \\
max\_concurrency (sub-agent) & 128 \\
hardware & 8 $\times$ 141\,GB (H200) \\
RL framework & verl \\
\bottomrule
\end{tabular}
\caption{Key hyper-parameters used in training and reinforcement learning.}
\label{tab:hyperparameters}
\end{table}

\subsection{Correctness Computation}

For math correctness (AIME), we use string matching via Hugging Face’s Math-Verify.\footnote{\url{https://github.com/huggingface/Math-Verify}} For the other benchmarks, we use Llama-3.3-70B-Instruct \cite{grattafiori2024llama3herdmodels} as an LLM-as-a-judge to determine correctness.


%% file: sections/appendix/discuss_control.tex
\section{Additional Discussion on Factors Affecting MAS Performance}
\label{ap:discuss_control}

\subsection{MAS vs. SAS for Robustness Axis}
\label{ap:mas_sas_robustness}

\begin{center}
\vspace{-0.5\baselineskip}
\includegraphics[width=0.5\columnwidth]{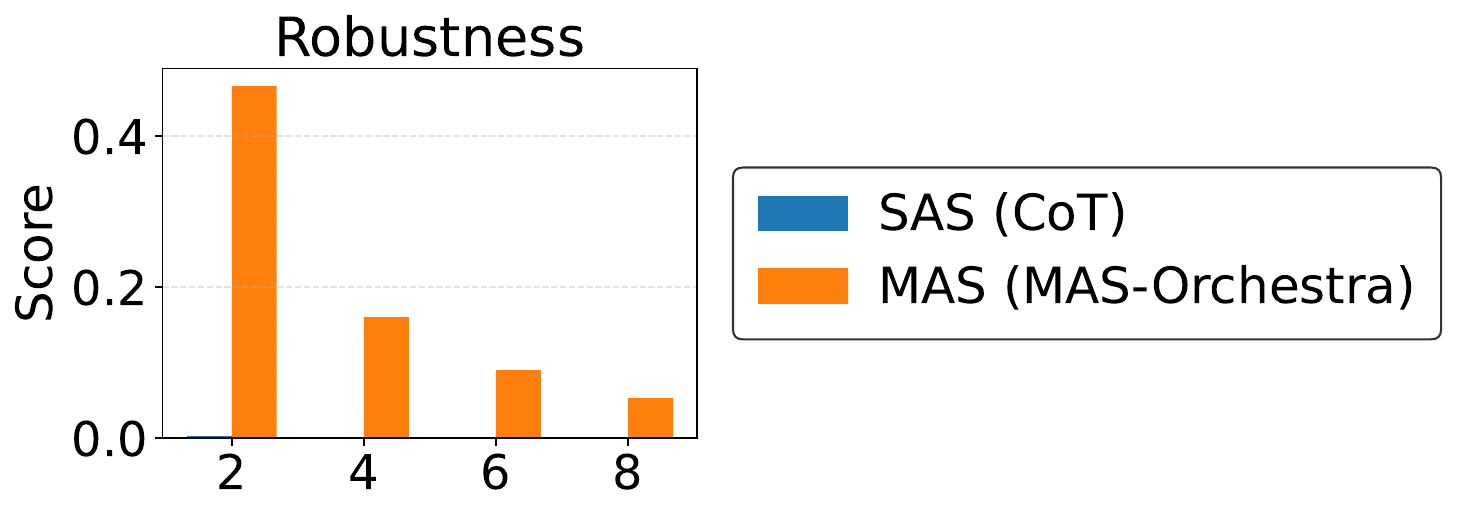}
\vspace{-0.5\baselineskip}
\captionof{figure}{\small \avgat{8} in the  \robustness setting with \ossonetwentylow as the sub-agent. (SAS performance is too low to be visible.)}
\label{fig:robustness}
\end{center}

In \cref{tab:sas_mas_example}, we demonstrate a sample from the \robustness axis (\robustness equals to 2) using \qwen as the orchestrator and \ossonetwenty as the sub-agent (or agent if SAS). See \cref{sec:subtask_structure} for detailed descriptions and observations.

\begin{table*}[h]
\centering
\small
\begin{tabularx}{\textwidth}{l X X}
\toprule
 & \textbf{Single-agent System (SAS)} & \textbf{Multi-agent System (MAS)} \\
\midrule
\textbf{Problem}
& \multicolumn{2}{p{0.76\linewidth}}{
From the passage, identify and return (1) the magic number for thundering-autumn and (2) the number of Accordion File that Beach Backpack has, wrapped in \texttt{\textbackslash boxed\{\}} and separated by a double newline.
} \\
\midrule
\textbf{Thinking}
&
We identified the magic number as 7381734 from the note.  
The Insulated Backpack's Accordion File is 21, and the Beach Backpack's Accordion File is twice that, so it is 42.  
The answer consists of two boxed numbers separated by a double newline.
&
The task is decomposed into two sub-problems:  
(1) extract the magic number for thundering-autumn from the passage;  
(2) compute the number of Accordion File that Beach Backpack has using the given relations.  
A sink agent formats the final output.
\\
\midrule
\textbf{MAS Design}
&
N/A
&
\hspace{-50mm}
\raisebox{-0.5\height}{\includegraphics[height=4.1cm]{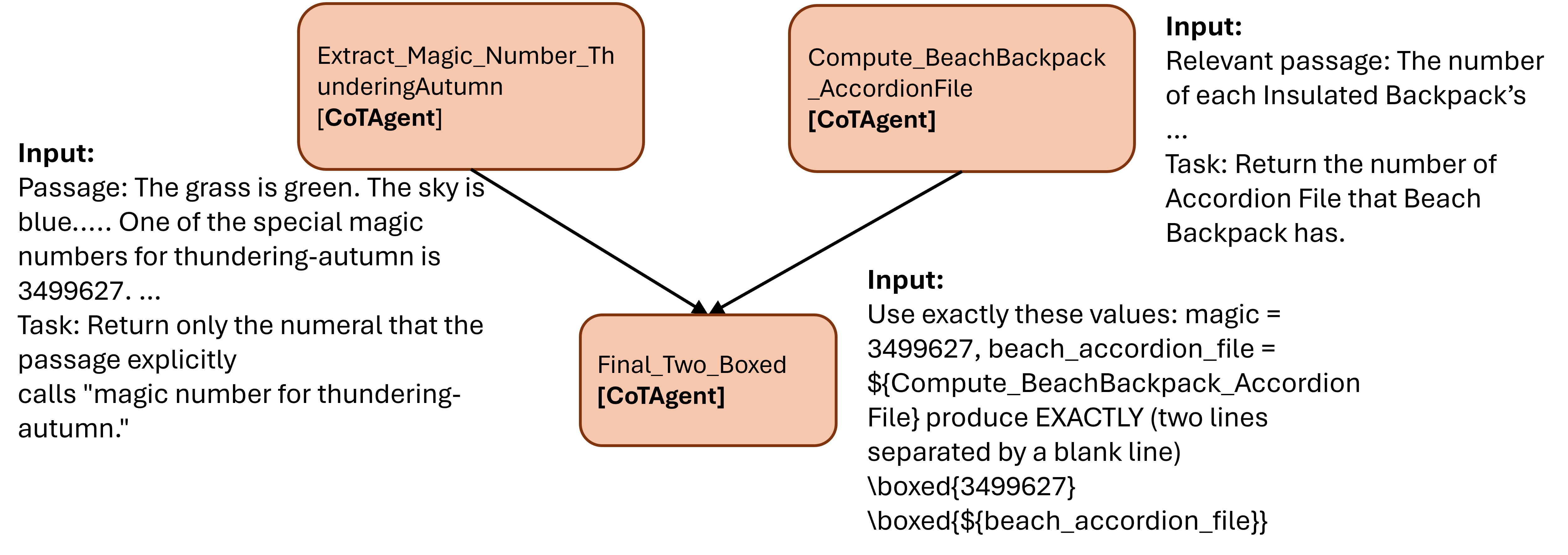}}
\\
\midrule
\textbf{Execution Output}
&
\boxed{7381734}

\boxed{42}
&
\boxed{3499627}  

\boxed{19}
\\
\midrule
\textbf{Ground Truth Answer}
& \multicolumn{2}{c}{
\boxed{3499627} \boxed{19}
} \\
\bottomrule
\end{tabularx}
\caption{Comparison of Single-agent System (SAS) and Multi-agent System (MAS) on the same problem instance. 
}
\label{tab:sas_mas_example}
\end{table*}

\subsection{Robustness and Adversarial-aware Training.} Importantly, performance on the \robustness axis in \cref{fig:mas_combined_comparison} incorporate {combined training data that include} adversarial examples. In practice, however, such examples is often unavailable or omitted. To study this effect, \cref{fig:robustness_combine} compares models
trained {on combined data} with and without adversarial examples. We observe that when adversarial data is excluded, performance on the \robustness axis remains poor (nearly
zero and comparable to SAS performance in \cref{fig:robustness}). This result
indicates that while combined training generalizes well across most axes, it
fails to generalize to robustness. Explicit inclusion of adversarial data is therefore necessary to improve robustness performance.

\textbf{Remark.} Separate training and combined training require different numbers of training
steps, as the combined dataset is naturally larger. The goal of our experiments
is not to fully converge in the RL training, but rather to
examine how different factors affect the performance of MAS. In our experiments,
we allocate twice as many training steps for combined training compared to
separate training. We expect that further performance gains are possible for
combined training with additional training steps.


\subsection{Statistics of Agents for All Axes}

\begin{figure*}[h!]
\centering
\includegraphics[width=\columnwidth]{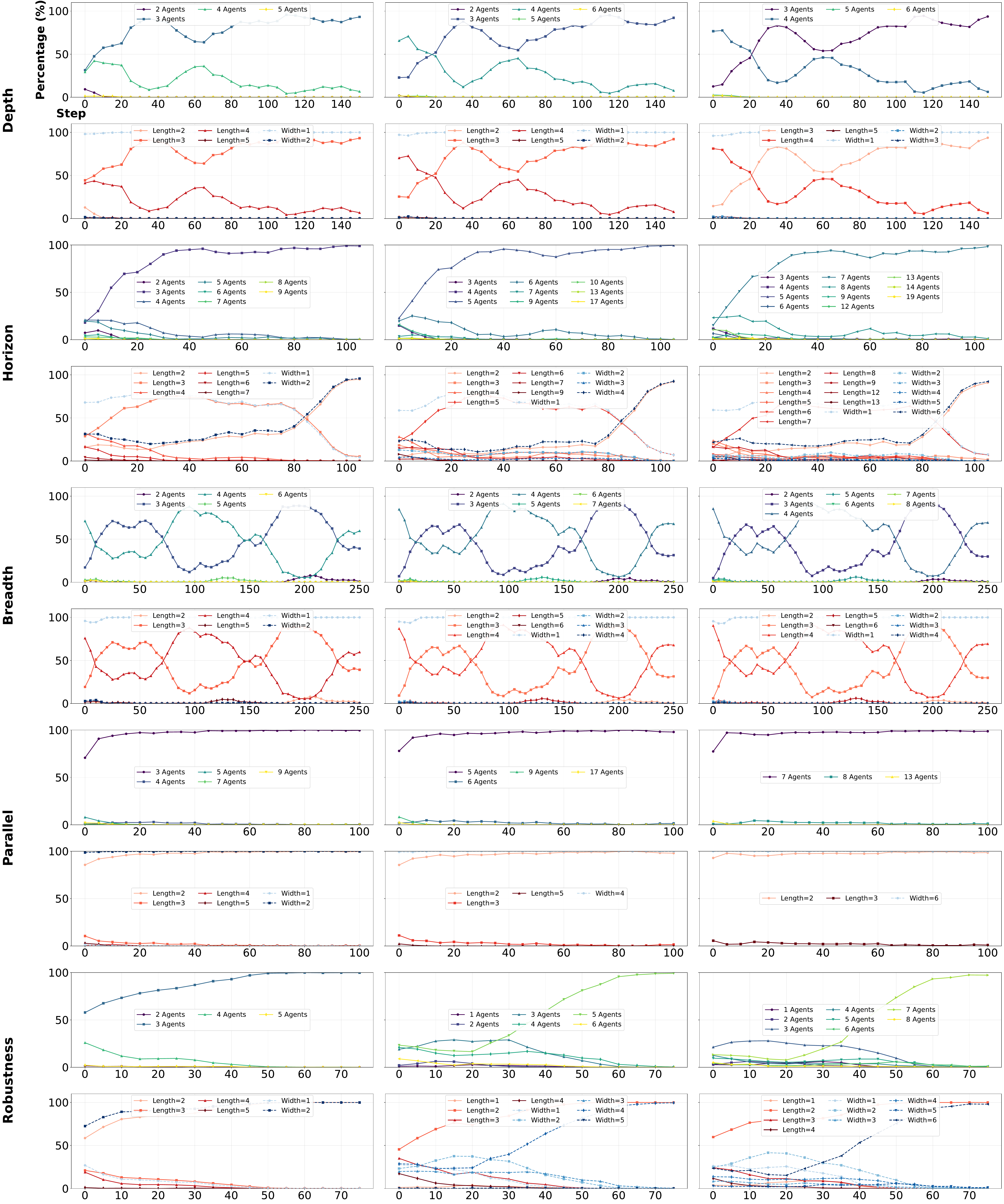}
\vspace{-\baselineskip}
\captionof{figure}{\small Statistics of agents for different axes (the correspnding values are 2, 4, 6 from left to right).}
\label{fig:axes_detail}
\end{figure*}

{\cref{fig:axes_detail} reports detailed statistics about the total number of sub-agents generated for different evaluation axes (\depth, \horizon,\ breadth, \myparallel and \robustness).} These results show that the orchestrator can largely learn to mirror the underlying task structure in its orchestration design. However, this structural alignment does not necessarily translate into performance gains over SAS, {as discussed earlier}.


Below, we provide several detailed observations.

\subsubsection{Given Axes, Observations Across Values.}

We first examine observations within a given axis. As task complexity increases (i.e., as the axis value becomes larger), \textbf{more sub-agents are generated}. This behavior is expected and indicates that the orchestrator effectively adapts to increasing task complexity.

We further observe that, within the same axis, the overall orchestration pattern remains similar across different values, despite the increase in the number of sub-agents. This suggests that the \textbf{orchestrator is stable} with respect to its learned structural strategy and scales the number of sub-agents without changing the underlying orchestration structure as task complexity increases.




\subsubsection{Given Value, Observations Across Axes.}

We observe that \depth and \breadth tend to converge to a general sequential pattern. Specifically, the length typically ranges from 1 to 5, while the width remains much smaller, usually up to 2. A closer inspection of the generated examples shows that the resulting MAS commonly follows a structure of \text{Parse} $\rightarrow$ \text{Solve} $\rightarrow$ \text{Verify} $\rightarrow$ \text{Final Answer}. This indicates that, for tasks with sequential dependency chains or fan-in structure, a linear decomposition with explicit intermediate reasoning is sufficient and stable under the given setting.

In contrast, \myparallel tends to converge to parallel MAS structures, where the width is typically aligned with the specified \myparallel value. This behavior closely mirrors the underlying task structure and suggests that the orchestrator can effectively learn and exploit independent sub-task structures under the current training method.

For \horizon, the observed structural diversity reflects uncertainty about when intermediate results should be made explicit. Some tasks benefit from early synthesis of partial answers, whereas others require delayed aggregation after several reasoning steps. The orchestrator adapts by varying both the number and the ordering of sub-agents across horizon settings, suggesting that \ourframework learns temporal control strategies from task-level feedback rather than relying on a fixed reasoning depth.

For \robustness, we observe a tendency toward parallel MAS structures. This orchestration strategy allows different sub-agents to address different sub-tasks and helps isolate and mitigate adversarial or incorrect information, either within individual sub-agents or through additional verification or moderation components.

\subsection{Separate vs. combined training.} 

In \cref{fig:sub_task_sub_agent}, we train the orchestrators using training
data from different axes. Now, we examine: \textit{can data from different axes be combined for
training, and can the resulting orchestrator generalize when tested on
individual axes?} \cref{fig:mas_combined_comparison} reports results obtained by training on
combined data from all five axes. We observe that the
performance across different axes is highly similar, indicating that the
 the orchestrator can effectively generalize when
trained on combined data.




\begin{center}
\vspace{-0.5\baselineskip}
\includegraphics[width=0.6\columnwidth]{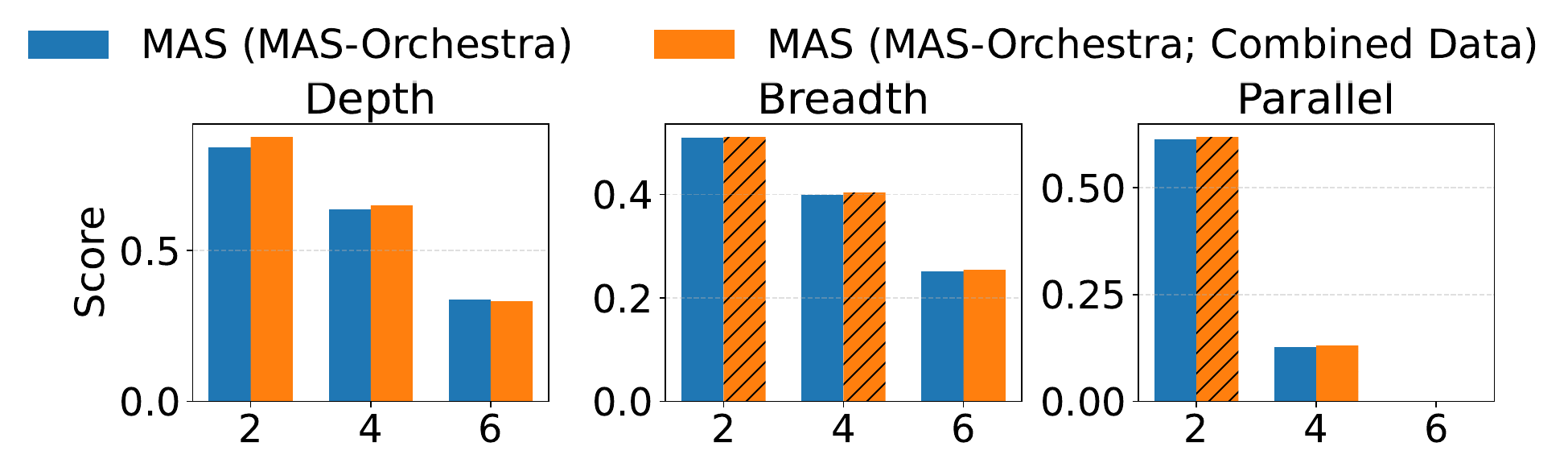}
\vspace{-10pt}
\captionof{figure}{\small \avgat{8} comparing separate and combined training. 
}
\label{fig:mas_combined_comparison}
\end{center}

\textbf{\robustness.} 
\cref{fig:robustness_combine} compares models
trained on the combined data, with and without adversarial samples. We observe that incorporating adversarial training data leads to a substantial improvement over training without such
data. This result indicates that explicitly including adversarial examples in the combined training data is
necessary for improving robustness performance.

\begin{center}
\includegraphics[width=0.6\columnwidth]{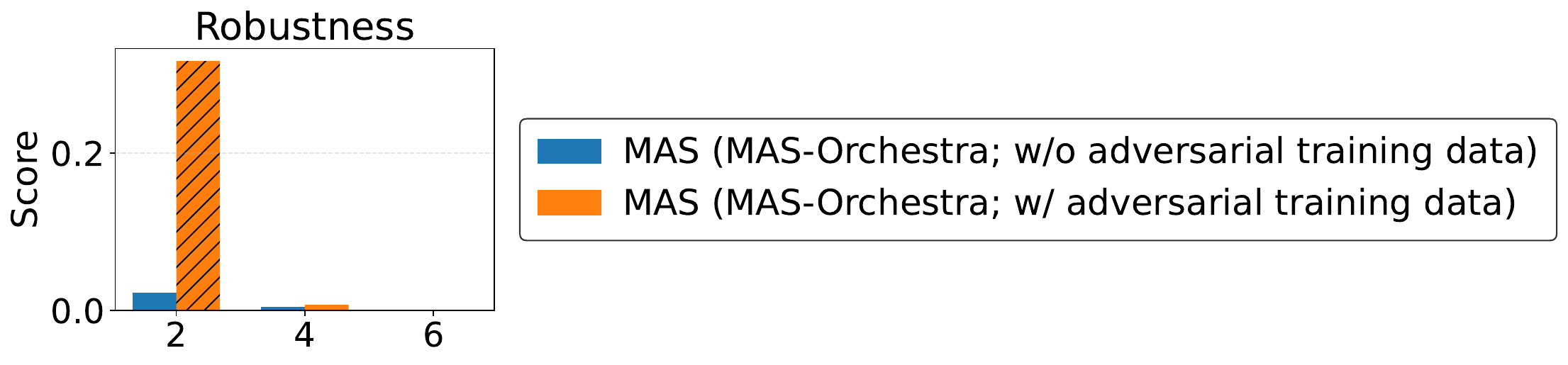}
\vspace{-0.5\baselineskip}
\captionof{figure}{\small \avgat{8} of \robustness combined training w/ and w/o adversarial-aware training}
\label{fig:robustness_combine}
\end{center}


\subsection{How Does RLM Orchestrator Size Affect MAS Performance?} 

Since \qwen is a 7b model while the RLM used above has 20b  parameters, one may
wonder whether model size plays a role in the observed performance gap. To
investigate this, we compare a 7b RLM DeepSeek-R1-Distill-Qwen-7B (\deepseekqwen) \cite{deepseekai2025deepseekr1incentivizingreasoningcapability} with \qwen under
combined training, using a similar number of training steps. As shown in
\cref{fig:deepseek}, even when controlling for model size, the RLM consistently underperforms the LLM across different axes. This result suggests that the
current RLM is not well suited to serve as an orchestrator. 


\cref{fig:deepseek} compares performance when using the same-size instruction-tuned LLM versus RLM. A detailed description is provided in \cref{sec:reasoning_meta_agent}.

\begin{center}
\vspace{-0.5\baselineskip}
\includegraphics[width=0.6\columnwidth]{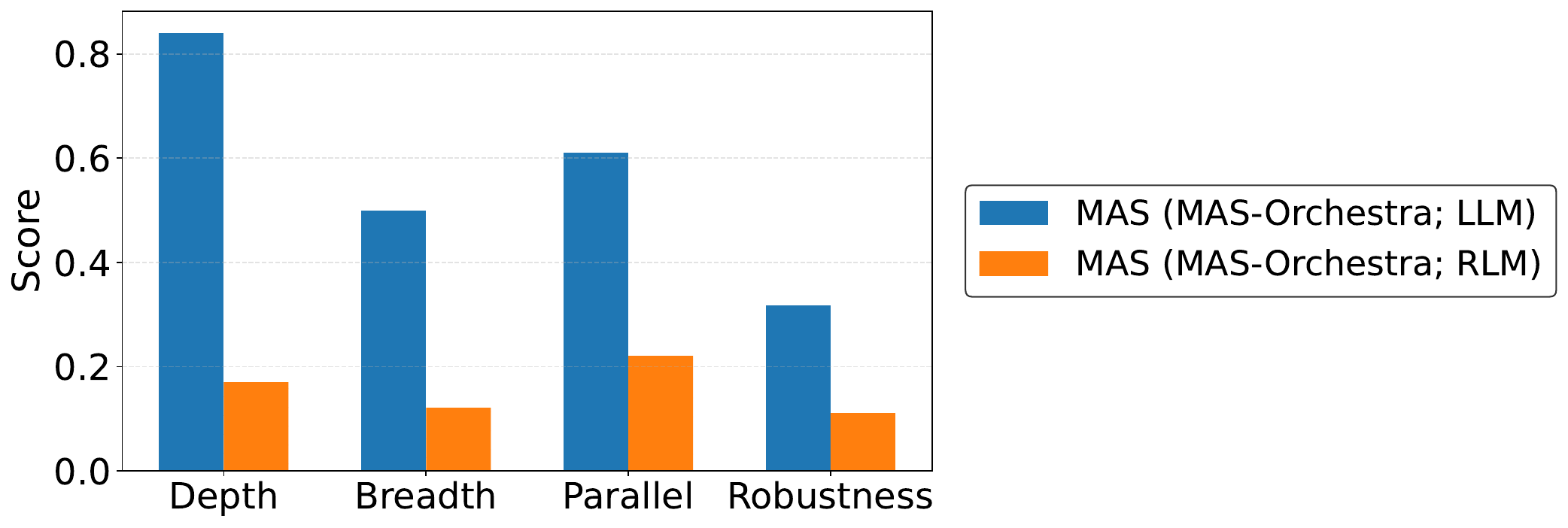}
\captionof{figure}{\small \avgat{8} comparing LLM (\qwen) and RLM (\deepseekqwen) \textbf{with the same size} as orchestrator.}
\label{fig:deepseek}
\end{center}

\subsection{Complexity Generalization}
\label{ap:complexity_generalization}

 
To facilitate the investigation of  \textit{complexity generalization}, we train the orchestrator on instances with smaller axis values and evaluate on both the training values and higher, held-out axis values in \ourbenchmark. {Specifically, the model is trained on \depth values from 2 to 8 and evaluated up to 12. For \breadth and \myparallel, training covers values from 2 to 4, with evaluation extending to 8. For \horizon and \robustness, training values range from 2 to 6, and evaluation values range up to 8.} 

As the tasks become more complex, performance decreases. Importantly, out-of-distribution complexity follows the same trend as in-distribution complexity.
This suggests that \ourframework induces controlled increases in
task complexity and exhibits consistent generalization behavior as the task
complexity grows. 

\subsection{LLM vs. RLM as Orchestrator}
\label{ap:rlm_orchestrator}

\textbf{Statistics of Agents.} \cref{fig:7b_120b_vs_20b_120b_depth4} shows agent statistics in the
MAS proposed by MAS-Orchestra. See \cref{sec:reasoning_meta_agent} for detailed descriptions and observations.

\begin{center}
\vspace{-0.5\baselineskip}
\includegraphics[width=0.8\columnwidth]{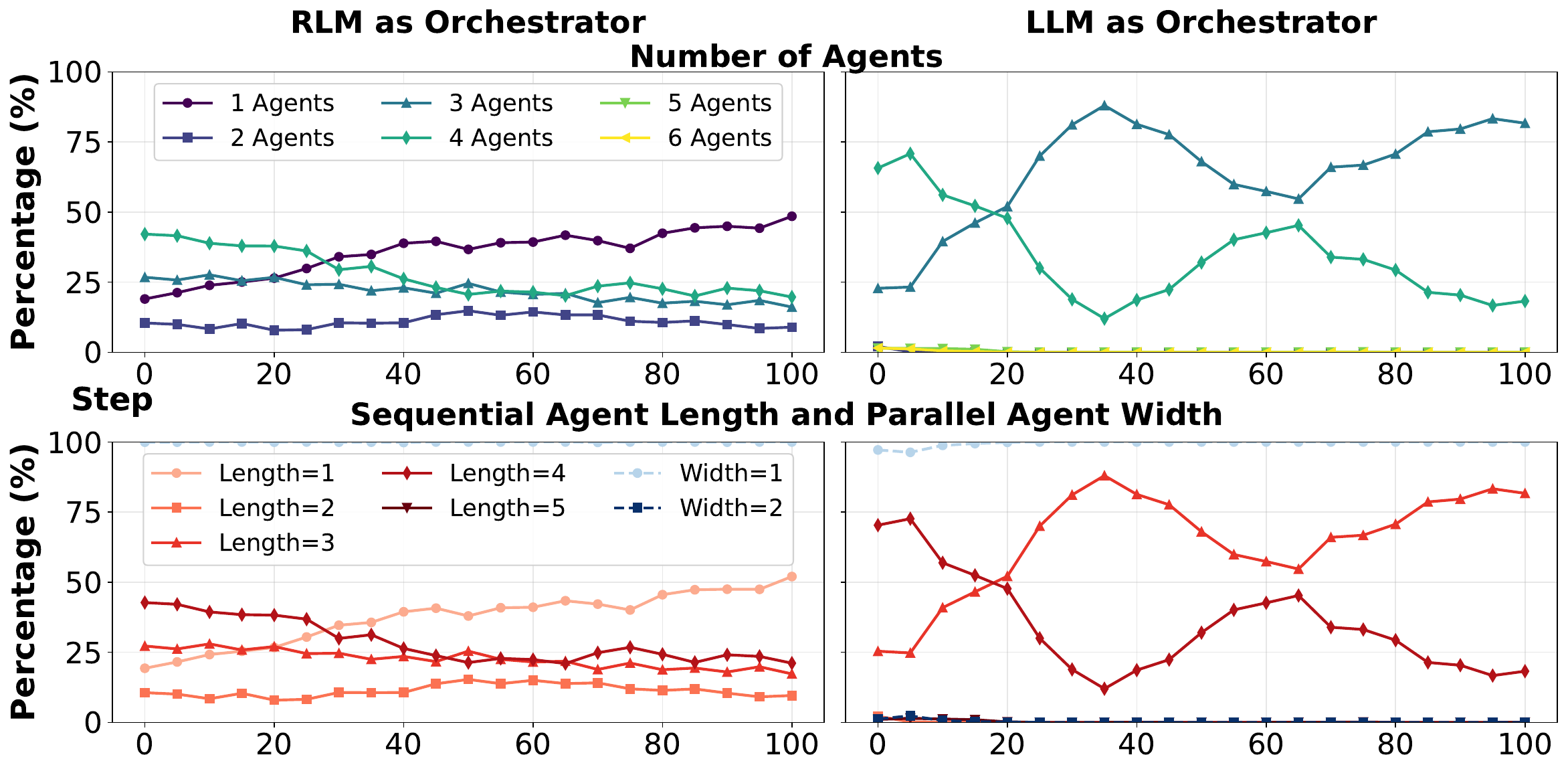}
\captionof{figure}{\small Statistics of agents in the generated MAS over the training steps (use \depth equals 4 as an example). {The \textit{number of agents} measures the total number of sub-agents, the \textit{sequential agent length} measures  the length of dependency chain  and \textit{parallel agent width} measures the in degree of sub-agent.}}
\label{fig:7b_120b_vs_20b_120b_depth4}
\end{center}

\textbf{Example.}
In \cref{tab:rlm_example}, we demonstrate a sample from the \depth axis (\depth equals to 4) using \qwen as the orchestrator and \ossonetwenty as the sub-agent. See \cref{sec:reasoning_meta_agent} for detailed descriptions and observations.

\begin{table*}[h!]
\centering
\small
\begin{tabularx}{\textwidth}{l X X}
\toprule
\textbf{Orchestrator} & \textbf{\osstwenty} & \textbf{\qwen}\\
\midrule
\textbf{Problem} & \multicolumn{2}{p{0.76\linewidth}}{
 The number of each Golden Gate Park in San Francisco's Anteater equals 21. The number of each Bois de Boulogne in Paris's Kangaroo equals each Chapultepec Park in Mexico City's Wolf. The number of each Bois de Boulogne in Paris's Wolf equals 6. .... How many Moose does Chapultepec Park in Mexico City have?}\\
\midrule
\textbf{Thinking} &
analysisWe need to design agents. Problem: find number of Moose in Chapultepec Park in Mexico City. Let's parse quickly.

Variables: Let's denote:

GGP_SF_Anteater = 21 (given)
GGP_SF_Wolf = 12
GGP_SF_Moose = ?
GGP_SF_Kangaroo = ?
GGP_SF_Moose equals each Chapultepec Park in Mexico City's Wolf. So GGP_SF_Moose = CP_MX_Wolf.

...

No other constraints involve CP_MX_Creatures. So no issue.

Thus answer 42.

We can create single agent? Single agent can solve. But we may want multi-agent to show reasoning. However constraints allow single agent. But we can do multi-agent for demonstration.

Simpler: one CoTAgent with empty input to solve directly. That is simplest. But we must ensure final answer. Provide final agent with empty input. That yields final answer.

But we might want to show intermediate steps? Not necessary.

Given we can do single agent. Let's do single agent.

...

Let's produce.

assistantfinal<thinking>
  The problem is a straightforward system of linear equations with a single unknown of interest: the number of Moose in Chapultepec Park, Mexico City.  
  All other variables are either given directly or can be expressed in terms of the known constants.  
  A single CoTAgent can parse the statements, instantiate the variables, perform the necessary substitutions, and produce the final numeric answer.  
  Since no intermediate agents are needed, the graph consists of just this one sink agent, which also serves as the start node. 
  & 
  <thinking>
  This problem involves setting up and solving a system of equations. Using a single agent would be insufficient as it would mix the steps of parsing the problem and solving it, increasing the risk of errors. We need to parse the problem into equations first, then solve the system of equations, and finally verify the solution. This separation ensures accuracy and allows for independent verification of each step.
  The final answer will be provided by the Final_Answer agent.
\\
\midrule
\textbf{Execution Output} &
21
&
19
\\
\midrule
\textbf{Ground Truth Answer} & \multicolumn{2}{c}{
19}
\\
\bottomrule
\end{tabularx}
\caption{Comparison of instruction-tuned LLM (\qwen) and RLM (\osstwenty) on the same problem instance.
}
\label{tab:rlm_example}
\end{table*}

\subsection{\robustness under high reasoning effort when context lengths are longer}

\begin{center}
\vspace{-0.5\baselineskip}
\includegraphics[width=\columnwidth]{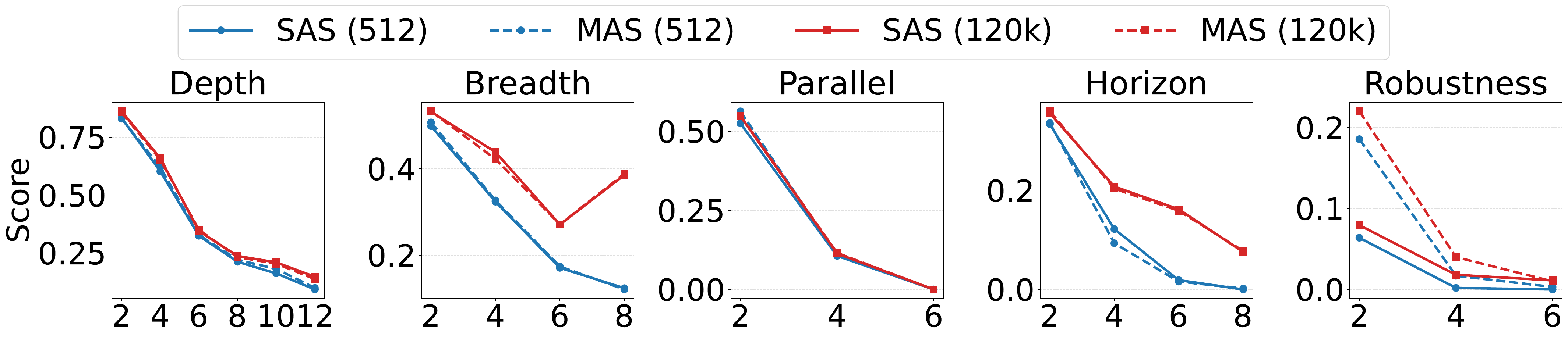}
\vspace{-\baselineskip}
\captionof{figure}{\small \avgat{8} comparing different maximum context length with \qwen as orchestrator and \ossonetwenty as sub-agent.}
\label{fig:max_length}
\end{center}

%% file: sections/appendix/discuss_public.tex
\section{Additional Discussion for Public Benchmark Evaluation}
\label{ap:disucuss_public}


\subsection{Sub-agents and Baselines Setup}
\label{ap:discuss_baselines}

For \texttt{SearchAgent}, we adapt the OpenDeepResearch code\footnote{\url{https://github.com/langchain-ai/open_deep_research}}
 to use DuckDuckGo Search\footnote{\url{https://duckduckgo.com/}}
 for HotpotQA and a BM25 retriever \citep{robertson1994okapi} for BrowseComp+. \cref{tab:baseline_setup} summarizes the sub-agents, including the orchestration decisions in \ourframework (i.e, configurations required to be generated by the orchestrator), and their corresponding implementations.

\begin{table*}[h]
\centering
\small
\begin{tabular}{l l l}
\toprule
\textbf{Sub-agent} & \textbf{Orchestration Decisions} & \textbf{Implementation} \\
\midrule
CoTAgent & Input & \cref{app:sub_agents} \\
SCAgent & Input & \cref{app:sub_agents} \\
ReflexionAgent & Input & \cref{app:sub_agents} \\
DebateAgent & Input, Debate role & \cref{app:sub_agents} \\
SearchAgent (HotpotQA) & Input & Autonomous (Multi-turn + DuckDuckGoSearch) \\
SearchAgent (BrowseComp+) & Input & Autonomous (Multi-turn + BM25) \\
\bottomrule
\end{tabular}
\caption{Summary of the sub-agents.
}
\label{tab:baseline_setup}
\end{table*}

For baseline systems, we run all baselines using the official code released by the original authors.

\subsection{Additional Observations on Overall Results}
\label{ap:addition_observation}

\cref{sec:experiemnt} reports the superiority of \ourframework over all baselines. In this section, we provide additional observations based on our experience running these systems.

\paragraph{Inference-time orchestration baselines.}
Inference-time orchestration methods are often computationally costly, particularly AFlow and MAS-Zero. MAS-Zero requires an orchestrator larger than 32B parameters to produce meaningful results, as also noted in the original paper \citep{Ke2025MASZero}. AFlow achieves the strongest baseline performance among inference-time methods, but still underperforms \ourframework by a clear margin.

\paragraph{Training-based orchestration baselines.}
Training-based orchestration baselines perform unexpectedly worse than many inference-time methods. We observe that MAS-GPT frequently falls back to Self-Consistency (SC), which itself is a strong single-agent system, when no valid MAS are generated. This fallback behavior contributes substantially to its reported gains when compared with ToolOrchestra.

\begin{figure*}[h]
\centering
\includegraphics[width=0.5\columnwidth]{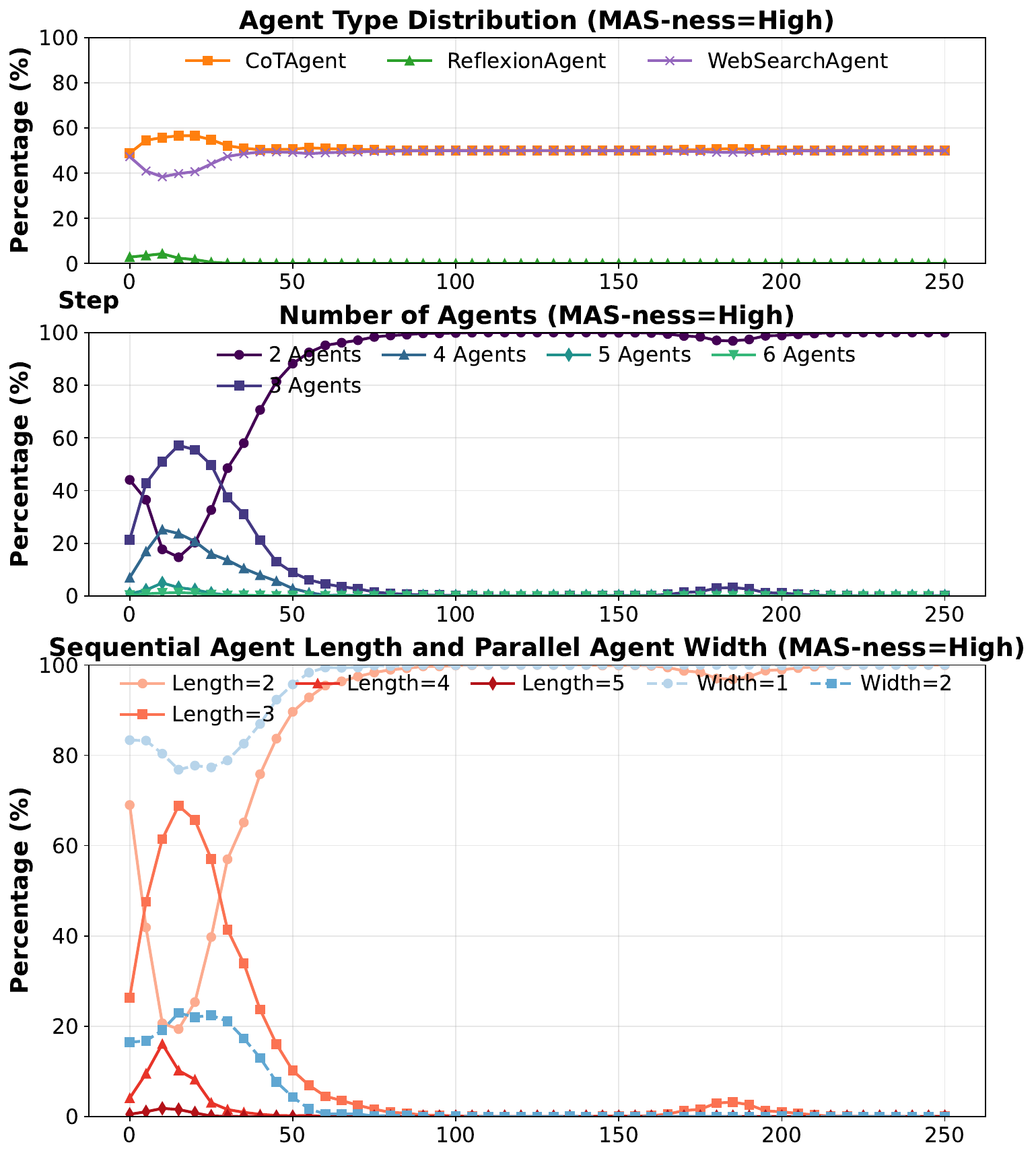}
\captionof{figure}{\small Statistics of agents for \masness as high (HotpotQA).}
\label{fig:7b_120b_hotpotqa_high}
\end{figure*}

We further observe that ToolOrchestra often refuses to call any tool. In ToolOrchestra, the sub-agent is treated as a tool, and such refusal results in degraded performance when the sub-agent is stronger than the orchestrator for solving the task. In contrast, \ourframework does not rely on any manually designed fallback strategy. All MAS are generated directly by the learned orchestrator. Under the DoM notation, sub-agents in \ourframework are consistently and effectively utilized, as illustrated in Figures \ref{fig:7b_120b_aime24_low}, \ref{fig:7b_120b_browse_comp_plus_high}, and \ref{fig:7b_120b_hotpotqa_high}.

 

\begin{center}
\vspace{-0.5\baselineskip}
\includegraphics[width=0.5\columnwidth]{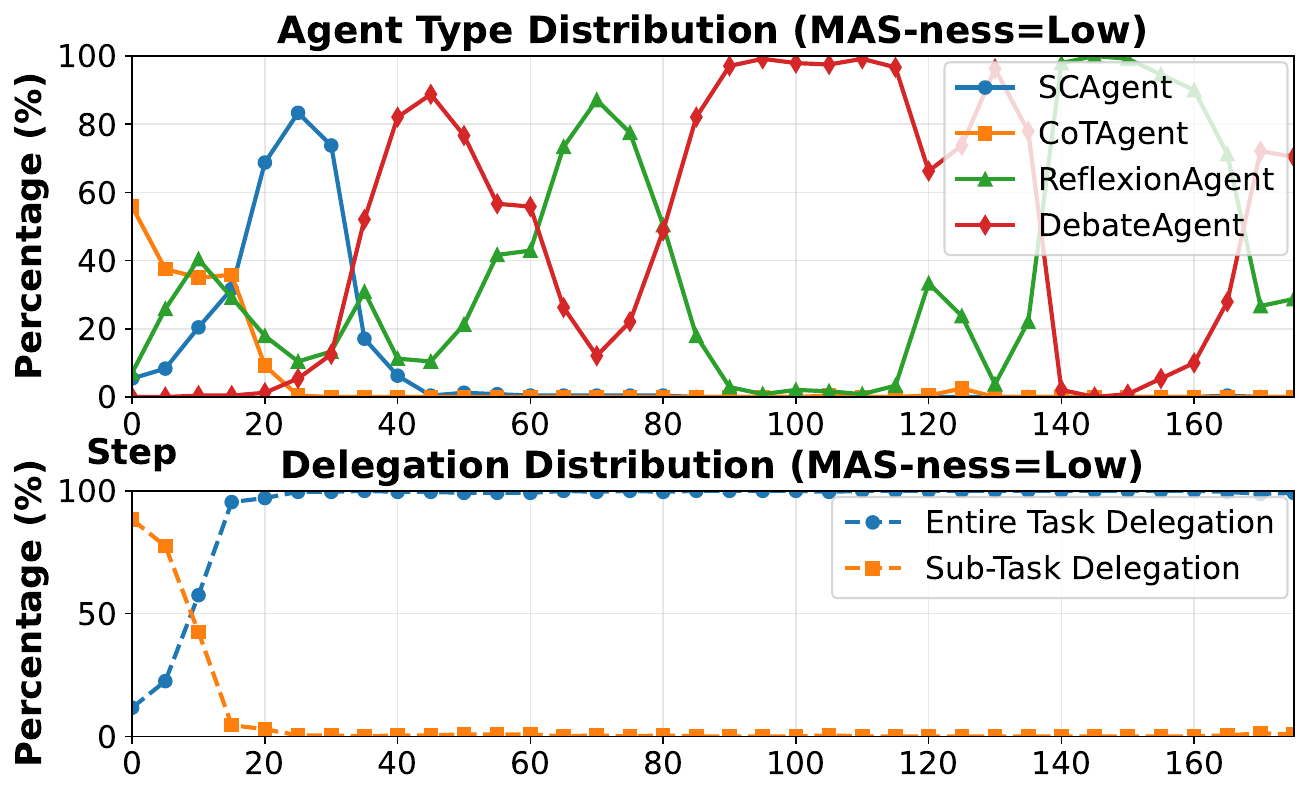}
\captionof{figure}{\small Statistics of agents for low \masness (AIME24).}
\label{fig:7b_120b_aime24_low}
\end{center}

\begin{center}
\includegraphics[width=0.5\columnwidth]{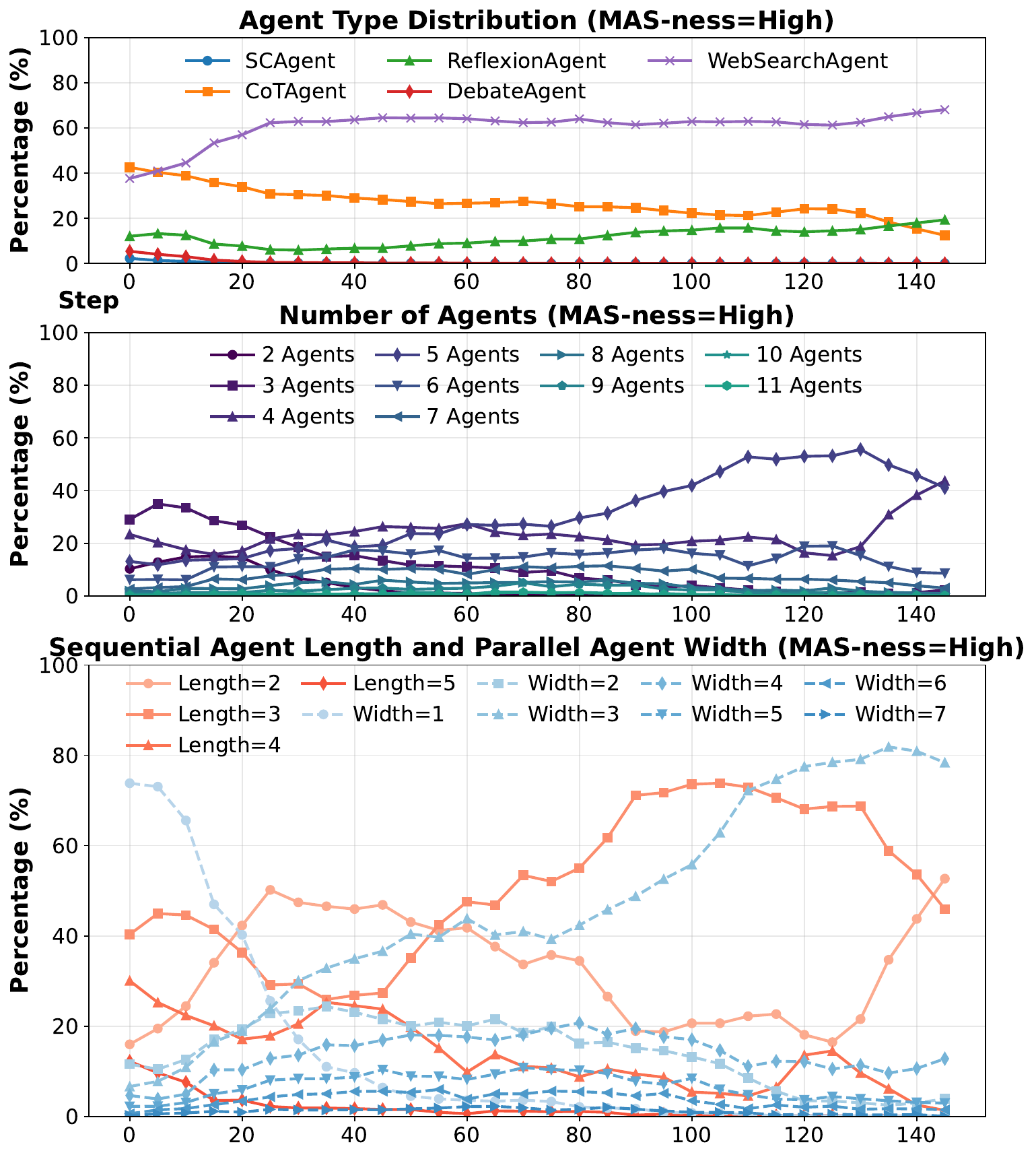}
\vspace{-0.5\baselineskip}
\captionof{figure}{\small Statistics of agents for high \masness (BrowseComp+).}
\label{fig:7b_120b_browse_comp_plus_high}
\end{center}


\subsection{Benchmarks vs. Generated MAS}
\label{ap:benchmark_generated_mas}

Beyond the results reported in \cref{sec:experiemnt}, We show the statistics of HotpotQA in \cref{fig:7b_120b_hotpotqa_high}. 
For HotpotQA, the orchestrator often decides to use only a single \texttt{SearchAgent} (together with a final \texttt{CoTAgent}, for a total of 2 agents), even though the question may allow parallel searches. This suggests that the questions are relatively simple and that one search step is sufficient to solve them. 

Together with the AIME results in Figures \ref{fig:7b_120b_aime24_low} and the BrowseComp+ results in \ref{fig:7b_120b_browse_comp_plus_high}, these observations show that \ourframework can dynamically adapt to a given task by proposing MAS designs that align with the underlying sub-task structure and by delegating execution to the most effective agent configurations. We note that the generated MAS designs do not necessarily perfectly match the underlying task structure (for example, in HotpotQA), due to diverse underlying LLM capabilities, which highlights an advantage over manually designed orchestration strategies.





\subsection{Cost Comparison}
\label{ap:cost_breakdown}

In \cref{fig:parato_front}, we illustrate the cost–performance trade-off. In \cref{tab:cost_breakdown}, we report the corresponding detailed inference costs on AIME24 and GPQA. Holistic orchestration is expected to be more efficient than sequential orchestration systems, and even more so compared to inference-time orchestration methods. Consistent with this intuition, \ourframework achieves more than \textbf{10$\times$} efficiency in terms of the number of LLM calls, the number of tokens (including both prompt and completion tokens), and overall cost (based on Together AI pricing at the time of writing). We also observe substantially faster wall-time performance. We note that wall-time reflects not only the algorithmic design but also implementation details, where we apply extensive optimization (e.g. asynchronous sub-agent execution) and will release the full codebase to the community.



\begin{table}[h]
\centering
\small
\setlength{\tabcolsep}{6pt}
\renewcommand{\arraystretch}{1.15}
\begin{tabular}{lcccccccc}
\toprule
& \multicolumn{4}{c}{\textbf{AIME24}} & \multicolumn{4}{c}{\textbf{GPQA}} \\
\cmidrule(lr){2-5} \cmidrule(lr){6-9}
\textbf{Method}
& \textbf{Calls}
& \textbf{Tokens (M)}
& \textbf{Cost (\$)}
& \textbf{Time (ks)}
& \textbf{Calls}
& \textbf{Tokens (M)}
& \textbf{Cost (\$)}
& \textbf{Time (ks)} \\
\midrule
CoTAgent
& 30 & 0.05 & 0.03 & 0.38
& 198 & 0.16 & 0.06 & 1.18 \\
SCAgent
& 156 & 0.25 & 0.13 & 1.88
& 990 & 0.80 & 0.29 & 5.43 \\
DebateAgent
& 332 & 0.37 & 0.13 & 2.29
& 2178 & 2.04 & 0.55 & 7.65 \\
ReflexionAgent
& 112 & 0.14 & 0.06 & 0.67
& 518 & 0.44 & 0.13 & 2.50 \\
\midrule
MaAS
& 501 & 1.28 & 0.62 & 6.96
& 2936 & 7.52 & 3.11 & 39.52 \\
AFlow
& 1288 & 3.47 & 1.46 & 19.28
& 7404 & 12.38 & 4.52 & 56.94 \\
\midrule
MAS-GPT
& 228 & 0.74 & 0.37 & 5.47
& 1516 & 3.90 & 1.76 & 29.56 \\
ToolOrchestra
& 1319 & 3.77 & 1.76 & 31.35
& 7829 & 15.38 & 5.79 & 92.47 \\
\midrule
\textbf{\ourframework}
& \textbf{51} & \textbf{0.03} & \textbf{0.01} & \textbf{0.50}
& \textbf{261} & \textbf{0.22} & \textbf{0.07} & \textbf{0.46} \\
\bottomrule
\end{tabular}
\caption{Inference-time statistics on AIME24 and GPQA. Token counts are in millions (M) and wall time in thousands of seconds (ks).}
\label{tab:cost_breakdown}
\end{table}

%% file: sections/appendix/related.tex
\section{Additional Related Work on Inference-Time Orchestration}
\label{ap:related}

Recent work on automatic MAS design typically operates at inference time and relies on a validation set to guide adaptation.  MASS~\citep{zhou2025multiagentdesignoptimizingagents} uses rejection sampling to select effective agent configurations, and MaAS~\citep{zhang2025multiagentarchitecturesearchagentic} extends MASS with a question-wise masking mechanism to adapt subnetworks. ADAS~\citep{hu2025automated} and AFlow~\citep{zhang2024aflowautomatingagenticworkflow} frame MAS design as a code generation task. ADAS stores and searches historical designs based on validation performance, while AFlow enhances this with Monte Carlo Tree Search. One exception is MAS-Zero \citep{Ke2025MASZero}, which performs fully inference-time automatic MAS design without relying on a validation set. However, these inference-time methods exhibit limited adaptability, as unreliable validation signals or incorrect guidance from LLM-based judges can misdirect the adaptation process.

%% file: sections/appendix/grpo.tex
\section{Policy Optimization in \ourframework}
\label{ap:grpo}

The policy in \cref{eq:general_objetive} is optimized by maximizing
\begin{equation}
\label{eq:grpo}
\begin{aligned}
&J_{\text{GRPO}}(\theta)
=\;\\
&\mathbb{E}_{x \sim \mathcal{D},\, \{a_i\}_{i=1}^{K} \sim \pi_{\theta_{\text{old}}}(\cdot \mid x,m)}
\frac{1}{K}\sum_{i=1}^{K}\frac{1}{|a_i|}\sum_{t=1}^{|a_i|}
\Bigg\{
\min\!\Bigg[
r_{i,t}(\theta)\,
\hat{A}_{i,t},\;
\mathrm{clip}\!\Bigg(
r_{i,t}(\theta),\,
1-\epsilon,1+\epsilon
\Bigg)\hat{A}_{i,t}
\Bigg]
-\;\beta\, D_{\mathrm{KL}}\!\left(\pi_\theta \,\|\, \pi_{\mathrm{ref}}\right)
\Bigg\}.
\end{aligned}
\end{equation}

where
\begin{equation}
r_{i,t}(\theta)
=
\frac{\pi_\theta(a_{i,t}\mid x, a_{i,<t}, m)}
{\pi_{\theta_{\text{old}}}(a_{i,t}\mid x, a_{i,<t}, m)}.
\end{equation}
For outcome-supervised training, we assign a sequence-level group-normalized advantage
to all tokens in $a_i$:
\begin{equation}
\hat{A}_{i,t}=\hat{R}_i,
\qquad
\hat{R}_i
=
\frac{R_i - \mathrm{mean}(\{R_j\}_{j=1}^K)}
{\mathrm{std}(\{R_j\}_{j=1}^K)}.
\end{equation}
where $\bar{R} = \frac{1}{K}\sum_{i=1}^K R_i$ is the group-wise average reward,
and $\theta_{\text{old}}$ denotes the parameters of the policy used to generate
the samples.

%% file: sections/appendix/prompt.tex
\section{Prompts for Orchestrator}
\label{app:prompt}

\subsection{Low \masness}

\begin{systemprompt}[title={System Prompt}]
You are a helpful assistant.

\masness (degree of MAS): Low

Valid Channels: thinking, agent, answer

Model: \texttt{[MODEL]}\\

An agent is a pre-configured AI personalities that can delegate tasks to. Each subagent:

1, Has a specific purpose and expertise area\\
2. Uses its own context window separate from the main conversation\\
3. (Optional) Can be configured with specific tools it's allowed to use\\
4. Includes a custom system prompt that guides its behavior \\
\\
An agent can only call within the \xmltag{thinking} channel, it should be defined in tag \xmltag{agent}. Each agent must contain \xmltag{agent_name}, \xmltag{agent_description}, \xmltag{required_arguments}, \xmltag{agent_output_id}\\

DO NOT MISS ANY REQUEST FIELDS and ensure that your response is a well-formed XML object!
\end{systemprompt}

\begin{developprompt}[title={Develop Prompt}]
Channels: \\
    \xmltag{thinking}: internal reasoning and planning \\
    \xmltag{agent}: definition of sub-agents \\
    \xmltag{answer}: final user-facing answer \\

Model (the model used in sub-agent):

    \texttt{gpt-4.1-nano}: [Introduction of the model from official website] \\
    \texttt{gpt-oss-120b}:[Introduction of the model from official website] \\
    (...list of all candidate models)\\

\masness Levels:

    \texttt{low}: direct solve or at most one agent \\
    \texttt{high}: complex multi-agent delegation\\

Sub-agent Schema (all fields required):\\

\xmltag{agent}

\quad\xmltag{agent_name}
...
\xmlendtag{agent_name}
(select one of the agents: CoTAgent, SCAgent, DebateAgent, ReflexionAgent)

\quad\xmltag{agent_description}
...
\xmlendtag{agent_description}

\quad\xmltag{required_arguments}
(make sure all required parameters are set. Must follow XML format)

\qquad\xmltag{...}...\xmlendtag{...}

\qquad\xmltag{...}...\xmlendtag{...}

\quad\xmlendtag{required_arguments}

\quad\xmltag{agent_output_id}
...
\xmlendtag{agent_output_id}

\xmlendtag{agent} \\

Available Agents: \\

\texttt{CoT}: [Description, Name, Required Argument and Summary of Implementation]

\texttt{SC}: [Description, Name, Required Argument and Summary of Implementation]

\texttt{Debate}: [Description, Name, Required Argument and Summary of Implementation]

\texttt{Reflexion}: [Description, Name, Required Argument and Summary of Implementation]

\texttt{Search}: [Description, Name, Required Argument and Summary of Implementation]

\end{developprompt}

\begin{userprompt}[title={User Prompt}]
Please solve the question step-by-step. During the reasoning process, you can address the task yourself and output the final answer in the \xmltag{answer} tag. You can also delegate the task to the agent that you designed, and output the corresponding \texttt{agent_output_id} in the \xmltag{answer} tag. You must output ALL required parameters and use EXACTLY the same field names for the agent. \\
\\
For example,
\\
If you can solve the task yourself, you will output the following:\\

\xmltag{thinking}

(20+9)*(30+7) = 600 + 140 + 270 + 63 = 1073.

\xmlendtag{thinking}

\xmltag{answer}1073\xmlendtag{answer}
\\

If you can solve the task via single-agent delegation with one tool call, you will output the following:\\

\xmltag{thinking}

This problem involves symbolic integration and applying the Fundamental Theorem of Calculus.
It requires structured reasoning rather than simple numeric computation.
I will use a calculus agent that can perform step-by-step Chain-of-Thought reasoning.
The final answer to the original question will be the output of the CoTAgent.

\xmlendtag{thinking}

\xmltag{agent}

\quad\xmltag{agent_name}CoTAgent\xmlendtag{agent_name}

\quad\xmltag{agent_description}{Definite integrals with one Chain-of-Thought call.}\xmlendtag{agent_description}

\quad\xmltag{required_arguments}

\qquad\xmltag{agent_input}\xmlendtag{agent_input}

\quad\xmlendtag{required_arguments}

\quad\xmltag{agent_output_id}
calc_agent_output
\xmlendtag{agent_output_id}

\xmlendtag{agent}

\xmltag{answer}{calc_agent_output}\xmlendtag{answer} \\

Another example:\\

\xmltag{thinking}

This question requires comparison between two close numeric choices.
To ensure accuracy, I will let two reasoning roles debate: one focusing on mathematical precision and the other on practical rounding.
The DebateAgent can capture both perspectives and reach a justified final answer.
The final answer will be the output of the DebateAgent.

\xmlendtag{thinking}

\xmltag{agent}

\quad\xmltag{agent_name}DebateAgent\xmlendtag{agent_name}

\quad\xmltag{agent_description}
Near-tie numeric choice using one Debate call.
\xmlendtag{agent_description}

\quad\xmltag{required_arguments}

\qquad\xmltag{agent_input}\xmlendtag{agent_input}

\qquad\xmltag{debate_roles}
["Mathematics Professor", "Statistics Teacher"]
\xmlendtag{debate_roles}

\quad\xmlendtag{required_arguments}

\quad\xmltag{agent_output_id}
compare_agent_output
\xmlendtag{agent_output_id}

\xmlendtag{agent}

\xmltag{answer}compare_agent_output\xmlendtag{answer}\\

More examples:\\

\xmltag{thinking}

Computing 17³ can be done directly, but arithmetic mistakes are easy to make.
Using a Chain-of-Thought with Self-Consistency (CoT\_SC) allows sampling multiple reasoning paths and combining results for accuracy.
The final answer to the original question will be the output of the SCAgent.

\xmlendtag{thinking}

\xmltag{agent}

\quad\xmltag{agent_name}SCAgent\xmlendtag{agent_name}

\quad\xmltag{agent_description}
Performs arithmetic calculations using Chain-of-Thought with Self-Consistency (CoT\_SC).
\xmlendtag{agent_description}

\quad\xmltag{required_arguments}

\qquad\xmltag{agent_input}\xmlendtag{agent_input}

\quad\xmlendtag{required_arguments}

\quad\xmltag{agent_output_id}math_agent_output\xmlendtag{agent_output_id}

\xmlendtag{agent}

\xmltag{answer}math_agent_output\xmlendtag{answer}
\\

Final example: \\

\xmltag{thinking}

This task requires reasoning with a formula and ensuring units are handled correctly.
I will use a Reflexion agent that can reflect on and refine its reasoning if errors occur.
The final answer to the original question will be the output of the ReflexionAgent.

\xmlendtag{thinking}

\xmltag{agent}

\quad\xmltag{agent_name}ReflexionAgent\xmlendtag{agent_name}

\quad\xmltag{agent_description}
Solves reasoning tasks with iterative self-reflection and critique using Reflexion.
\xmlendtag{agent_description}

\quad\xmltag{required_arguments}

\qquad\xmltag{agent_input}\xmlendtag{agent_input}

\quad\xmlendtag{required_arguments}

\quad\xmltag{agent_output_id}reflexion_agent_output\xmlendtag{agent_output_id}

\xmlendtag{agent}

\xmltag{answer}reflexion_agent_output\xmlendtag{answer} \\

If you combine one direct step with a single Reflexion tool call, you will output the following:

\xmltag{thinking}

Step 1: Compute the average manually:
\quad sum = 2 + 3 + 5 + 7 + 11 = 28
\quad avg = 28 / 5 = 5.6

Step 2:
\quad The remaining step — computing sqrt(5.6) to three decimals — requires precision and numeric refinement.
\quad I will delegate that part to a ReflexionAgent using the ReflexionAgent for self-correction if rounding is wrong.
\quad The final answer to the original question will be the output of the ReflexionAgent.

\xmlendtag{thinking}

\xmltag{agent}

\quad\xmltag{agent_name}ReflexionAgent\xmlendtag{agent_name}

\quad\xmltag{agent_description}
Square root with a light self-refine loop (single agent call).
\xmlendtag{agent_description}

\quad\xmltag{required_arguments}

\qquad\xmltag{agent_input}Compute sqrt(5.6) to 3 decimals\xmlendtag{agent_input}

\quad\xmlendtag{required_arguments}

\quad\xmltag{agent_output_id}numeric_agent_output\xmlendtag{agent_output_id}

\xmlendtag{agent}

\xmltag{answer}numeric_agent_output\xmlendtag{answer} \\

Below is the question to solve: 

[QUESTION]

\end{userprompt}

\subsection{High \masness}


\begin{systemprompt}[title={System Prompt}]
You are a helpful assistant.

\masness (degree of MAS): High

Valid Channels: thinking, agent, edge

Model: \texttt{[MODEL]}\\

An agent is a pre-configured AI personalities that can delegate tasks to. Each subagent:
1. Has a specific purpose and expertise area\\
2. Uses its own context window separate from the main conversation\\
3. (Optional) Can be configured with specific tools it's allowed to use\\
4. Includes a custom system prompt that guides its behavior\\
\\
An agent can only call within the \xmltag{thinking} channel, it should be defined in tag \xmltag{agent}. Each agent must contain \xmltag{agent_id}, \xmltag{agent_name}, \xmltag{agent_description}, \xmltag{required_arguments}. To connect multiple agents to form a multi-agent system, use \xmltag{edge} channel.\\

DO NOT MISS ANY REQUEST FIELDS and ensure that your response is a well-formed XML object!
\end{systemprompt}

\begin{developprompt}[title={Develop Prompt}]
Channels: \\
    \xmltag{thinking}: internal reasoning and planning \\
    \xmltag{agent}: definition of sub-agents \\
    \xmltag{answer}: final user-facing answer \\

Model (the model used in sub-agent):

    \texttt{gpt-4.1-nano}: [Introduction of the model from official website] \\
    \texttt{gpt-oss-120b}:[Introduction of the model from official website] \\
    (...list of all candidate models)\\

\masness Levels:

    \texttt{low}: direct solve or at most one agent \\
    \texttt{high}: complex multi-agent delegation\\
    
Sub-agent Schema (all fields required):\\

\xmltag{agent}

\quad\xmltag{agent_name}
...
\xmlendtag{agent_name}
(select one of the agents: CoTAgent, SCAgent, DebateAgent, ReflexionAgent)

\quad\xmltag{agent_description}
...
\xmlendtag{agent_description}

\quad\xmltag{required_arguments}
(make sure all required parameters are set. Must follow XML format)

\qquad\xmltag{...}...\xmlendtag{...}

\qquad\xmltag{...}...\xmlendtag{...}

\quad\xmlendtag{required_arguments}

\quad\xmltag{agent_output_id}
...
\xmlendtag{agent_output_id}

\xmlendtag{agent} \\

Edge Schema (single block; all fields required. Each pair defines a directed link: output of \xmltag{from} → input of \xmltag{to}. List ALL links here; use exactly one \xmltag{edge} block per solution:\\

\xmltag{edge}

\quad\xmltag{from}
...
\xmlendtag{from}
(the source agent\_id)

\quad\xmltag{to}
...
\xmlendtag{to}
(the target agent\_id)

\xmlendtag{edge}\\

Available Agents: \\

\texttt{CoT}: [Description, Name, Required Argument and Summary of Implementation]

\texttt{SC}: [Description, Name, Required Argument and Summary of Implementation]

\texttt{Debate}: [Description, Name, Required Argument and Summary of Implementation]

\texttt{Reflexion}: [Description, Name, Required Argument and Summary of Implementation]

\texttt{Search}: [Description, Name, Required Argument and Summary of Implementation]

\end{developprompt}

\begin{userprompt}[title={User Prompt}]

Please solve the given question by creating one or more agents and connecting them into a valid computational graph that collaboratively produces the final answer. To create an agent, you must define that agent by outputting \xmltag{agent} with \xmltag{agent_id} (a unique id for the agent, must be unique and contain only alphanumeric or underscore characters (e.g., A1, Refine_1, WS_Japan)), \xmltag{agent_name} (exactly one of: CoTAgent, SCAgent, DebateAgent, ReflexionAgent and WebSearchAgent), \xmltag{agent_description}, \xmltag{required_arguments} (must include at least one \xmltag{agent_input} tag. DebateAgents must define \xmltag{debate_roles} with two or more roles. If  \xmltag{agent_input} left empty (""), the parser will automatically replace it with the original question.). \\

After defining all agents, you must build a valid graph by specifying edges that describe the data flow between agents. Output exactly one \xmltag{edge} block, Each \xmltag{from} \xmltag{to} pair connects the output of one agent to the input of another: \\

\xmltag{edge}

\quad\xmltag{from}source_agent_id\xmlendtag{from}

\quad\xmltag{to}target_agent_id\xmlendtag{to}

\xmlendtag{edge} \\

You can output multiple \xmltag{from} and \texttt{\xmltag{/to}} inside the \xmltag{edge}. Each \xmltag{from} and \xmltag{to} value must exactly match an existing \xmltag{agent_id}. \\

To be valid, your graph must satisfy all of the following constraints: 
\begin{itemize}[leftmargin=*,noitemsep,topsep=2pt]
\item Node consistency: Every \xmltag{from} and \xmltag{to} must reference a valid \xmltag{agent_id} that appears in an \xmltag{agent} block.
\item  Directionality: Edges are directed: data flows from \xmltag{from} → \xmltag{to}.
\item  Connectivity: Every agent must be connected directly or indirectly to the main flow. Isolated agents are not allowed.
\item  Start node(s): At least one agent must have no incoming edge. These are “entry points” (e.g., WebSearch or initial reasoning).
\item  Sink node: There must be exactly one agent with no outgoing edge — this is the FINAL agent that produces the answer.
\item  No undefined edges: It is invalid to reference an agent in \xmltag{from} or \xmltag{to} that was not declared.
\item  No loops or cycles: No self-loop: \xmltag{from}X\xmltag{/from}\xmltag{to}X\xmltag{/to} is not allowed. No cycles: The graph must be acyclic; a topological order must exist.
\item  Parallelism allowed: Multiple agents may have the same \xmltag{from} or \xmltag{to} (fan-out/fan-in).
\item  Unambiguous sink: The parser will reject graphs with multiple sinks (add a final “collector” agent if needed).
\item  Order-independent: The XML order of edges does not need to follow execution order; topological sorting is handled automatically.
\item  Sink answer completeness: The unique sink agent's output must directly answer the original question in a user-ready form. It must not be an intermediate artifact (e.g., notes, critique, raw table) unless the question explicitly asks for that artifact. If \xmltag{agent_input} is empty for the sink, it inherits the original question and must return the final answer. If \xmltag{agent_input} is non-empty, the runner still prepends the original question as context; the sink must still produce the final, user-facing answer to that question.
\item Edge-Data Flow Consistency (BIDIRECTIONAL): Edges represent execution order. \$\{\} represents data flow. As a result, you must ensure they are consistent with each other.
    (a) If an \xmltag{agent_input} references \$\{X\}, there MUST be an edge \xmltag{from}X\xmltag{/from}\xmltag{to}THIS AGENT\xmltag{/to}
    (b) If there is an edge \xmltag{from}X\xmltag{/from}\xmltag{to}Y\xmltag{/to}, then Y's \xmltag{agent_input} MUST reference \$\{X\}
    In other words: edges exist if and only if there is actual data passing from one agent to another. Do not create edges solely for execution ordering without data flow.
\end{itemize}

Thinking Section (Required):\\

Before defining agents and edges, you must include a \xmltag{thinking} section.
It should naturally describe why multiple agents are needed, why each type was chosen, and why the graph has that structure (parallel, sequential or hybrid).
It must justify both planning and design rationale.\\

Example structure:\\

\xmltag{thinking}
  Explain why a single agent is insufficient.
  Describe each agent's role and how they connect.
  Justify the flow pattern (parallel, sequential, hybrid).
  End by clearly stating which agent produces the final output.
\xmltag{/thinking}\\

Single-agent example:\\

If you decide to solve via single agent, you will output the following. In this case, since the \xmltag{agent_input} is the same as the original task, you must set the \xmltag{agent_input} as empty (""), and the parser will replace it with the original question.\\

1st example:\\

Question: Compute the definite integral of (2x + 5) dx from 0 to 3.\\

\xmltag{thinking}

This problem involves symbolic integration and applying the Fundamental Theorem of Calculus.
It requires structured reasoning rather than simple numeric computation.
I will use a calculus agent that can perform step-by-step Chain-of-Thought reasoning.
The final answer to the original question will be the output of the CoTAgent.

\xmlendtag{thinking}

\xmltag{agent}

\quad\xmltag{agent_id}calc\_agent\xmlendtag{agent_id}

\quad\xmltag{agent_name}CoTAgent\xmlendtag{agent_name}

\quad\xmltag{agent_description}
Definite integrals with one Chain-of-Thought call.
\xmlendtag{agent_description}

\quad\xmltag{required_arguments}

\qquad\xmltag{agent_input}\xmlendtag{agent_input}

\quad\xmlendtag{required_arguments}

\xmlendtag{agent}\\

2nd example:\\

Question: What is 17 cubed?\\

\xmltag{thinking}

Computing 17³ can be done directly, but arithmetic mistakes are easy to make.
Using a Chain-of-Thought with Self-Consistency (CoT\_SC) allows sampling multiple reasoning paths and combining results for accuracy.
The final answer to the original question will be the output of the SCAgent.

\xmlendtag{thinking}

\xmltag{agent}

\quad\xmltag{agent_id}math\_agent\xmlendtag{agent_id}

\quad\xmltag{agent_name}SCAgent\xmlendtag{agent_name}

\quad\xmltag{agent_description}
Performs arithmetic calculations using Chain-of-Thought with Self-Consistency (CoT\_SC).
\xmlendtag{agent_description}

\quad\xmltag{required_arguments}

\qquad\xmltag{agent_input}\xmlendtag{agent_input}

\quad\xmlendtag{required_arguments}

\xmlendtag{agent} \\

3rd examples: \\

Question: Given x² = 46.694444, which target number is closer, 45 or 46? \\

\xmltag{thinking}

This question requires comparison between two close numeric choices.
To ensure accuracy, I will let two reasoning roles debate — one focusing on mathematical precision and the other on practical rounding.
The DebateAgent can capture both perspectives and reach a justified final answer.
The final answer will be the output of the DebateAgent.

\xmlendtag{thinking}

\xmltag{agent}

\quad\xmltag{agent_id}compare\_agent\xmlendtag{agent_id}

\quad\xmltag{agent_name}DebateAgent\xmlendtag{agent_name}

\quad\xmltag{agent_description}
Near-tie numeric choice using one Debate call.
\xmlendtag{agent_description}

\quad\xmltag{required_arguments}

\qquad\xmltag{agent_input}\xmlendtag{agent_input}

\qquad\xmltag{debate_roles}
["Mathematics Professor", "Statistics Teacher"]
\xmlendtag{debate_roles}

\quad\xmlendtag{required_arguments}

\xmlendtag{agent}

4th example: \\

Question: A train travels 60 miles per hour. How far does it go in 2.5 hours? \\

\xmltag{thinking}

This task requires reasoning with a formula and ensuring units are handled correctly.
I will use a Reflexion agent that can reflect on and refine its reasoning if errors occur.
The final answer to the original question will be the output of the ReflexionAgent.

\xmlendtag{thinking}

\xmltag{agent}

\quad\xmltag{agent_id}reflexion\_agent\xmlendtag{agent_id}

\quad\xmltag{agent_name}ReflexionAgent\xmlendtag{agent_name}

\quad\xmltag{agent_description}
Solves reasoning tasks with iterative self-reflection and critique using Reflexion.
\xmlendtag{agent_description}

\quad\xmltag{required_arguments}

\qquad\xmltag{agent_input}\xmlendtag{agent_input}

\quad\xmlendtag{required_arguments}

\xmlendtag{agent}\\

5th example:\\

Question: What is the current inflation rate in Japan as of this month?\\

\xmltag{thinking}

This question depends on up-to-date factual information that cannot be reliably recalled from static knowledge.  
Using a single WebSearchAgent is sufficient because the task only requires retrieving accurate, cited facts from the web, not further reasoning or synthesis.  
The agent will search online sources and return a concise, citation-based summary.  
The WebSearchAgent is therefore both the only node and the final output of the flow.

\xmlendtag{thinking}

\xmltag{agent}

\quad\xmltag{agent_id}SEARCH\xmlendtag{agent_id}

\quad\xmltag{agent_name}WebSearchAgent\xmlendtag{agent_name}

\quad\xmltag{agent_description}
Retrieves recent and cited factual information from the internet.
\xmlendtag{agent_description}

\quad\xmltag{required_arguments}

\qquad\xmltag{agent_input}\xmlendtag{agent_input}

\quad\xmlendtag{required_arguments}

\xmlendtag{agent}\\

Multi-Agent Example: \\

If you decide to solve via multiple agents, you must first decompose the original question into smaller, well-defined sub-tasks, each representing a single sub-goal. Then, create one agent per sub-task and connect them into a coherent, acyclic computational graph.\\

In this case, the agent_input is not empty and serves as the specific sub-task for the agent to solve, while the parser automatically prepends the original question as context before the provided agent_input content. \\

When decomposing:\\

1. Keep sub-tasks minimal and focused. Each agent should handle one atomic objective (e.g., one query, one reasoning step, or one verification task). \\
2. Use multiple WebSearchAgents for different pieces of factual evidence, rather than a single broad search.\\
3. Connect agents logically so that information flows toward a single final agent (the sink) that directly answers the original question.\\

1st example:\\

Question: Give a short, cited summary of the most recent housing vacancy rates for New York City, Los Angeles, and Chicago\\

\xmltag{thinking}

We need up-to-date numbers with sources, so we will run three WebSearch agents in parallel.
Then a CoT agent will normalize the three snippets into a small table.
An SC agent will run small variants and vote to reduce extraction errors.
A Reflexion agent will check units, recency, and citations.
A final CoT agent will write the short summary.
The final sink is FINAL.

\xmlendtag{thinking}

\xmltag{agent}

\quad\xmltag{agent_id}WS\_NYC\xmlendtag{agent_id}

\quad\xmltag{agent_name}WebSearchAgent\xmlendtag{agent_name}

\quad\xmltag{agent_description}
Find the latest NYC housing vacancy rate with source text.
\xmlendtag{agent_description}

\quad\xmltag{required_arguments}

\qquad\xmltag{agent_input}
Search for the latest official or reputable estimate of the housing vacancy rate for New York City. Return a short snippet with the number, date, and a citation line.
\xmlendtag{agent_input}

\quad\xmlendtag{required_arguments}

\xmlendtag{agent}

\xmltag{agent}

\quad\xmltag{agent_id}WS\_LA\xmlendtag{agent_id}

\quad\xmltag{agent_name}WebSearchAgent\xmlendtag{agent_name}

\quad\xmltag{agent_description}
Find the latest LA vacancy rate with source text.
\xmlendtag{agent_description}

\quad\xmltag{required_arguments}

\qquad\xmltag{agent_input}
Search for the latest official or reputable estimate of the housing vacancy rate for Los Angeles. Return a short snippet with the number, date, and a citation line.
\xmlendtag{agent_input}

\quad\xmlendtag{required_arguments}

\xmlendtag{agent}

\xmltag{agent}

\quad\xmltag{agent_id}WS\_CHI\xmlendtag{agent_id}

\quad\xmltag{agent_name}WebSearchAgent\xmltag{agent_name}

\quad\xmltag{agent_description}
Find the latest Chicago vacancy rate with source text.
\xmlendtag{agent_description}

\quad\xmltag{required_arguments}

\qquad\xmltag{agent_input}
Search for the latest official or reputable estimate of the housing vacancy rate for Chicago. Return a short snippet with the number, date, and a citation line.
\xmlendtag{agent_input}

\quad\xmlendtag{required_arguments}

\xmlendtag{agent}

\xmltag{agent}

\quad\xmltag{agent_id}EXT\xmlendtag{agent_id}

\quad\xmltag{agent_name}CoTAgent\xmlendtag{agent_name}

\quad\xmltag{agent_description}
Extract numbers and standardize the three rates with dates and citations.
\xmlendtag{agent_description}

\quad\xmltag{required_arguments}

\qquad\xmltag{agent_input}
From the snippets below, extract for each city: city name, vacancy rate (as a percent), reference date (YYYY-MM or YYYY), and a short citation. Output a compact 3-line table.

NYC:
\$\{WS\_NYC\}

LA:
\$\{WS\_LA\}

Chicago:
\$\{WS\_CHI\}
\xmlendtag{agent_input}

\quad\xmlendtag{required_arguments}

\xmlendtag{agent}

\xmltag{agent}

\quad\xmltag{agent_id}VOTE\xmlendtag{agent_id}

\quad\xmltag{agent_name}SCAgent\xmlendtag{agent_name}

\quad\xmltag{agent_description}
Ensemble the extraction to reduce copy or parse errors.
\xmlendtag{agent_description}

\quad\xmltag{required_arguments}

\qquad\xmltag{agent_input}
Given the 3-line table below, produce 5 independent extractions and vote on a single corrected 3-line table.

Table:
\$\{EXT\}
\xmlendtag{agent_input}

\quad\xmlendtag{required_arguments}

\xmlendtag{agent}

\xmltag{agent}

\quad\xmltag{agent_id}QA\xmlendtag{agent_id}

\quad\xmltag{agent_name}ReflexionAgent\xmlendtag{agent_name}

\quad\xmltag{agent_description}
Check units, date freshness, and cite presence; list fixes if needed.
\xmlendtag{agent_description}

\quad\xmltag{required_arguments}

\qquad\xmltag{agent_input}
Audit the voted table for: units as \%, dates present, and a citation per city. If any issue is found, list concrete fixes in 3 lines; else say OK. End with a 1-line verdict.

Voted table:
\$\{VOTE\}
\xmlendtag{agent_input}

\quad\xmlendtag{required_arguments}

\xmlendtag{agent}

\xmltag{agent}

\quad\xmltag{agent_id}FINAL\xmlendtag{agent_id}

\quad\xmltag{agent_name}CoTAgent\xmlendtag{agent_name}

\quad\xmltag{agent_description}
Write the short, cited summary.
\xmlendtag{agent_description}

\quad\xmltag{required_arguments}

\qquad\xmltag{agent_input}
Using the checked table and notes, write a 3--4 sentence summary with one sentence per city and a final sentence comparing the rates. Keep citations as inline source lines from the table.

Table:
\$\{VOTE\}

QA notes:
\$\{QA\}
\xmlendtag{agent_input}

\quad\xmlendtag{required_arguments}

\xmlendtag{agent}

\xmltag{edge}

\quad\xmltag{from}WS\_NYC\xmlendtag{from}\xmltag{to}EXT\xmlendtag{to}

\quad\xmltag{from}WS\_LA\xmlendtag{from}\xmltag{to}EXT\xmlendtag{to}

\quad\xmltag{from}WS\_CHI\xmlendtag{from}\xmlendtag{from}\xmltag{to}EXT\xmlendtag{to}

\quad\xmltag{from}EXT\xmlendtag{from}\xmltag{to}VOTE\xmlendtag{to}

\quad\xmltag{from}VOTE\xmlendtag{from}\xmltag{to}QA\xmlendtag{to}

\quad\xmltag{from}VOTE\xmlendtag{from}\xmltag{to}FINAL\xmlendtag{to}

\quad\xmltag{from}QA\xmlendtag{from}\xmltag{to}FINAL\xmlendtag{to}

\xmlendtag{edge}\\

2nd example:\\

Question: During Pope John Paul II's first foreign journey in the late 1970s, he visited a country known for its rich Mesoamerican history and home to a large population. On which other date did he visit a major city on the Adriatic Sea, known for its significant port and a famous basilica dedicated to a saint with the initial “S”, and which other nearby city did he visit on the same day?\\

\xmltag{thinking}

The question needs historical reasoning: identify Pope John Paul II's first foreign trip, find an Adriatic city he visited with a basilica of a saint starting with “S”, get the date, and determine another city visited the same day.

One agent cannot do all of this because it mixes retrieval and reasoning.
I will decompose it into smaller sub-tasks: three WebSearchAgents for each factual lookup, one CoTAgent to build a timeline, one ReflexionAgent to verify consistency, and a final CoTAgent to write the answer.
The FINAL agent outputs the final answer.

\xmlendtag{thinking}

\xmltag{agent}

\quad\xmltag{agent_id}WS\_FIRST\_TRIP\xmlendtag{agent_id}

\quad\xmltag{agent_name}WebSearchAgent\xmlendtag{agent_name}

\quad\xmltag{agent_description}
Retrieve Pope John Paul II's first foreign trip details in the late 1970s.
\xmlendtag{agent_description}

\quad\xmltag{required_arguments}

\qquad\xmltag{agent_input}
Search for Pope John Paul II's first foreign journey (year, destination country, and duration).
Return the trip date range, destination, and a reliable citation.
\xmlendtag{agent_input}

\quad\xmlendtag{required_arguments}

\xmlendtag{agent}

\xmltag{agent}

\quad\xmltag{agent_id}WS\_ADRIATIC\xmlendtag{agent_id}

\quad\xmltag{agent_name}WebSearchAgent\xmlendtag{agent_name}

\quad\xmltag{agent_description}
Find Adriatic Sea city visits with basilicas dedicated to saints starting with `S'.
\xmlendtag{agent_description}

\quad\xmltag{required_arguments}

\qquad\xmltag{agent_input}
Search for any Adriatic city visited by Pope John Paul II that has a basilica dedicated to a saint with the initial `S'
(e.g., Saint Nicholas, Saint Mark).
Return the city name, basilica name, and visit date, with citation.
\xmlendtag{agent_input}

\quad\xmlendtag{required_arguments}

\xmlendtag{agent}

\xmltag{agent}

\quad\xmltag{agent_id}WS\_SAME\_DAY\xmlendtag{agent_id}

\quad\xmltag{agent_name}WebSearchAgent\xmlendtag{agent_name}

\quad\xmltag{agent_description}
Identify any other nearby city visited by Pope John Paul II on the same day as the Adriatic visit.
\xmlendtag{agent_description}

\quad\xmltag{required_arguments}

\qquad\xmltag{agent_input}
Search for other cities visited by Pope John Paul II on the same date as his Adriatic visit.
Return the nearby city name, distance estimate, and source citation.
\xmlendtag{agent_input}

\quad\xmlendtag{required_arguments}

\xmlendtag{agent}

\xmltag{agent}

\quad\xmltag{agent_id}TIMELINE\xmlendtag{agent_id}

\quad\xmltag{agent_name}CoTAgent\xmlendtag{agent_name}

\quad\xmltag{agent_description}
Integrate trip data into a single verified historical timeline.
\xmlendtag{agent_description}

\quad\xmltag{required_arguments}

\qquad\xmltag{agent_input}
Using the results below:

First foreign journey:
\$\{WS\_FIRST\_TRIP\}

Adriatic visit:
\$\{WS\_ADRIATIC\}

Same-day visit:
\$\{WS\_SAME\_DAY\}

Build a clear timeline confirming the Adriatic visit date and same-day nearby city.
\xmlendtag{agent_input}

\quad\xmlendtag{required_arguments}

\xmlendtag{agent}

\xmltag{agent}

\quad\xmltag{agent_id}VERIFY\xmlendtag{agent_id}

\quad\xmltag{agent_name}ReflexionAgent\xmlendtag{agent_name}

\quad\xmltag{agent_description}
Validate chronology, geography, and saint-basilica link.
\xmlendtag{agent_description}

\quad\xmltag{required_arguments}

\qquad\xmltag{agent_input}
Check consistency among timeline facts:
- Ensure the same date appears across all sources.
- Confirm the Adriatic city is geographically near the second city.
- Verify that the basilica indeed honors a saint whose name starts with `S'.

Return OK if consistent, otherwise list corrections.

Timeline to verify:
\$\{TIMELINE\}
\xmlendtag{agent_input}

\quad\xmlendtag{required_arguments}

\xmlendtag{agent}

\xmltag{agent}

\quad\xmltag{agent_id}FINAL\xmlendtag{agent_id}

\quad\xmltag{agent_name}CoTAgent\xmlendtag{agent_name}

\quad\xmltag{agent_description}
Produce the final concise answer with the date and both cities.
\xmlendtag{agent_description}

\quad\xmltag{required_arguments}

\qquad\xmltag{agent_input}
Using verified results, answer in one sentence:

``Pope John Paul II visited [ADRIATIC\_CITY] on [DATE], home to the Basilica of Saint [S], and also visited [NEARBY\_CITY] on the same day.''

Include a brief verification sentence citing sources.

Timeline:
\$\{TIMELINE\}

Verification:
\$\{VERIFY\}
\xmlendtag{agent_input}

\quad\xmlendtag{required_arguments}

\xmlendtag{agent}

\xmltag{edge}

\quad\xmltag{from}WS\_FIRST\_TRIP\xmlendtag{from}\xmltag{to}TIMELINE\xmlendtag{to}

\quad\xmltag{from}WS\_ADRIATIC\xmlendtag{from}\xmltag{to}TIMELINE\xmlendtag{to}

\quad\xmltag{from}WS\_SAME\_DAY\xmlendtag{from}\xmltag{to}TIMELINE\xmlendtag{to}

\quad\xmltag{from}TIMELINE\xmlendtag{from}\xmltag{to}VERIFY\xmlendtag{to}

\quad\xmltag{from}VERIFY\xmlendtag{from}\xmltag{to}FINAL\xmlendtag{to}

\quad\xmltag{from}TIMELINE\xmlendtag{from}\xmltag{to}FINAL\xmlendtag{to}

\xmlendtag{edge}\\

(... more examples) \\

Below is the question to solve: 

[QUESTION]

\end{userprompt}

\subsection{Sub-agents}
\label{app:sub_agents}


\begin{subagent}[title={CoTAgent}]

\pykw{async} \pykw{def} \pyfn{CoTAgent}(\pykw{self}, agent\_input, model: \pykw{str}):

\quad \pycom{\# Basic setting}

\quad temperature = 0.5

\quad \pycom{\# Chain-of-Thought instruction}

\quad cot\_instruction = \pystr{"Please think step by step and then solve the task."}

\quad \pycom{\# Instantiate CoT LLM}

\quad cot\_agent = LLMAgentBase([\pystr{thinking}, \pystr{answer}],\pystr{Chain-of-Thought LLM}, model=model, temperature=temperature)

\quad thinking, answer = \pykw{await} cot\_agent([agent\_input], cot\_instruction)

\quad final\_answer = self.make\_final\_answer(thinking, answer)

\quad \pykw{return} final\_answer \\

func_string = inspect.getsource(CoTAgent) \\

COT = \code{\{}

\quad \jsonkey{description}: 
\jsonstr{By encouraging the LLM to think step by step rather than directly outputting an answer, chain-of-thought reasoning enables complex problem-solving through intermediate steps.}

\quad \jsonkey{name}: \jsonstr{Chain-of-Thought Agent (CoTAgent)}

\quad \jsonkey{required_arguments}: \code{\{}

\quad\quad \jsonkey{agent_input}: 
\jsonstr{The input for the CoTAgent. If empty, the parser replaces it with the original question.}

\quad \code{\}}

\quad \jsonkey{implementation}: \jsonstr{\{func\_string\}}

\code{\}}

\end{subagent}

\begin{subagent}[title={SCAgent}]
    \pykw{async} \pykw{def} \pyfn{SCAgent}(\pykw{self}, agent\_input, model: \pykw{str}):

\quad \pycom{\# Basic setting}

\quad temperature = 0.5

\quad num\_repeated\_samples = 5

\quad \pycom{\# Chain-of-Thought instruction}

\quad cot\_instruction = \pystr{"Please think step by step and then solve the task."}

\quad \pycom{\# Initialize multiple CoT agents for self-consistency}

\quad cot\_agents = [
\quad\quad LLMAgentBase([\pystr{thinking}, \pystr{answer}],
\quad\quad \pystr{Chain-of-Thought LLM},
\quad\quad model=model,
\quad\quad temperature=temperature)
\quad\quad \pykw{for} \_ \pykw{in} range(num\_repeated\_samples)
\quad ]

\quad thinking\_mapping = \{\}

\quad answer\_mapping = \{\}

\quad possible\_answers = []

\quad \pykw{for} i \pykw{in} range(num\_repeated\_samples):

\quad\quad thinking, answer = \pykw{await} cot\_agents[i]([agent\_input], cot\_instruction)

\quad\quad possible\_answers.append(answer.content)

\quad\quad thinking\_mapping[answer.content] = thinking

\quad\quad answer\_mapping[answer.content] = answer

\quad \pycom{\# Ensembling answers via majority voting}

\quad answer = self.majority\_voting(possible\_answers)

\quad thinking = thinking\_mapping[answer]

\quad answer = answer\_mapping[answer]

\quad final\_answer = self.make\_final\_answer(thinking, answer)

\quad \pykw{return} final\_answer \\

func\_string = inspect.getsource(SCAgent) \\

COT\_SC = \code{\{}

\quad \jsonkey{description}:
\jsonstr{While an LLM can arrive at the correct answer, its reasoning may vary. By repeatedly asking the same question with higher temperature settings, multiple reasoning paths are generated. These answers from Chain-of-Thought agents are then combined through ensembling to produce a more accurate final answer. This approach is best suited for problems requiring high confidence through consensus.}

\quad \jsonkey{name}:
\jsonstr{Self-Consistency with Chain-of-Thought (SCAgent)}

\quad \jsonkey{required_arguments}: \code{\{}

\quad\quad \jsonkey{agent_input}:
\jsonstr{The input for the SCAgent. If empty (""), the parser automatically replaces it with the original question.}

\quad \code{\}}

\quad \jsonkey{implementation}:
\jsonstr{\{func\_string\}}

\code{\}}

\end{subagent}

\begin{subagent}[title={DebateAgent}]
\pykw{async} \pykw{def} \pyfn{DebateAgent}(\pykw{self}, agent\_input, model: \pykw{str}, debate\_roles: List[\pykw{str}]):

\quad \pycom{\# Basic setting}

\quad temperature = 0.5

\quad max\_debate\_round = 5

\quad \pycom{\# Instruction for initial reasoning}

\quad debate\_initial\_instruction = 
\quad \pystr{"Please think step by step and then solve the task."}

\quad \pycom{\# Instruction for debate updates}

\quad debate\_instruction = 
\quad \pystr{"Given solutions from other agents, consider their opinions as advice and provide an updated answer."}

\quad \pycom{\# Initialize debate agents with different roles}

\quad debate\_agents = [
\quad\quad LLMAgentBase([\pystr{thinking}, \pystr{answer}],
\quad\quad \pystr{Debate LLM},
\quad\quad model=model,
\quad\quad role=role,
\quad\quad temperature=temperature)
\quad\quad \pykw{for} role \pykw{in} debate\_roles
\quad ]

\quad \pycom{\# Instruction for final decision}

\quad final\_decision\_instruction =
\quad \pystr{"Given all reasoning and answers, carefully provide a final answer."}

\quad final\_decision\_agent = LLMAgentBase(
\quad\quad [\pystr{thinking}, \pystr{answer}],
\quad\quad \pystr{Final Decision LLM},
\quad\quad model=model,
\quad\quad temperature=temperature
\quad )

\quad all\_thinking = [[] \pykw{for} \_ \pykw{in} range(max\_debate\_round)]

\quad all\_answer = [[] \pykw{for} \_ \pykw{in} range(max\_debate\_round)]

\quad \pycom{\# Perform debate rounds}

\quad \pykw{for} r \pykw{in} range(max\_debate\_round):

\quad\quad \pykw{for} i \pykw{in} range(len(debate\_agents)):

\quad\quad\quad \pykw{if} r == 0:

\quad\quad\quad\quad thinking, answer = 
\quad\quad\quad\quad \pykw{await} debate\_agents[i]([agent\_input], debate\_initial\_instruction)

\quad\quad\quad \pykw{else}:

\quad\quad\quad\quad input\_infos = 
\quad\quad\quad\quad [agent\_input] 
\quad\quad\quad\quad + [all\_thinking[r-1][i]] 
\quad\quad\quad\quad + all\_thinking[r-1][:i] 
\quad\quad\quad\quad + all\_thinking[r-1][i+1:]

\quad\quad\quad\quad thinking, answer = 
\quad\quad\quad\quad \pykw{await} debate\_agents[i](input\_infos, debate\_instruction)

\quad\quad\quad all\_thinking[r].append(thinking)

\quad\quad\quad all\_answer[r].append(answer)

\quad \pycom{\# Final decision based on all debate results}

\quad thinking, answer = 
\quad \pykw{await} final\_decision\_agent(
\quad\quad [agent\_input] 
\quad\quad + all\_thinking[max\_debate\_round-1] 
\quad\quad + all\_answer[max\_debate\_round-1],
\quad\quad final\_decision\_instruction
\quad )

\quad final\_answer = self.make\_final\_answer(thinking, answer)

\quad \pykw{return} final\_answer \\

func\_string = inspect.getsource(DebateAgent) \\

LLM\_debate = \code{\{}

\quad \jsonkey{description}:
\jsonstr{By letting different LLMs debate with each other, diverse perspectives are leveraged to reach better solutions. This agent is best suited for problems that benefit from multiple viewpoints.}

\quad \jsonkey{name}:
\jsonstr{LLM Debate (DebateAgent)}

\quad \jsonkey{required_arguments}: \code{\{}

\quad\quad \jsonkey{agent_input}:
\jsonstr{The input for the DebateAgent. If empty (""), the parser automatically replaces it with the original question.}

\quad\quad \jsonkey{debate_roles}:
\jsonstr{A list of roles (must include more than one) representing distinct perspectives, such as "Mathematics Professor" or "Statistician".}

\quad \code{\}}

\quad \jsonkey{implementation}:
\jsonstr{\{func\_string\}}

\code{\}}

\end{subagent}

\begin{subagent}[title=ReflexionAgent]
\pykw{async} \pykw{def} \pyfn{ReflexionAgent}(\pykw{self}, agent\_input, model: \pykw{str}):

\quad \pycom{\# Basic setting}

\quad temperature = 0.5

\quad max\_reflection\_round = 5

\quad \pycom{\# Instruction for initial reasoning}

\quad initial\_instruction = 
\quad \pystr{"Please think step by step and then solve the task."}

\quad \pycom{\# Instruction for reflection and refinement}

\quad reflect\_instruction = 
\quad \pystr{"Given previous attempts and feedback, carefully consider where you could go wrong in your latest attempt. Using insights from previous attempts, try to solve the task better."}

\quad cot\_agent = LLMAgentBase(
\quad\quad [\pystr{thinking}, \pystr{answer}],
\quad\quad \pystr{Chain-of-Thought LLM},
\quad\quad model=model,
\quad\quad temperature=temperature
\quad )

\quad \pycom{\# Instruction for critic feedback}

\quad critic\_instruction =
\quad \pystr{"Please review the answer above and criticize where it might be wrong. If you are absolutely sure it is correct, output exactly 'True' in 'correct'."}

\quad critic\_agent = LLMAgentBase(
\quad\quad [\pystr{feedback}, \pystr{correct}],
\quad\quad \pystr{Critic LLM},
\quad\quad model=model,
\quad\quad temperature=temperature
\quad )

\quad \pycom{\# Initial attempt}

\quad cot\_inputs = [agent\_input]

\quad thinking, answer = 
\quad \pykw{await} cot\_agent(cot\_inputs, initial\_instruction, 0)

\quad \pycom{\# Iterative self-reflection loop}

\quad \pykw{for} i \pykw{in} range(max\_reflection\_round):

\quad\quad feedback, correct = 
\quad\quad \pykw{await} critic\_agent(
\quad\quad\quad [agent\_input, thinking, answer],
\quad\quad\quad critic\_instruction,
\quad\quad\quad i
\quad\quad )

\quad\quad \pykw{if} correct.content == \pystr{'True'}:

\quad\quad\quad \pykw{break}

\quad\quad cot\_inputs.extend([thinking, answer, feedback])

\quad\quad thinking, answer =
\quad\quad \pykw{await} cot\_agent(cot\_inputs, reflect\_instruction, i + 1)

\quad final\_answer = self.make\_final\_answer(thinking, answer)

\quad \pykw{return} final\_answer \\

func\_string = inspect.getsource(ReflexionAgent) \\

Reflexion = \code{\{}

\quad \jsonkey{description}:
\jsonstr{To enhance performance, an LLM can iteratively improve its answer based on feedback. By reflecting on previous attempts and incorporating critique, the model refines its reasoning and produces a more accurate solution. This agent is best suited for complex problems that benefit from self-correction.}

\quad \jsonkey{name}:
\jsonstr{Self-Refine (Reflexion)}

\quad \jsonkey{required_arguments}: \code{\{}

\quad\quad \jsonkey{agent_input}:
\jsonstr{The input for the ReflexionAgent. If empty (""), the parser automatically replaces it with the original question.}

\quad \code{\}}

\quad \jsonkey{implementation}:
\jsonstr{\{func\_string\}}

\code{\}}

\end{subagent}

%% file: sections/appendix/parser.tex
\section{Rule-based Parser}
\label{app:parser}

\subsection{Low \masness}

\begin{subagent}[title=Low \masness Parser]

\pykw{async} \pykw{def} \pyfn{LowDoMParser}(
\pykw{self}, response\_text, original\_task\_info):

\quad \pycom{\# Step 1: Check whether an agent is specified}

\quad agent\_name = extract\_xml(response\_text, \pystr{agent\_name})

\quad \pykw{if} \pykw{not} agent\_name:

\quad\quad \pycom{\# No agent: fall back to direct answer}

\quad\quad answer = extract\_xml(response\_text, \pystr{answer})

\quad\quad \pykw{return} \pystr{direct\_answer}, answer \\

\quad \pycom{\# Step 2: Parse required arguments}

\quad args\_xml = extract\_xml(response\_text, \pystr{required\_arguments})

\quad required\_args = \code{\{\}} \\

\quad \pycom{\# agent\_input handling}

\quad agent\_input = extract\_xml(args\_xml, \pystr{agent\_input})

\quad \pykw{if} agent\_input == \pystr{""} \pykw{or} global\_no\_decompose:

\quad\quad required\_args[\pystr{agent\_input}] = \pystr{""}

\quad \pykw{else}:

\quad\quad required\_args[\pystr{agent\_input}] = agent\_input \\

\quad \pycom{\# Optional DebateAgent arguments}

\quad debate\_roles = extract\_xml(args\_xml, \pystr{debate\_roles})

\quad \pykw{if} agent\_name.lower() == \pystr{debateagent}:

\quad\quad roles = parse\_list(debate\_roles)

\quad\quad \pykw{assert} len(roles) $\geq$ 2

\quad\quad required\_args[\pystr{debate\_roles}] = roles \\

\quad \pycom{\# Step 3: Construct agent call}

\quad agent\_call = \code{\{}

\quad\quad \jsonkey{name}: agent\_name,

\quad\quad \jsonkey{required\_arguments}: required\_args

\quad \code{\}} \\

\quad \pycom{\# Step 4: Generate unified forward() dispatcher}

\quad \pykw{async} \pykw{def} \pyfn{forward}(
\pykw{self}, original\_task\_info):

\quad\quad agent\_input = required\_args[\pystr{agent\_input}] \\

\quad\quad \pykw{if} agent\_input == \pystr{""}:

\quad\quad\quad task\_info = original\_task\_info

\quad\quad \pykw{else}:

\quad\quad\quad task\_info = CombineTask(original\_task\_info,
agent\_input) \\

\quad\quad \pycom{\# Dispatch based on agent type}

\quad\quad \pykw{if} agent\_name == \pystr{SCAgent}:

\quad\quad\quad \pykw{return} \pykw{await} self.SCAgent(task\_info, model=global\_node\_model)

\quad\quad \pykw{elif} agent\_name == \pystr{CoTAgent}:

\quad\quad\quad \pykw{return} \pykw{await} self.CoTAgent(task\_info, model=global\_node\_model)

\quad\quad \pykw{elif} agent\_name == \pystr{ReflexionAgent}:

\quad\quad\quad \pykw{return} \pykw{await} self.ReflexionAgent(task\_info, model=global\_node\_model)

\quad\quad \pykw{elif} agent\_name == \pystr{DebateAgent}:

\quad\quad\quad \pykw{return} \pykw{await} self.DebateAgent(task\_info, model=global\_node\_model, debate\_roles=required\_args[\pystr{debate\_roles}])

\quad \pykw{return} forward 
\end{subagent}

\subsection{High \masness}


\begin{subagent}[title=High DoM Parser]

\pykw{async} \pykw{def} \pyfn{HighDoMParser}(
\pykw{self}, response\_text, original\_task\_info):

\quad \pycom{\# Step 1: Extract all agent nodes}

\quad agents = ExtractAllAgents(response\_text)

\quad \pykw{if} \pykw{not} agents:

\quad\quad \pycom{\# No agents: fall back to direct answer}

\quad\quad answer = extract\_xml(response\_text, \pystr{answer})

\quad\quad \pykw{return} \pystr{direct\_answer}, answer \\

\quad \pycom{\# Step 2: Extract edges (directed dependencies)}

\quad edges = ExtractEdges(response\_text) \\

\quad \pycom{\# Step 3: Handle the single-agent case}

\quad \pykw{if} len(agents) == 1 \pykw{and} len(edges) == 0:

\quad\quad \pykw{return} GenerateSingleAgentForward(agents[0]) \\

\quad \pycom{\# Step 4: Reject invalid multi-agent case with missing edges}

\quad \pykw{if} len(agents) $>$ 1 \pykw{and} len(edges) == 0:

\quad\quad \pykw{raise} ValueError(\pystr{multiple agents require edges}) \\

\quad \pycom{\# Step 5: Validate that the graph is a DAG with exactly one sink}

\quad ValidateGraph(agents, edges) \\

\quad \pycom{\# Step 6: Compute execution order and sink node}

\quad execution\_order = TopologicalSort(agents, edges)

\quad sink = FindUniqueSink(agents, edges) \\

\quad \pycom{\# Step 7: Generate multi-agent forward() with graph execution}

\quad \pykw{return} GenerateMultiAgentForward(agents, edges, execution\_order, sink) \\

\pykw{def} \pyfn{ExtractAllAgents}(response\_text):

\quad agents = []

\quad \pycom{\# Find all \texttt{<agent>...</agent>} blocks}

\quad \pykw{for} block \pykw{in} FindAllAgentBlocks(response\_text):

\quad\quad agent\_id = extract\_xml(block, \pystr{agent\_id})

\quad\quad agent\_name = extract\_xml(block, \pystr{agent\_name})

\quad\quad agent\_desc = extract\_xml(block, \pystr{agent\_description})

\quad\quad args\_xml = extract\_xml(block, \pystr{required\_arguments})

\quad\quad \pykw{if} \pykw{not} agent\_id \pykw{or} \pykw{not} agent\_name:

\quad\quad\quad \pykw{continue} \\

\quad\quad required\_args = \code{\{\}}

\quad\quad agent\_input = extract\_xml(args\_xml, \pystr{agent\_input})

\quad\quad \pykw{if} agent\_input == \pystr{""} \pykw{or} \pykw{not} agent\_input:

\quad\quad\quad required\_args[\pystr{agent\_input}] = \pystr{""}

\quad\quad \pykw{else}:

\quad\quad\quad required\_args[\pystr{agent\_input}] = agent\_input \\

\quad\quad debate\_roles = extract\_xml(args\_xml, \pystr{debate\_roles})

\quad\quad \pykw{if} debate\_roles:

\quad\quad\quad roles = ParseList(debate\_roles)

\quad\quad\quad \pykw{if} agent\_name.lower() == \pystr{debateagent}:

\quad\quad\quad\quad \pykw{assert} roles \pykw{is} \pykw{not} None

\quad\quad\quad\quad \pykw{assert} len(roles) $\geq$ 2

\quad\quad\quad\quad \pykw{assert} \pykw{all}(isinstance(r, \pykw{str}) \pykw{and} len(r) $\geq$ 2 \pykw{for} r \pykw{in} roles)

\quad\quad\quad\quad required\_args[\pystr{debate\_roles}] = roles

\quad\quad\quad \pykw{else}:

\quad\quad\quad\quad required\_args[\pystr{debate\_roles}] = roles \\

\quad\quad agents.append(\code{\{}

\quad\quad\quad \jsonkey{agent\_id}: agent\_id,

\quad\quad\quad \jsonkey{agent\_name}: agent\_name,

\quad\quad\quad \jsonkey{agent\_description}: agent\_desc,

\quad\quad\quad \jsonkey{required\_arguments}: required\_args

\quad\quad \code{\}})

\quad \pykw{return} agents \\

\pykw{def} \pyfn{ExtractEdges}(response\_text):

\quad edges = []

\quad edge\_xml = extract\_xml(response\_text, \pystr{edge})

\quad \pykw{if} \pykw{not} edge\_xml:

\quad\quad \pykw{return} edges \\

\quad \pycom{\# Read each (from,to) pair in order}

\quad \pykw{for} (u, v) \pykw{in} FindAllFromToPairs(edge\_xml):

\quad\quad edges.append((u.strip(), v.strip()))

\quad \pykw{return} edges \\

\pykw{def} \pyfn{ValidateGraph}(agents, edges):

\quad \pycom{\# Rule 1/6: Every endpoint must be a declared agent\_id}

\quad ids = \pykw{set}([a[\pystr{agent\_id}] \pykw{for} a \pykw{in} agents])

\quad \pykw{for} (u, v) \pykw{in} edges:

\quad\quad \pykw{if} u \pykw{not in} ids \pykw{or} v \pykw{not in} ids:

\quad\quad\quad \pykw{raise} ValueError(\pystr{undefined agent referenced by edge}) \\

\quad \pycom{\# Rule 4: At least one start node (no incoming edges)}

\quad incoming = \pykw{set}([v \pykw{for} (u, v) \pykw{in} edges])

\quad start\_nodes = ids - incoming

\quad \pykw{if} len(start\_nodes) == 0:

\quad\quad \pykw{raise} ValueError(\pystr{no start node}) \\

\quad \pycom{\# Rule 5/9: Exactly one sink node (no outgoing edges)}

\quad outgoing = \pykw{set}([u \pykw{for} (u, v) \pykw{in} edges])

\quad sink\_nodes = ids - outgoing

\quad \pykw{if} len(sink\_nodes) != 1:

\quad\quad \pykw{raise} ValueError(\pystr{must have exactly one sink node}) \\

\quad \pycom{\# Rule 3: Connectivity: reachable from starts and can reach sink}

\quad sink = \pykw{list}(sink\_nodes)[0]

\quad forward\_reach = BFSForward(start\_nodes, edges)

\quad backward\_reach = BFSBackward(\{sink\}, edges)

\quad connected = forward\_reach \pykw{and} backward\_reach

\quad isolated = ids - (forward\_reach \pykw{\&} backward\_reach)

\quad \pykw{if} isolated:

\quad\quad \pykw{raise} ValueError(\pystr{isolated agents exist}) \\

\quad \pycom{\# Rule 7: Must be acyclic (DAG)}

\quad order = TopologicalSort(agents, edges)

\quad \pykw{if} len(order) != len(agents):

\quad\quad \pykw{raise} ValueError(\pystr{cycle detected}) \\

\pykw{def} \pyfn{TopologicalSort}(agents, edges):

\quad \pycom{\# Kahn's algorithm (queue of zero in-degree nodes)}

\quad ids = [a[\pystr{agent\_id}] \pykw{for} a \pykw{in} agents]

\quad in\_deg = \code{\{}i: 0 \pykw{for} i \pykw{in} ids\code{\}}

\quad adj = \code{\{}i: [] \pykw{for} i \pykw{in} ids\code{\}}

\quad \pykw{for} (u, v) \pykw{in} edges:

\quad\quad adj[u].append(v)

\quad\quad in\_deg[v] += 1 \\

\quad Q = Deque([i \pykw{for} i \pykw{in} ids \pykw{if} in\_deg[i] == 0])

\quad order = []

\quad \pykw{while} Q:

\quad\quad u = Q.popleft()

\quad\quad order.append(u)

\quad\quad \pykw{for} v \pykw{in} adj[u]:

\quad\quad\quad in\_deg[v] -= 1

\quad\quad\quad \pykw{if} in\_deg[v] == 0:

\quad\quad\quad\quad Q.append(v)

\quad \pykw{return} order \\

\pykw{async} \pykw{def} \pyfn{GenerateMultiAgentForward}(
\pykw{self}, agents, edges, execution\_order, sink):

\quad global\_node\_model = get\_global(\pystr{global\_node\_model})

\quad \pycom{\# Map agent\_id to its outputs}

\quad results = \code{\{\}} \\

\quad \pykw{for} agent\_id \pykw{in} execution\_order:

\quad\quad cfg = Lookup(agents, agent\_id)

\quad\quad name = cfg[\pystr{agent\_name}].lower()

\quad\quad args = cfg[\pystr{required\_arguments}]

\quad\quad agent\_input = args.get(\pystr{agent\_input}, \pystr{""}) \\

\quad\quad \pycom{\# Replace $\{id\}$ placeholders with previous results}

\quad\quad \pykw{if} agent\_input:

\quad\quad\quad \pykw{for} prev\_id, prev\_out \pykw{in} results.items():

\quad\quad\quad\quad agent\_input = ReplaceAll(
agent\_input,
\pystr{\$\{prev\_id\}},
prev\_out
) \\

\quad\quad \pycom{\# Prepare task\_info: always include original task as context}

\quad\quad \pykw{if} \pykw{not} agent\_input:

\quad\quad\quad task\_info = original\_task\_info

\quad\quad \pykw{else}:

\quad\quad\quad combined = (
\pystr{Original task: } + original\_task\_info.content + \pystr{; Current Sub-task: } + agent\_input
)

\quad\quad\quad task\_info = Info(\pystr{task}, \pystr{User}, combined, \dots) \\

\quad\quad \pycom{\# Dispatch to the chosen agent}

\quad\quad \pykw{if} name == \pystr{cotagent}:

\quad\quad\quad out = \pykw{await} self.CoTAgent(task\_info, model=global\_node\_model)

\quad\quad \pykw{elif} name == \pystr{scagent}:

\quad\quad\quad out = \pykw{await} self.SCAgent(task\_info, model=global\_node\_model)

\quad\quad \pykw{elif} name == \pystr{reflexionagent}:

\quad\quad\quad out = \pykw{await} self.ReflexionAgent(task\_info, model=global\_node\_model)

\quad\quad \pykw{elif} name == \pystr{debateagent}:

\quad\quad\quad out = \pykw{await} self.DebateAgent(
task\_info,
model=global\_node\_model,
debate\_roles=args[\pystr{debate\_roles}]
)

\quad\quad \pykw{elif} name == \pystr{websearchagent}:

\quad\quad\quad out = \pykw{await} self.WebSearchAgent(task\_info, model=global\_node\_model)

\quad\quad \pykw{else}:

\quad\quad\quad \pykw{raise} ValueError(\pystr{unknown agent type}) \\

\quad\quad results[agent\_id] = out \\

\quad \pykw{return} results[sink] \\

\end{subagent}